\documentclass[11pt]{article}
\usepackage[preprint]{acl}

\usepackage{times}
\usepackage{latexsym}
\usepackage[T1]{fontenc}
\usepackage[utf8]{inputenc}
\usepackage{microtype}
\usepackage{inconsolata}
\usepackage{graphicx}
\usepackage{booktabs}
\usepackage{array}
\usepackage{amsmath,amssymb,amsthm}
\usepackage{mathtools}
\usepackage{xcolor}
\usepackage{algorithm}
\usepackage[noend]{algpseudocode}
\usepackage{multirow}
\usepackage[capitalise,noabbrev]{cleveref}

\theoremstyle{plain}
\newtheorem{theorem}{Theorem}
\newtheorem{proposition}[theorem]{Proposition}
\newtheorem{lemma}[theorem]{Lemma}
\newtheorem{corollary}[theorem]{Corollary}
\theoremstyle{definition}
\newtheorem{definition}{Definition}

\theoremstyle{remark}
\newtheorem{remark}{Remark}

\newcommand{\KL}{\mathrm{KL}}
\newcommand{\E}{\mathbb{E}}
\renewcommand{\P}{\mathbb{P}}

\newcommand{\Poi}{\mathrm{Poi}}

\newcommand{\eps}{\varepsilon}
\newcommand{\Th}{\Theta}                 
\newcommand{\cov}{M}                     
\newcommand{\rstar}{r^{\star}}           
\newcommand{\nstar}{n^{\star}}           
\newcommand{\ksz}{\kappa_s}              
\providecommand{\KLinf}{\KL_{\inf}}
\newcommand{\1}{\mathbf{1}}
\newcommand{\defeq}{\coloneqq}

\makeatletter
\newcommand{\probe}[1]{\typeout{PROBE #1 p=\thepage\space col=\if@firstcolumn 1\else 2\fi\space tot=\the\pagetotal}}
\makeatother

\title{Many Steps into One: Reasoning Contraction and the Limits of Safe Memory Reuse for LLM Agents}

\author{Jianghong Zhou \quad Hao Rong \quad Chen Lin \quad Daoan Zhang}

\begin{document}
\maketitle
\begin{abstract}
Thought templates, skill libraries, agent memories and caches turn many reasoning steps into one. We call this \emph{reasoning contraction}. A contracted memory saves reasoning steps and errors until its facts change, then quietly answers wrong. Most systems trust memories by counting successes and failures. We prove that such counters are badly tuned change detectors: a symmetric counter notices a partial change only after a power of its false-alarm horizon, and an unclipped one may never notice it. On five LLMs answering multi-hop questions over edited facts, ExpeL- and ACE-style counters serve $7.0$--$13.5\times$ more stale answers than certified memories given the same information, and on code tasks with an exact verifier $35$--$73\times$ more. Safety also has a price. When facts change and each memory is judged only by its own verified outcomes, one number, the \emph{certification load}, sets how much reasoning any safe memory can skip, and under Zipf demand it yields a scaling law. Our recipe weighs each memory's failures by its own likelihood ratio, certifies its utility against live reasoning, and restarts the certificate when facts change, with the likelihood-ratio detector as the safety net for unannounced changes; for large edits it also invalidates the memory.
\end{abstract}

\section{Introduction}
\label{sec:intro}

\begin{figure*}[t]
\centering
\includegraphics[width=\textwidth]{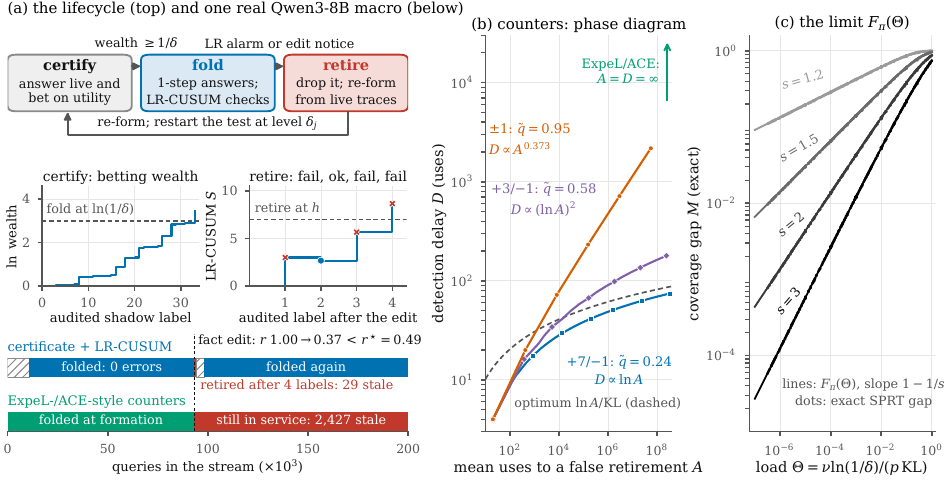}
\caption{\textbf{Trusting a contracted memory is a sequential test.} \textbf{(a)}~Top: the lifecycle of a contracted memory (a \emph{macro}). In \emph{shadow} mode a betting certificate scores the macro's verified answers and \emph{folds} it (one-step answers) once its wealth reaches $1/\delta$; a likelihood-ratio CUSUM (LR-CUSUM) alarm or an edit notice retires it, and the next candidate restarts the test. Middle: both tests on one real macro (Qwen3-8B, \textsc{KQA-Stream}; per candidate, no notices). Its log-wealth reaches $\ln 20$ ($\delta=0.05$) after 33 labels; an edit then drops its reliability from $1.00$ to $0.37$, below break-even ($r^\star=0.49$), and the LR-CUSUM crosses $h=7$ on the fourth label. Bottom: its timeline (hatched: certification); ExpeL- and ACE-style counters have not retired it by stream end (2{,}427 stale answers, against 29). \textbf{(b)}~Exact detection delay $D$ against uses to a false retirement $A$ for three clipped counters (score per failure/success), failure rate $0.05\to0.25$: the symmetric $\pm1$ rule is slow by a power of $A$; likelihood-ratio weights ($+7/{-1}$) track the optimum. $\tilde q$: the failure rate a rule implicitly tests (\cref{rem:implied}). \textbf{(c)}~The exact coverage gap $M$ (share not answered in one step) of a restarted test for 2{,}520 settings (dots) lies on the limit $F_\pi(\Th)$ (lines).}
\label{fig:hero}
\end{figure*}

\paragraph{Everyone builds reasoning memories.} Thought templates \citep{yang2024buffer}, reasoning and workflow memories \citep{ouyang2026reasoningbank,wang2025agent,suzgun2026dynamic}, behavior handbooks \citep{didolkar2025metacognitive}, skill libraries \citep{wang2024voyager,wang2025inducing}, answer caches \citep{bang2023gptcache,schroeder2026vcache}, and the memories that agents share \citep{zhang2025gmemory} or rewrite themselves from \citep{zhang2025darwin} all do one thing: they compile a derivation of $m\ge2$ steps into a one-step \emph{contracted memory} for later instances of the problem. We call this \emph{reasoning contraction}. Trusting such a memory is a sequential test, bound by the limits of change detection and certification.

\paragraph{Live reasoning is costly and fragile.} LLM reasoners buy accuracy with computation: voting over sampled chains \citep{wang2023self}, tree search \citep{yao2023tree}, tool calls \citep{yao2023react} and self-critique \citep{shinn2023reflexion,madaan2023self}. Every step is another chance to err. On the unedited facts of our multi-hop benchmark, Qwen3-8B's live accuracy falls from $0.63$ at two hops to $0.31$ at three and $0.19$ at four (\cref{app:llm-paths}). A verified contracted memory answers in one step; on our edited streams it raises Qwen3-8B's accuracy from $0.379$ to $0.648$ with fewer calls (\cref{tab:realllm}). The price is a new failure mode shared by all these systems: when the facts behind a memory change, it keeps answering, wrongly. This is the utility problem of macro learning \citep{minton1990quantitative}, now for LLM memories over changing facts.

\paragraph{Memories are trusted by counters, which are badly tuned change detectors.} ExpeL and ACE vote memories up and down \citep{zhao2024expel,zhang2026agentic}, and validation-gated memories clip a Hebbian strength \citep{singh2025validation}. Every clipped counter is a cumulative-sum (CUSUM) change detector in disguise, testing an \emph{implied} change that nobody chose (\cref{sec:hebbian}). A symmetric counter assumes that a stale memory is as often wrong as a fresh one was right, so counting votes equally is provably the wrong detector for \emph{partial} staleness: when an edit leaves a memory failing less than half the time, the counter's evidence drifts the wrong way, and its delay grows as a power of its false-alarm horizon, not its logarithm (\cref{thm:trichotomy}, \cref{fig:hero}b). Unclipped counters are worse: credit banked while the memory was healthy must be spent before retirement. On \textsc{KQA-Stream}, multi-hop questions over controlled knowledge-base edits, ExpeL- and ACE-style counters leave about half of the same $156$ stale memories in service at stream end, against $2\%$ for a likelihood-ratio rule, and need about $60\times$ more uses to retire them (restricted mean; \cref{fig:km}). On five open LLMs they serve $7.0$--$13.5\times$ more stale answers than our certificate given the same information (no notices; \cref{tab:models}); on code, larger still (\cref{sec:experiments}).

\paragraph{Safety has a price.} Can a memory be safe and still skip most reasoning? One number decides: the \emph{certification load} $\Th=\nu\ln(1/\delta)/(p\KL)$, which combines the rate of change ($\nu$), the risk tolerated ($\delta$), the audit rate ($p$) and how much one verified outcome reveals ($\KL$). Whatever its algorithm, every safe memory must, in the worst case and as $\delta\to0$, leave a share $F_\pi(\Th)$ of queries to multi-step reasoning, and a restarted test achieves it (\cref{thm:limit}, \cref{fig:hero}c). Under Zipf demand it scales as $\ksz\Th^{1-1/s}$ (\cref{cor:zipf}): popular problems can be contracted; rare ones change before they can be trusted.

\paragraph{The recipe.} Three changes turn a counter into a \emph{certificate}, a rule that lets a memory answer in one step only while the risk of trusting a bad one stays below $\delta$ (\cref{sec:certify}). (i)~Weigh each memory's failures against its successes by its own likelihood ratio, set by its healthy and stale failure rates; on \textsc{KQA-Stream} this detects staleness in $1.3$--$5.9\times$ fewer uses than the best tested fixed weighting. (ii)~Certify utility, not accuracy, by betting on the paired difference with live reasoning on the same query, with no estimate of live accuracy. (iii)~Restart a memory's certificate whenever its answers or the live reasoning it competes with change, so every trust decision rests on current evidence. Where edits are substantial, our deployment also invalidates memories an edit announces, and the likelihood-ratio detector is the safety net for unannounced edits; it cuts stale answers $9$--$12\times$ against invalidation alone. On five models it trusts under one below-break-even memory per stream (with notices; counters: $111$--$595$), and, scoring answers alone, makes no false certification at $\delta\le0.1$. Agent memories and skill libraries with a verifier should therefore adopt these changes, and memory benchmarks should edit facts.

\paragraph{Contributions and what is new.} Our results apply when facts can change and each memory can be judged only by its own verified outcomes. (1)~\emph{A concept}: reasoning contraction, one lens on templates, skills, memories and caches, whose two risky decisions, trust and retirement, are sequential tests bound by the limits of change detection and certification (\cref{fig:hero}a); done safely, \textsc{KQA-Stream}'s symbolic macros add $14$--$33$ points over live reasoning. (2)~\emph{Two theorems}: the phase diagram of counter maintenance (equal votes: delay $\asymp A^{0.373}$, not $\ln A$, at failure rate $0.05\to0.25$), with the learning rates that fix it (\cref{thm:trichotomy,cor:lr}), and the certification-load limit with its scaling law $\ksz\Th^{1-1/s}$, matched within $0.945$--$1.025\times$ by a restarted test (\cref{thm:limit,cor:zipf}). (3)~\emph{A certificate}: a per-template certify--retire--restart lifecycle from standard anytime-valid tools \citep{yu2026anytime}, tested on \textsc{KQA-Stream} (with invalidation, no stale answers under notices; $7.0$--$13.5\times$ fewer than counters without), natural Wikidata edits and code. Where memory systems vote and anytime-valid routing certifies one global threshold, we test each memory's trust over its lifecycle: memory systems with a verifier should give each memory its own weight and certificate, and price skipped reasoning by $\Th$.

\section{Contraction as a memory lifecycle}
\label{sec:setting}

\paragraph{Queries, templates, macros.}
A reasoner serves a stream of queries. Each query instantiates a \emph{template}~$i$, a canonical derivation skeleton such as a relation path in a knowledge base, a lemma chain, or a tool-call sequence. Template $i$ is drawn with probability $\pi_i$ and has a \emph{live} derivation of $m_i\ge2$ steps (LLM calls). A \emph{macro} for template $i$ is a one-step operator formed from a verified derivation: a materialized composed relation, a stored lemma, or a distilled direct-answer note. Its \emph{reliability} $r$ is the probability that it answers a random instance of $i$ correctly. On each query the system answers \emph{live} (cost $m$, correct with probability $a$), \emph{folded} (cost $1$, answer from the trusted macro), or in \emph{shadow} mode (live answer, macro evaluated alongside). A verifier (exact match, execution, or a unit test) labels the macro's answer on \emph{audited} queries, a fraction $p\in(0,1]$ of the template's shadow or folded queries. With error price $L$ (in steps), a folded query costs $1+L(1-r)$ and a live one $m+L(1-a)$, so folding pays if and only if
\begin{equation}
r>\rstar\defeq a-(m-1)/L .
\label{eq:breakeven}
\end{equation}

\paragraph{The lifecycle.}
Contraction is a memory lifecycle (\cref{fig:hero}a): \emph{form} a candidate from a verified trace, \emph{certify} it in shadow mode, \emph{fold} it, \emph{monitor} it, \emph{retire} it when it goes stale, and \emph{re-form}. Two decisions carry all the risk: \emph{when may a candidate be trusted} (certification) and \emph{when has a trusted macro gone stale} (retirement). Verifier outcomes $Y\in\{0,1\}$ of a macro are conditionally Bernoulli given the past, with mean equal to its current reliability; this allows adaptive usage and drift. A macro is \emph{stale} once an edit has raised its failure rate: healthy macros fail with probability $q_0$, stale ones with $q_1>q_0$.

\paragraph{Maintenance rules in current systems.}
ExpeL gives a new insight importance $2$, adds $1$ per up-vote or edit, subtracts $1$ per down-vote and deletes it at $0$ \citep{zhao2024expel}. ACE keeps unbounded helpful/harmful counts, and its released code flags a bullet when harmful $\ge$ helpful and harmful $>0$ \citep{zhang2026agentic}; our ACE-style baseline treats a flag as retirement. We idealize their LLM-issued votes as one vote per verified use. Validation-gated Hebbian memories clip a strength to $[0,1]$ with time decay \citep{singh2025validation}; we analyze the symmetric variant $c\gets\min\{1,c+\eta_+\}$ on success and $c\gets c-\eta_-$ on failure, retiring below $\tau$.
\section{Memory counters are change detectors}
\label{sec:hebbian}

A \emph{score rule} assigns a bounded score $f(Y)$ to each verifier outcome, keeps
\begin{equation}
S_n=\max\{0,\,S_{n-1}+f(Y_n)\},
\label{eq:scorerule}
\end{equation}
and retires the macro at the first $n$ with $S_n\ge h$.
Clipped Hebbian confidence has this form. With $S=(1-c)/\eta_-$ the score is $+1$ on a failure and $-\rho$ on a success, $\rho=\eta_+/\eta_-$, and the clip at full confidence is the reflection at $0$ (\cref{app:hebbian}). Let $P_0,P_1$ be the outcome laws of a healthy and a stale macro, with $\E_0 f<0<P_0(f>0)$, and let $\theta_j>0$ solve $\E_j e^{\theta_j f}=1$ (the Cram\'er--Lundberg \emph{adjustment coefficient} under $P_j$). We measure a rule by $A$, the mean number of healthy uses before it falsely retires a macro, and by $D$, its worst-case number of uses to detect a change \citep{lorden1971procedures}.

\begin{remark}[Implied alternative]
\label{rem:implied}
Every score rule \eqref{eq:scorerule} is exactly Page's CUSUM \citep{page1954continuous} for the change $P_0\to\tilde P_0\defeq e^{\theta_0 f}P_0$ and for no other change, as for quality-control charts \citep{munford1980control,brook1972approach}. The symmetric Hebbian rule ($\rho=1$) thus tests failure rate $q_0$ against its \emph{mirror} $1-q_0$: it assumes that a stale memory is as wrong as a fresh one was right (reliability $0.95\to0.05$), whereas a fact edit may corrupt only part of a template's instances (reliability $0.95\to0.75$).
\end{remark}

\begin{theorem}[Phase diagram of maintenance rules]
\label{thm:trichotomy}
Let $\mu_1=\E_1 f$ and $\gamma=\theta_1/\theta_0$. As $A\to\infty$,
\begin{align*}
\mu_1>0:&\quad D=\tfrac{\ln A}{\theta_0\mu_1}+O(1);\\
\mu_1=0:&\quad D=\tfrac{(\ln A)^2}{(\theta_0\sigma_1)^2}+O(\ln A),\ \sigma_1^2=\mathrm{Var}_1 f;\\
\mu_1<0:&\quad D\asymp A^{\gamma}\ \text{ if }P_1(f>0)>0,
\end{align*}
where $\gamma<1$ if and only if $\E_1e^{\theta_0 f}>1$, e.g.\ whenever $q_1>q_0$ for Bernoulli rules.
In the first regime the efficiency relative to the optimal delay $\ln A/\KL(P_1\|P_0)$ \citep{lorden1971procedures} is
\begin{equation}
\mathrm{Eff}=\frac{\theta_0\mu_1}{\KL(P_1\|P_0)}=1-\frac{\KL(P_1\|\tilde P_0)}{\KL(P_1\|P_0)} ,
\label{eq:eff}
\end{equation}
so $\mathrm{Eff}\le1$, with equality if and only if $f$ is proportional to the log-likelihood ratio.
\emph{When it applies:} any clipped counter with a bounded score per verified outcome, when outcomes are i.i.d.\ before and after one change (constants in \cref{app:hebbian}).
\end{theorem}

\paragraph{Why three regimes.} After a change the score drifts at rate $\mu_1$. With positive drift it climbs to the threshold $h\approx\ln A/\theta_0$ in linear time, hence $\ln A$. With zero drift it diffuses and needs time $h^2$. With negative drift it must make a rare excursion, like a false alarm but at the post-change exponent $\theta_1<\theta_0$, hence $e^{\theta_1h}=A^{\gamma}$. The sign is that of $\theta_0\mu_1=\KL(P_1\|P_0)-\KL(P_1\|\tilde P_0)$: \emph{a rule detects a change quickly only if the change moves the outcome law closer to the rule's implied alternative than to the healthy law.} For the symmetric rule this requires $q_1>\tfrac12$. Partial staleness (reliability $0.95\to0.75$) gives $\gamma=\ln 3/\ln 19=0.373$ (\cref{fig:hero}b). The logarithmic regime is known (\cref{sec:related}); the power-law regime and the unified diagram are new. \emph{Why it matters:} under partial staleness, equal votes wait $A^{\gamma}$ uses when clipped, not $\ln A$, and with positive probability forever when unclipped (\cref{rem:unclipped}).
\begin{corollary}[Likelihood-ratio learning rates]
\label{cor:lr}
A clipped Hebbian rule is the likelihood-ratio CUSUM for $q_0\to q_1$, optimal at the false-alarm horizons its thresholds attain, if and only if
\begin{equation}
\frac{\eta_+}{\eta_-}=\rho(q_0,q_1)\defeq\ln\frac{1-q_0}{1-q_1}\Big/\ln\frac{q_1}{q_0} .
\label{eq:rho}
\end{equation}
\emph{When it applies:} a memory whose healthy and stale failure rates $q_0<q_1$ can be estimated (if only an interval for $q_1$ is known, tuning at its lower end is first-order minimax; \cref{app:hebbian}).
\end{corollary}
We use the Bernoulli likelihood-ratio CUSUM (\emph{LR-CUSUM}) of \citet{reynolds1999cusum} to set memory learning rates. Under partial staleness ($q_0=0.05$, $q_1=0.25$), $\rho=0.147$: a failure must weigh about $7$ successes, not $1$ as in ExpeL and ACE. Scores $+7/{-1}$ detect in $41$ and $54$ uses at $A\approx1.6\times10^5$ and $3\times10^6$, $5.7\times$ and $13\times$ faster than the symmetric rule (\cref{fig:hero}b). \emph{Why it matters:} weighting each memory by its own $\rho$ makes it the optimal detector of its $q_0\to q_1$; if memories' $\rho$ differ, no global weight is optimal for all.

\begin{remark}[Unclipped counters]
\label{rem:unclipped}
An unclipped counter such as ExpeL's or ACE's (\cref{sec:setting}) rarely retires a healthy memory ($A=\infty$) but has no finite worst-case delay: credit banked before a change must be spent first, at rate $2q_1-1$. Under partial staleness ($q_1<\tfrac12$) the stale memory is \emph{never} retired with probability at least $1-2q_1$ (ACE) or $1-(q_1/(1-q_1))^2$ (ExpeL), i.e.\ $0.50$ and $0.89$ for reliability $0.95\to0.75$ (\cref{prop:unclipped} in \cref{app:hebbian}).
\end{remark}

\section{The price of safety: how far can reasoning contract?}
\label{sec:limit}

How large a share of queries can \emph{any} safe system answer in one step when facts keep changing?

\paragraph{Model.} Queries of template $i$ arrive at rate $\pi_i$ ($\sum_i\pi_i=1$), and each is audited with probability $p$; an audit reveals one verifier outcome of the candidate it evaluates. Each template's facts change at rate $\nu$. After a change the current macro is invalid, and a candidate must earn trust in the new \emph{epoch}. A candidate is \emph{good} if its reliability is $\ge r_1$ and \emph{bad} if it is $\le r_0<r_1$ (e.g.\ $r_0=\rstar$ of \cref{eq:breakeven}). A \emph{fold event} is a time when the system starts answering a template in one step with some candidate, formed in this or an earlier epoch. The \emph{coverage gap} $\cov$ is the long-run fraction of queries \emph{not} answered in one step by a valid, certified macro.

\begin{definition}[$\delta$-safety]
\label{def:safety}
A policy is $\delta$-safe if, for every epoch, every history before it and every outcome law within it, the probability that the epoch contains a fold event of a candidate that is bad in that epoch is at most $\delta$.
\end{definition}

Safety is per epoch, not per candidate: folding each of many candidates blindly with probability $\delta$ is safe per candidate but eventually folds a bad one. A sequential probability ratio test (SPRT), $\Lambda_n=\sum_{k\le n}\ln\frac{P_{r_1}(Y_k)}{P_{r_0}(Y_k)}$, that folds when $\Lambda_n\ge\ln(1/\delta)$ and restarts at each change is $\delta$-safe by Ville's inequality \citep{wald1945sequential,ville1939etude}.

\begin{theorem}[Fundamental limit of safe contraction]
\label{thm:limit}
Let $\KL=\KL(r_1\|r_0)$ (Bernoulli), define the \emph{certification load}
\begin{equation}
\Th\defeq\frac{\nu\,\ln(1/\delta)}{p\,\KL},
\label{eq:load}
\end{equation}
and let $F_\pi(\Th)\defeq\sum_i\pi_i\bigl(1-e^{-\Th/\pi_i}\bigr)$.
\emph{(i)~No safe policy does better:} every $\delta$-safe policy has $\cov\ge(1-\eps_\delta)F_\pi(\Th)$, where $\eps_\delta\to0$ as $\delta\to0$, whether or not changes are announced and however many candidates it forms. \emph{(ii)~A restarted test achieves it:} the SPRT restarted at each change is $\delta$-safe and has $\cov\le(1+\ln(r_1/r_0)/\ln(1/\delta))\,F_\pi(\Th)$.
\emph{When it applies:} facts change at random times, each audit reveals one outcome, only audited outcomes are informative (shadow agreement excluded; \cref{thm:app-agree}), and good memories have reliability only $r_1$ (the worst case; exact model in \cref{app:limit}).
\end{theorem}

\paragraph{Why $F_\pi(\Th)$.} A good candidate cannot be told apart from a bad one before its log-likelihood ratio reaches $\ln(1/\delta)$, which takes $\nstar=\ln(1/\delta)/\KL$ audited uses, i.e.\ time $\nstar/(p\pi_i)$. A query is folded only if its epoch is older than that. Epoch ages are exponential with rate $\nu$, so this happens with probability $e^{-\Th/\pi_i}$. Popular templates certify quickly and are almost always covered; templates with $\pi_i\lesssim\Th$ change before certification. A change of measure makes this rigorous \citep{lai1981asymptotic,tartakovsky2014sequential,kaufmann2016complexity}. Under Zipf demand it becomes a scaling law:

\begin{corollary}[Scaling law of one-step reasoning]
\label{cor:zipf}
If $\pi_i=i^{-s}/\zeta(s)$ with $s>1$, then for all $\Th>0$
\begin{equation}
\ksz\Th^{1-1/s}-\Th\ \le\ F_\pi(\Th)\ \le\ \ksz\Th^{1-1/s},
\label{eq:zipflaw}
\end{equation}
with $\ksz=\Gamma(1/s)/((s-1)\zeta(s)^{1/s})$ ($\kappa_2\approx1.38$). Hence the optimal gap is $\ksz\Th^{1-1/s}(1+o(1))$ for small $\delta$ and small load.
\end{corollary}

\paragraph{Reading the law.} Drift, risk, audit rate, reliability gap and verifier noise (through the $\KL$ of the \emph{observed} channel) enter only through $\Th$, and the exponent $1-1/s$ is set by the demand tail alone. \emph{Why it matters:} in this model no safe algorithm, candidate supply or notice beats the limit as $\delta\to0$; for given demand only a lower load contracts more. \Cref{fig:hero}c checks the law exactly: the gap of the restarted SPRT, computed by dynamic programming for 2{,}520 settings of reliabilities, verifier noise, $\delta\in[10^{-32},0.03]$, audit and drift rates, has $\cov_{\rm SPRT}/F_\pi(\Th)\in[0.945,1.025]$, and the fitted log-log slopes are within $0.008$ of $1-1/s$ (finite-$\delta$ bounds: \cref{app:limit}).

\providecommand{\KLinf}{\KL_{\inf}}
\section{The recipe: a certificate that needs no live accuracy}
\label{sec:certify}

The SPRT of \cref{thm:limit} tests $r\le\rstar$, but $\rstar$ involves the unknown live accuracy $a$ (\cref{eq:breakeven}), and plugging in an estimate breaks the guarantee.
\begin{proposition}[Plug-in tests are not level $\delta$]\label{prop:plugin}
If the SPRT uses $r_0=\hat a-\frac{m-1}{L}$ and $r_1=\hat a$ with $\hat a$ estimated from $n_a$ live runs, its false-fold probability at break-even can exceed $\delta$. At $a=0.9$ and $\frac{m-1}{L}=\delta=0.05$ it is $0.34$, $0.25$ and $0.08$ for $n_a=20$, $100$ and $1000$ (exact; \cref{prop:app-plugin}).
\end{proposition}
Following testing by betting \citep{shafer2021testing,waudbysmith2024estimating}, with inverse-propensity weights as in B-PAC \citep{yu2026anytime}, we bet on the paired utility of each shadow evaluation, with live reasoning on the \emph{same} instance as control: $U_t=C^L_t-C^M_t+L(Y^M_t-Y^L_t)$, the steps saved plus $L$ times the correctness gained, where $C^L_t,C^M_t$ are the costs and $Y^L_t,Y^M_t\in\{0,1\}$ the verified correctness of the live and macro answers; $Z_t=U_t/B\ge-1$ for a known bound $B$.
\begin{theorem}[Betting certificate]\label{thm:betting}
Fold when the wealth of predictable bets on the inverse-propensity-weighted $Z_t$ reaches $1/\delta$.
(i)~\emph{Validity.} A macro that is never more useful than live reasoning is folded with probability at most $\delta$, whatever the live accuracy and however live and macro errors are correlated.
(ii)~\emph{Rate.} If the unweighted $Z_t$ are i.i.d.\ with law $P_Z$ and audits have rate $p$, certification takes $(1+o(1))\ln(1/\delta)/(p\,\KLinf(P_Z))$ uses, and no $\delta$-safe rule that sees only audited utilities is faster \citep{kaufmann2016complexity,garivier2016optimal}. Here $\KLinf(P)\defeq\inf\{\KL(P\|Q):\E_QZ\le0\}$ over laws $Q$ on the same bounded range: the divergence from $P_Z$ to the nearest useless macro.
\emph{When it applies:} audits have known probabilities, bounded below, independent of the hidden outcome (exact conditions in \cref{app:utility}).
\end{theorem}
Under the null, reaching $1/\delta$ is rare (Ville's inequality); a useful macro grows log-wealth by $\KLinf(P_Z)$ per audited pair \citep{honda2010asymptotically}. For a certificate restarted at each change, \cref{thm:limit} holds with $\KLinf(P_Z)$ for $\KL$ and unknown $a$ (\cref{cor:app-scaling}). Unlike a discordant-pair test \citep{shawn2026pace}, it can certify macros less accurate than live but above break-even (\cref{thm:app-pairing}).

\paragraph{From certificate to deployment.} The \emph{restart-only} certificate (\cref{tab:algos}) runs a template's $j$-th test at level $\delta_j=6\delta/(\pi^2j^2)$, so that all its tests together spend at most $\delta$; it \emph{restarts}, at the next level, at each notice that the template's answers changed, and retires a folded macro by its own LR-CUSUM (\cref{cor:lr}). Our \emph{recommended deployment} (``ours'') also invalidates and re-forms folded macros at each notice, so the LR-CUSUM is the safety net for missed notices. \emph{When it applies:} both are $\delta$-safe per epoch for their own null, paired utility $\le0$, when notices cover every change of that null, including changes of live reasoning (\cref{prop:app-ledger}). So one $\delta$ bounds \emph{false certification}, a fold of a candidate whose expected paired utility is $\le0$ at the fold, in every epoch, however many successive candidates and edits; later staleness is left to retirement (\cref{sec:hebbian}).

\begin{table}[t]
\centering
\scriptsize
\setlength{\tabcolsep}{2pt}
\renewcommand{\arraystretch}{1.05}
\begin{tabular}{@{}>{\raggedright\arraybackslash}p{1.95cm}>{\raggedright\arraybackslash}p{1.05cm}>{\raggedright\arraybackslash}p{4.38cm}@{}}
\toprule
Certificate & Needs & Guarantee \\
\midrule
\textbf{Restart only}: ledger $\delta_j$, restarts & notices$^\dagger$ & $\delta$-safe per epoch (\cref{prop:app-ledger}); \emph{as run}, growth $\KLinf/(1.03$--$1.07)$ on the laws checked, plus $2\ln j+O(1)$ nats (\cref{app:ledger}) \\
\addlinespace[1pt]
\textbf{Ours} (recommended): + invalidate on notice & notices$^\dagger$, some missed & as above; a missed notice adds the carried-over term of \cref{prop:app-ledger}(iii) \\
\addlinespace[1pt]
\textbf{No notice}: discard at own alarm & none & $\delta$-safe for folds on post-change evidence (\cref{prop:app-detect}); not per epoch if $\alpha_ir_0>\delta$ (\cref{prop:app-nonotice-lb}) \\
\addlinespace[1pt]
\textbf{Per candidate}: level $\delta$ each & none & per candidate only (\cref{fig:hero}a, code) \\
\bottomrule
\end{tabular}
\caption{\textbf{Restarting at each notice makes certification safe per epoch.} Bold: ours. All retire folded macros by LR-CUSUM (\cref{cor:lr}). The \emph{optimum}, an SPRT or grid bet restarted at known change times, attains $F_\pi(\Th)$ (\cref{thm:limit,cor:app-scaling}); \emph{as run} is a rule's own growth rate; \emph{measured} rates are in \cref{sec:experiments}. $^\dagger$Of every change of the null, including live reasoning's (\cref{app:r4-silent}).}
\label{tab:algos}
\end{table}

\section{Experiments}
\label{sec:experiments}

\paragraph{Benchmark.} \textsc{KQA-Stream} is built from the KQA Pro knowledge base \citep{cao2022kqa}: 300 templates of 2--4 relation hops with 60 instances each (18k LLM-verbalized questions) and three sequences of 24 fact edits (\cref{app:experiments}). The \emph{live} reasoner is a hop-by-hop agent that chooses one relation per LLM call. A \emph{macro} materializes a verified trace's relation path as an anchor-to-answer table, reached through one routing call; it goes stale when edits change the facts it cached. A \emph{compiled plan} stores the path and re-executes it on the current knowledge base. Every LLM call is recorded once per model, with exact selective re-runs after edits, and policies are replayed on 200k-query Zipf streams ($s=1.2$; 10 seeds $\times$ 3 edit sequences; 10\% random audits), so all policies see identical queries, edits and audit coins. We use five open models: Qwen3-8B and Qwen3-4B \citep{yang2025qwen3} (thinking disabled), Granite-3.3-8B \citep{ibm2025granite}, Mistral-7B \citep{jiang2023mistral} and OLMo-3-7B \citep{olmo2025olmo3}. Prompts, seeds and scripts are in our code release.

\paragraph{Policies.} \emph{Fold once} folds a verified candidate at formation and never retires it, like a semantic cache. ExpeL- and ACE-style baselines transplant each system's retention rule onto our macros, one vote per audited outcome. Detector baselines certify with the plug-in SPRT and retire with an LR-CUSUM; the \emph{optimistic} one folds at formation. The certificates of \cref{tab:algos} run at $\delta=0.05$; a \emph{notice} is an edit that changes the gold answer of a query routed to the template; a \emph{vCache-style} cache re-implements \citet{schroeder2026vcache} (\cref{app:focus-vcache}; more variants: \cref{tab:app-full}). \emph{Metrics:} \emph{stale} answers are wrong one-step answers that were right when the macro was formed; a \emph{bad fold} trusts a candidate below break-even; a \emph{stale episode} runs from the edit that makes a folded macro stale to its retirement or the stream end; $J_{20}$ is LLM calls per query plus $20\bar m$ times the error rate ($\bar m$: mean live cost; $L=20\bar m$ in \cref{eq:breakeven}; lower is better). Audits query the knowledge base at no LLM-call cost.

\begin{table}[t]
\centering
\scriptsize
\setlength{\tabcolsep}{2.1pt}
\begin{tabular}{@{}lrrrrrrrr@{}}
\toprule
& \multicolumn{3}{c}{stale per $10^4$ queries} & \multicolumn{3}{c}{unretired at stream end} & \multicolumn{2}{c}{bad folds}\\
\cmidrule(lr){2-4}\cmidrule(lr){5-7}\cmidrule(l){8-9}
Model & ExpeL & ACE & \textbf{ours}$^\circ$ & ExpeL & ACE & \textbf{LR} & counters & \textbf{ours} \\
\midrule
Qwen3-8B   & 967 & 946 & 72 & 53\% & 53\% & 3\% & 284--400 & 0.0 \\
Qwen3-4B   & 795 & 783 & 85 & 57\% & 51\% & 0\% & 111--148 & 0.1 \\
Granite-8B & 657 & 650 & 82 & 22\% & 22\% & 0\% & 282--369 & 0.4 \\
Mistral-7B & 298 & 294 & 42 & 25\% & 25\% & 0\% & 486--595 & 0.6 \\
OLMo-3-7B  & 480 & 445 & 51 & 64\% & 55\% & 9\% & 172--211 & 0.3 \\
\bottomrule
\end{tabular}
\caption{\textbf{Counters serve $7.0$--$13.5\times$ more stale answers on every model, even when ours gets no edit notice.} $^\circ$Ours with every notice missed, i.e.\ the counters' information; with notices it serves none. Unretired: share of the stale episodes of \cref{fig:km} (66, 53, 18, 8 and 11) still in service at stream end. Bad folds: per stream (ours: with notices). Bold: ours.}
\label{tab:models}
\end{table}

\begin{figure}[t]
\centering
\includegraphics[width=\columnwidth]{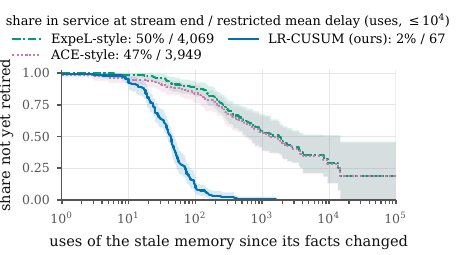}
\caption{\textbf{Counters leave half the stale memories in service at stream end; likelihood ratios retire them about $60\times$ sooner (restricted mean).} Kaplan--Meier share of the 156 stale memories not yet retired (same audits for all rules; counters start at the fold without credit). Legend: share still in service when the stream ends (open episodes, so it differs from the curve's tail at $10^4$ uses) / restricted mean delay ($\le10^4$ uses). Bands: 95\% template bootstrap.}
\label{fig:km}
\end{figure}

\paragraph{Counters leave stale memories in service.} As \cref{rem:unclipped} predicts, counters serve $7.0$--$13.5\times$ more stale answers than ours on every model, even with every notice missed (\cref{tab:models}). To isolate detection, we replay every macro that restart only folds, which only detection retires, under each rule with the same audits: $156$ stale episodes (\cref{app:acl3}). Counters leave $50\%$ and $47\%$ in service at stream end against $2\%$ for the LR-CUSUM, with restricted mean delays of $4{,}069$ and $3{,}949$ uses against $67$ (\cref{fig:km}); each episode costs counters about $1{,}350$ stale answers, against $38.5$. Loss hides this: their $J_{20}$ is close to ours because, skipping certification, they answer $28$--$76\%$ of queries in one step (ours: $20$--$53\%$), so benchmarks must count stale answers.

\paragraph{Detection follows the phase diagram.} A \emph{false retirement} (FR) retires a macro still above break-even. Staleness here is nearly total (mean $q_1=0.75$--$0.90$, $q_0=0.04$--$0.24$), and \cref{thm:trichotomy,cor:lr} predict three things, all observed. \emph{(i)~The symmetric rule detects, but slowly}: with $q_1>\tfrac12$ its delay is logarithmic, but the LR-CUSUM needs $3.7$--$6.1\times$ fewer uses at $0.8$--$5$ FR per $10^4$ healthy uses (\cref{fig:app-detect}). \emph{(ii)~Per-memory weights beat the best tested global one}, as $q_0$ varies: $1.3$--$5.9\times$ fewer uses (significant on three of five models; \cref{app:acl2-decomp}); that best ratio is itself small (below $1{:}1$ on four models; predicted $1/\rho$: $0.67$--$2.14$; \cref{fig:e1}). \emph{(iii)~Partial staleness moves the optimum up}: on $217$ real partially stale episodes (predicted $1/\rho=4.8$) the best tested weight is $5$--$7$, and the symmetric rule needs $1.2$--$1.5\times$ more uses (\cref{app:partial}). The theory thus ranks the rules on real LLM errors.

\begin{figure}[t]
\centering
\includegraphics[width=\columnwidth]{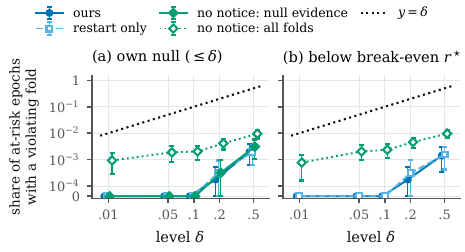}
\caption{\textbf{False certifications stay far below $\delta$.} At-risk epochs with a fold that violates \textbf{(a)}~the certificate's own null, the event $\delta$ bounds, or \textbf{(b)}~break-even $r^\star$; answers scored alone; five models. Without notices, folds on null evidence stay at zero up to $\delta=0.1$, and all folds more than $10\times$ below $\delta$. Bars: 95\% template bootstrap.}
\label{fig:epoch}
\end{figure}

\paragraph{Restarted certificates are safe in every epoch.} Scoring answers alone, as the certificate's paired test does (\cref{app:r3-safety}), ours makes no false certification (\cref{sec:certify}) at $\delta\le0.1$ ($0$ of $5{,}779$ at-risk epochs at $\delta=0.05$) and stays far below $\delta$ up to $0.5$, whether ``bad'' means null at the fold or below break-even (\cref{fig:epoch}); counting routing errors as wrong, $0.26\%$ of at-risk epochs have a bad fold at $\delta=0.05$ (\cref{app:r3-info}). Restarts remove the $11$ bad folds that candidates good at formation carry across edits without them (standard scoring; \cref{app:acl2-epoch}). Although $16\%$ of template-versions change live reasoning unannounced, none of the $9{,}835$ answer-scored folds ($2$ of $8{,}294$ under standard scoring) rests on a test spanning such a change (\cref{app:r4-silent}). $\delta$ controls which memories are trusted; how often a trusted memory errs is set by $r^\star$ ($10$--$11\%$ of one-step answers, mostly from macros above break-even).

\begin{table}[t]
\centering
\small
\setlength{\tabcolsep}{3pt}
\resizebox{\columnwidth}{!}{%
\begin{tabular}{llrrrrr}
\toprule
& Policy & Acc & Calls & Stale & Bad & $J_{20}$ \\
\midrule
no memory & live always & .379 & 3.26 & 0 & 0 & 43.8 \\
\midrule
\multirow{3}{*}{no retire} & fold once & .534 & 1.45 & 1290 & 34 & 31.8 \\
& compiled plan, fold once & .639 & 1.45 & 95 & 34 & 24.8 \\
& vCache-style, $\delta_c=.02$ & .377 & 2.86 & 30 & -- & 43.5 \\
\midrule
\multirow{2}{*}{counters} & ExpeL-style counter & .613 & 1.75 & 967 & 284 & 27.0 \\
& ACE-style counter & .618 & 1.86 & 946 & 400 & 26.7 \\
\midrule
\multirow{2}{*}{detectors} & SPRT + LR-CUSUM & .622 & 2.64 & 54 & 0.7 & 27.3 \\
& optimistic + LR-CUSUM & .632 & 1.58 & 573 & 82 & 25.5 \\
\midrule
\multirow{4}{*}{certified} & \textbf{ours (recommended)}$^\star$ & \textbf{.648} & 2.56 & 0 & 0.0 & 25.5 \\
& \textbf{restart only} & \textbf{.642} & 2.57 & 66 & 0.0 & 26.0 \\
& \textbf{no notice (own detector)} & \textbf{.650} & 2.52 & 70 & 0.1 & 25.3 \\
& \textbf{compiled plan + betting} & \textbf{.668} & 2.46 & 16 & 0.0 & 24.1 \\
\bottomrule
\end{tabular}}
\caption{\textbf{Certification removes bad folds and invalidation removes stale answers, with fewer calls than live reasoning} (Qwen3-8B, 30 streams of 200k queries). Calls: LLM calls per query. Stale: per $10^4$ queries. Bad: bad folds per stream (--: no macros). $J_{20}$: loss, lower is better ($\bar m=3.26$). Bold: ours. $^\star$Perfect notices; with every notice missed it serves $72$ stale answers per $10^4$ queries. Compiled plans (with restarts) gain $+2.6$ points [$1.7$, $3.8$] over restart-only macros (paired 95\% template bootstrap).}
\label{tab:realllm}
\end{table}

\paragraph{Certified contraction pays.} On Qwen3-8B, ours raises accuracy from $0.379$ (live) to $0.648$ with $21\%$ fewer LLM calls, and certified compiled plans to $0.668$ (\cref{tab:realllm}), because a macro executes a verified path that live reasoning re-derives with errors. Uncertified reuse (fold once) serves $1{,}290$ stale answers per $10^4$ queries; certification cuts this $20\times$, invalidation to zero. Across models ours adds $14$--$33$ points over live reasoning (\cref{tab:app-models}).

\paragraph{The detector matters exactly where notices fail.} With perfect notices ours serves no stale answers and matches restart only in accuracy ($+0.24$ points [$-0.24$, $+0.81$], pooled). With $80\%$, $50\%$ or no edits announced, invalidation alone serves $125$, $325$ and $810$ stale answers per $10^4$ queries, ours $13$, $28$ and $66$ (\cref{app:r3-info}), $9$--$12\times$ fewer: invalidation needs the detector wherever notices can be missed.

\paragraph{Natural edits: safety holds, and mild staleness shows its price.} On a 1992--2016 timeline of Wikidata temporal qualifiers (four models), where changes are small and partial (median $3.3\%$ of a template's answers per year), ours keeps bad-fold epochs below $\delta$ at every level, the LR-CUSUM detects in $1.1$--$1.6\times$ fewer uses than the best tested fixed weight, and counters leave $84\%$ of the same stale memories in service against $37\%$ (\cref{app:natural-edits,app:r4-natural-sens}). The price of safety (\cref{sec:limit}) shows: counters make $149$--$207$ bad folds per stream against $0.1$ and serve $3\times$ more stale answers than restart only, yet are $3.5$ points more accurate because folding pays when staleness is mild; for the same reason invalidation costs ours $1.9$ points against restart only.

\paragraph{On code, counters serve $35$--$73\times$ more stale answers.} On $521$ EvalPlus tasks \citep{liu2023evalplus} (five models), live reasoning writes, tests and repairs a program; a memory is a verified program rerun on its task's later queries; cost counts steps, an execution being one. Memories go stale under announced requirement revisions and unannounced Python API changes, which also change live reasoning and so exercise the safety net. Ratios are per query, against our certificate given the counters' information (per candidate, no notices), on all five models ($38$--$86\times$ in LLM-call accounting). Here most stale memories never failed before ($q_0=0$, so $\rho\to0$), and the per-memory LR-CUSUM leaves them folded for $10\times$ fewer uses than the best fixed weight at matched false retirements ($7.6$--$11.4\times$ per model, weights up to $5000{:}1$; \cref{app:codestream}). Ours folds a below-break-even candidate in $12$ of $946{,}678$ epochs under test (counters: $4.8$--$5.4\%$) and gains $8.7$ accuracy points [$7.1$, $10.5$] over live reasoning, with lower $J_{20}$ in both units: it uses $23\%$ fewer steps, while its shadow tests add $17\%$ LLM calls. Zero-utility ties ($84$--$86\%$ of audited pairs in LLM-call accounting) make audits nearly uninformative and halve one-step coverage ($0.61\to0.32$), as \cref{thm:betting}(ii) predicts: ties shrink $\KLinf$ and slow certification.

\section{Related work and what is new}
\label{sec:related}

\emph{Detection.} A mismatched CUSUM with positive drift, our first regime and \eqref{eq:eff}, has the known delay $\ln A/(\theta_0\mu_1)$ \citep{molloy2017misspecified,cooper2024quickest}. We add the zero- and negative-drift regimes, including the finite power law $A^{\theta_1/\theta_0}$ where \citet[Thm.~2]{wu2016quickest} state an infinite delay (counterexample: \cref{app:hebbian}), read deployed counters' implied alternative and set learning rates by it. \emph{Limit.} Ours extends crawler freshness \citep{azar2018tractable}, where one refresh restores a page, to items whose trust needs $\ln(1/\delta)/\KL$ audited uses; the Zipf law is an occupancy sum \citep{karlin1967central} at the load $\Th$. \emph{Certificate.} With standard betting, inverse-propensity weights and anytime validity, B-PAC \citep{yu2026anytime} certifies one global routing threshold; new is a per-template lifecycle that certifies, retires and restarts. \emph{Concurrent work} extends reasoning contraction to self-modification \citep{anonymous2026certification} and shared multi-agent memory \citep{anonymous2026verifiers}. More: \cref{app:related}.

\section{Conclusion}
\label{sec:conclusion}

Reusing a reasoning memory safely is a sequential test, and vote counters (ExpeL, ACE) are change detectors tuned to the wrong change. Three rules follow. (1)~Weigh a memory's failures by its own likelihood ratio: at failure rate $0.05\to0.25$ a failure weighs about $7$ successes, not $1$. (2)~Certify utility against live reasoning; the load $\Th$ prices skipped reasoning in the worst case. (3)~Restart at every notice ($0$ false certifications in $5{,}779$ at-risk epochs at $\delta=0.05$, answers scored alone) and, for substantial edits, invalidate too, with the detector as safety net ($9$--$12\times$ fewer stale answers when notices are missed). Memory benchmarks should edit facts and count stale answers.

\makeatletter\par\typeout{MAINEND p=\thepage\space col=\if@firstcolumn 1\else 2\fi\space tot=\the\pagetotal\space goal=\the\pagegoal}\makeatother
\section*{Limitations}

\paragraph{Scope.} Our results concern memories whose answers an exact verifier checks, such as knowledge-base execution or unit tests. The evidence comes from three domains and five open models of up to 8B parameters: multi-hop reasoning over a knowledge graph with controlled fact edits, a natural timeline of Wikidata temporal qualifiers (four models; \cref{app:natural-edits}), and EvalPlus code tasks, whose memories go stale under requirement revisions and re-enacted Python API changes (\cref{app:codestream}). Each instance--version is decoded once; seed variance is measured on a subset (\cref{app:r3-decode}). The theorems assume that each audit reveals one verified outcome whose law is set by the memory's current reliability, and that facts change at random times; \cref{app:hebbian,app:limit,app:utility} state the exact conditions and \cref{app:experiments} the replay setting. Per-epoch safety needs change notices that cover every change of the certificate's paired-utility null, including changes of live reasoning; our notices cover answer changes (\cref{app:r4-silent}). Some change information is necessary on every template whose epochs contain an audit with probability above $\delta/r_0$ (\cref{prop:app-nonotice-lb}); without an edit log, restarts triggered by the memory's own detector keep every decision that rests on post-change evidence safe (\cref{prop:app-detect}).

\paragraph{Future work.} Five directions extend this scope. (i)~A formation-aware limit: on real streams, coverage also depends on how often the live agent produces a verified trace from which a candidate can form, and a limit that models formation would predict real-system coverage. (ii)~Noisy and LLM-judge verifiers, which would bring text memories into reach; the certification load already prices noise through the observed channel, and a certificate built for noisy paired labels is the next step (\cref{app:focus-noise}). (iii)~Longer and denser natural edit histories, such as full knowledge-base revision logs, where change notices arrive with their own delays and errors, and real library drift on benchmarks with version-specific APIs, where staleness reaches more programs than re-enacted deprecations do. (iv)~Larger models. (v)~Open problems in the theory: a square-root law for adaptive audit budgets, a joint detection--certification limit that prices stale answers, the exact optimal certifier for one template, near-optimal retirement for an unknown post-change rate by parallel or GLR rules, and the agreement-aware version of the limit.

\section*{Ethical considerations}

This work analyzes when stored reasoning may be trusted, and its main practical implication is safety-positive: it bounds how often a memory-augmented reasoner serves stale or spurious shortcuts, and it shows that common counter rules may never retire a stale memory. The datasets we use (KQA Pro, CC BY-SA 4.0; MuSiQue, CC BY 4.0; Wikidata, CC0; EvalPlus, Apache-2.0, with MBPP, CC BY 4.0, and HumanEval, MIT) are public and contain no personal data beyond public facts about notable entities; our synthetic knowledge-base edits, natural-edit timeline and code requirement revisions are released as edit logs for research (under CC BY-SA 4.0 where derived from KQA Pro; our code is MIT-licensed). All questions and tasks are in English or Python. All models are open-weight and were run locally without paid APIs. AI assistants were used to draft proofs, code and text and to run experiments, and independent AI referee agents checked them; the authors take responsibility for all content. The guarantees hold only under the stated assumptions, and deploying certified memory in high-stakes settings still requires a verifier whose errors are understood.

\bibliography{refs}

\begin{thebibliography}{159}
\providecommand{\natexlab}[1]{#1}

\bibitem[{Aggarwal et~al.(2024)Aggarwal, Madaan, Anand, Potharaju, Mishra, Zhou, Gupta, Rajagopal, Kappaganthu, Yang, Upadhyay, Faruqui, and {Mausam}}]{aggarwal2024automix}
Pranjal Aggarwal, Aman Madaan, Ankit Anand, Srividya~Pranavi Potharaju, Swaroop Mishra, Pei Zhou, Aditya Gupta, Dheeraj Rajagopal, Karthik Kappaganthu, Yiming Yang, Shyam Upadhyay, Manaal Faruqui, and {Mausam}. 2024.
\newblock \href {https://arxiv.org/abs/2310.12963} {{AutoMix}: Automatically mixing language models}.
\newblock In \emph{Advances in Neural Information Processing Systems (NeurIPS)}, volume~37.

\bibitem[{Anderson et~al.(2004)Anderson, Bothell, Byrne, Douglass, Lebiere, and Qin}]{anderson2004integrated}
John~R. Anderson, Daniel Bothell, Michael~D. Byrne, Scott Douglass, Christian Lebiere, and Yulin Qin. 2004.
\newblock \href {https://doi.org/10.1037/0033-295X.111.4.1036} {An integrated theory of the mind}.
\newblock \emph{Psychological Review}, 111(4):1036--1060.

\bibitem[{Angelopoulos et~al.(2025)Angelopoulos, Bates, Cand{\`e}s, Jordan, and Lei}]{angelopoulos2025learn}
Anastasios~N. Angelopoulos, Stephen Bates, Emmanuel~J. Cand{\`e}s, Michael~I. Jordan, and Lihua Lei. 2025.
\newblock \href {https://doi.org/10.1214/24-AOAS1998} {{Learn Then Test}: Calibrating predictive algorithms to achieve risk control}.
\newblock \emph{The Annals of Applied Statistics}, 19(2):1641--1662.

\bibitem[{Asadolahi et~al.(2026)Asadolahi, Amini, Talebi, Farhadi, and Zamanifar}]{asadolahi2026memory}
Mohammad Asadolahi, Amir Amini, Samira Talebi, Amirfarhad Farhadi, and Azadeh Zamanifar. 2026.
\newblock \href {https://arxiv.org/abs/2608.00017} {Memory reward inflation in self-improving {LLM} agents}.
\newblock arXiv preprint arXiv:2608.00017.

\bibitem[{Asmussen(2003)}]{asmussen2003applied}
S{\o}ren Asmussen. 2003.
\newblock \href {https://doi.org/10.1007/b97236} {\emph{Applied Probability and Queues}}, 2 edition.
\newblock Stochastic Modelling and Applied Probability. Springer, New York.

\bibitem[{Austin et~al.(2021)Austin, Odena, Nye, Bosma, Michalewski, Dohan, Jiang, Cai, Terry, Le, and Sutton}]{austin2021program}
Jacob Austin, Augustus Odena, Maxwell Nye, Maarten Bosma, Henryk Michalewski, David Dohan, Ellen Jiang, Carrie Cai, Michael Terry, Quoc Le, and Charles Sutton. 2021.
\newblock Program synthesis with large language models.
\newblock \emph{arXiv preprint arXiv:2108.07732}.

\bibitem[{Azar et~al.(2018)Azar, Horvitz, Lubetzky, Peres, and Shahaf}]{azar2018tractable}
Yossi Azar, Eric Horvitz, Eyal Lubetzky, Yuval Peres, and Dafna Shahaf. 2018.
\newblock \href {https://doi.org/10.1073/pnas.1801519115} {Tractable near-optimal policies for crawling}.
\newblock \emph{Proceedings of the National Academy of Sciences}, 115(32):8099--8103.

\bibitem[{Banerjee and Veeravalli(2013)}]{banerjee2013data}
Taposh Banerjee and Venugopal~V. Veeravalli. 2013.
\newblock \href {https://doi.org/10.1109/TIT.2013.2272313} {Data-efficient quickest change detection in minimax settings}.
\newblock \emph{IEEE Transactions on Information Theory}, 59(10):6917--6931.

\bibitem[{Bang(2023)}]{bang2023gptcache}
Fu~Bang. 2023.
\newblock \href {https://doi.org/10.18653/v1/2023.nlposs-1.24} {{GPTC}ache: An open-source semantic cache for {LLM} applications enabling faster answers and cost savings}.
\newblock In \emph{Proceedings of the 3rd Workshop for Natural Language Processing Open Source Software (NLP-OSS 2023)}, pages 212--218. Association for Computational Linguistics.

\bibitem[{Bastopcu and Ulukus(2021)}]{bastopcu2021information}
Melih Bastopcu and Sennur Ulukus. 2021.
\newblock \href {https://doi.org/10.1109/TWC.2020.3037144} {Information freshness in cache updating systems}.
\newblock \emph{IEEE Transactions on Wireless Communications}, 20(3):1861--1874.

\bibitem[{Brook and Evans(1972)}]{brook1972approach}
D.~Brook and D.~A. Evans. 1972.
\newblock \href {https://doi.org/10.1093/biomet/59.3.539} {An approach to the probability distribution of {CUSUM} run length}.
\newblock \emph{Biometrika}, 59(3):539--549.

\bibitem[{Cai et~al.(2026)Cai, Zhang, Nie, Ran, Zheng, Song, Tian, Guo, and Xue}]{cai2026safeharness}
Qianshu Cai, Yonggang Zhang, Jun Nie, Maohao Ran, Huajiang Zheng, Jun Song, Xinmei Tian, Yike Guo, and Wei Xue. 2026.
\newblock \href {https://arxiv.org/abs/2609.08175} {A theory of reliable self-evolution for agent harnesses}.
\newblock arXiv preprint arXiv:2609.08175.
\newblock Version 1 (8 Sep 2026) titled ``Safe Harness Self-Evolution: A Theoretical Analysis of Feasibility and Limits''.

\bibitem[{Cai et~al.(2024)Cai, Wang, Ma, Chen, and Zhou}]{cai2024large}
Tianle Cai, Xuezhi Wang, Tengyu Ma, Xinyun Chen, and Denny Zhou. 2024.
\newblock Large language models as tool makers.
\newblock In \emph{International Conference on Learning Representations}.
\newblock ArXiv:2305.17126.

\bibitem[{Cao et~al.(2022)Cao, Shi, Pan, Nie, Xiang, Hou, Li, He, and Zhang}]{cao2022kqa}
Shulin Cao, Jiaxin Shi, Liangming Pan, Lunyiu Nie, Yutong Xiang, Lei Hou, Juanzi Li, Bin He, and Hanwang Zhang. 2022.
\newblock \href {https://doi.org/10.18653/v1/2022.acl-long.422} {{KQA Pro}: A dataset with explicit compositional programs for complex question answering over knowledge base}.
\newblock In \emph{Proceedings of the 60th Annual Meeting of the Association for Computational Linguistics (Volume 1: Long Papers)}, pages 6101--6119, Dublin, Ireland. Association for Computational Linguistics.

\bibitem[{Chandak et~al.(2020)Chandak, Jordan, Theocharous, White, and Thomas}]{chandak2020nonstationary}
Yash Chandak, Scott~M. Jordan, Georgios Theocharous, Martha White, and Philip~S. Thomas. 2020.
\newblock \href {https://arxiv.org/abs/2010.12645} {Towards safe policy improvement for non-stationary {MDP}s}.
\newblock In \emph{Advances in Neural Information Processing Systems (NeurIPS)}, volume~33.

\bibitem[{Che et~al.(2002)Che, Tung, and Wang}]{che2002hierarchical}
Hao Che, Ye~Tung, and Zhijun Wang. 2002.
\newblock \href {https://doi.org/10.1109/JSAC.2002.801752} {Hierarchical {Web} caching systems: Modeling, design and experimental results}.
\newblock \emph{IEEE Journal on Selected Areas in Communications}, 20(7):1305--1314.

\bibitem[{Chen et~al.(2024{\natexlab{a}})Chen, Zaharia, and Zou}]{chen2024frugalgpt}
Lingjiao Chen, Matei Zaharia, and James Zou. 2024{\natexlab{a}}.
\newblock \href {https://arxiv.org/abs/2305.05176} {{FrugalGPT}: How to use large language models while reducing cost and improving performance}.
\newblock \emph{Transactions on Machine Learning Research}.

\bibitem[{Chen et~al.(2024{\natexlab{b}})Chen, Zaharia, and Zou}]{chen2024how}
Lingjiao Chen, Matei Zaharia, and James Zou. 2024{\natexlab{b}}.
\newblock \href {https://doi.org/10.1162/99608f92.5317da47} {How is {ChatGPT}'s behavior changing over time?}
\newblock \emph{Harvard Data Science Review}, 6(2).

\bibitem[{Chen et~al.(2021)Chen, Tworek, Jun, Yuan, Pinto, Kaplan, Edwards, Burda, Joseph, Brockman et~al.}]{chen2021evaluating}
Mark Chen, Jerry Tworek, Heewoo Jun, Qiming Yuan, Henrique Ponde de~Oliveira Pinto, Jared Kaplan, Harri Edwards, Yuri Burda, Nicholas Joseph, Greg Brockman, et~al. 2021.
\newblock Evaluating large language models trained on code.
\newblock \emph{arXiv preprint arXiv:2107.03374}.

\bibitem[{Chen et~al.(2026)Chen, Bellucci, Sun, and Jacucci}]{chen2026skilldroid}
Qijia Chen, Andrea Bellucci, Zhida Sun, and Giulio Jacucci. 2026.
\newblock \href {https://arxiv.org/abs/2604.14872} {{SkillDroid}: Compile once, reuse forever}.
\newblock arXiv preprint arXiv:2604.14872.

\bibitem[{Chhikara et~al.(2025)Chhikara, Khant, Aryan, Singh, and Yadav}]{chhikara2025mem0}
Prateek Chhikara, Dev Khant, Saket Aryan, Taranjeet Singh, and Deshraj Yadav. 2025.
\newblock \href {https://arxiv.org/abs/2504.19413} {{Mem0}: Building production-ready {AI} agents with scalable long-term memory}.
\newblock arXiv preprint arXiv:2504.19413.

\bibitem[{Cho and Garcia-Molina(2003)}]{cho2003effective}
Junghoo Cho and Hector Garcia-Molina. 2003.
\newblock \href {https://doi.org/10.1145/958942.958945} {Effective page refresh policies for {Web} crawlers}.
\newblock \emph{ACM Transactions on Database Systems}, 28(4):390--426.

\bibitem[{Choungmo~Fofack et~al.(2014)Choungmo~Fofack, Nain, Neglia, and Towsley}]{fofack2014performance}
Nicaise Choungmo~Fofack, Philippe Nain, Giovanni Neglia, and Don Towsley. 2014.
\newblock \href {https://doi.org/10.1016/j.comnet.2014.03.006} {Performance evaluation of hierarchical {TTL}-based cache networks}.
\newblock \emph{Computer Networks}, 65:212--231.

\bibitem[{Cohen and Shenker(2002)}]{cohen2002replication}
Edith Cohen and Scott Shenker. 2002.
\newblock \href {https://doi.org/10.1145/633025.633043} {Replication strategies in unstructured peer-to-peer networks}.
\newblock In \emph{Proceedings of the 2002 Conference on Applications, Technologies, Architectures, and Protocols for Computer Communications (SIGCOMM)}, pages 177--190. ACM.

\bibitem[{Cooper and Meyn(2024)}]{cooper2024quickest}
Austin Cooper and Sean Meyn. 2024.
\newblock \href {https://doi.org/10.1109/Allerton63246.2024.10735265} {Quickest change detection using mismatched {CUSUM} (extended abstract)}.
\newblock In \emph{Proceedings of the 60th Annual Allerton Conference on Communication, Control, and Computing}, pages 1--7. IEEE.
\newblock Extended version: arXiv:2409.07948.

\bibitem[{Didolkar et~al.(2025)Didolkar, Ballas, Arora, and Goyal}]{didolkar2025metacognitive}
Aniket Didolkar, Nicolas Ballas, Sanjeev Arora, and Anirudh Goyal. 2025.
\newblock \href {https://arxiv.org/abs/2509.13237} {Metacognitive reuse: Turning recurring {LLM} reasoning into concise behaviors}.
\newblock arXiv preprint arXiv:2509.13237.

\bibitem[{Dong et~al.(2025)Dong, Xu, He, Li, Tang, Liu, Liu, and Xiang}]{dong2025minja}
Shen Dong, Shaochen Xu, Pengfei He, Yige Li, Jiliang Tang, Tianming Liu, Hui Liu, and Zhen Xiang. 2025.
\newblock \href {https://arxiv.org/abs/2503.03704} {Memory injection attacks on {LLM} agents via query-only interaction}.
\newblock arXiv preprint arXiv:2503.03704.

\bibitem[{Ellis et~al.(2021)Ellis, Wong, Nye, Sabl{\'e}-Meyer, Morales, Hewitt, Cary, Solar-Lezama, and Tenenbaum}]{ellis2021dreamcoder}
Kevin Ellis, Catherine Wong, Maxwell Nye, Mathias Sabl{\'e}-Meyer, Lucas Morales, Luke Hewitt, Luc Cary, Armando Solar-Lezama, and Joshua~B. Tenenbaum. 2021.
\newblock \href {https://doi.org/10.1145/3453483.3454080} {{DreamCoder}: Bootstrapping inductive program synthesis with wake-sleep library learning}.
\newblock In \emph{Proceedings of the 42nd ACM SIGPLAN International Conference on Programming Language Design and Implementation (PLDI)}, pages 835--850. ACM.

\bibitem[{Fagin(1977)}]{fagin1977asymptotic}
Ronald Fagin. 1977.
\newblock \href {https://doi.org/10.1016/S0022-0000(77)80014-7} {Asymptotic miss ratios over independent references}.
\newblock \emph{Journal of Computer and System Sciences}, 14(2):222--250.

\bibitem[{Feng et~al.(2021)Feng, Emerson, and Simon}]{feng2021biocreep}
Jean Feng, Scott Emerson, and Noah Simon. 2021.
\newblock \href {https://doi.org/10.1111/biom.13379} {Approval policies for modifications to machine learning-based software as a medical device: A study of bio-creep}.
\newblock \emph{Biometrics}, 77(1):31--44.

\bibitem[{Fikes et~al.(1972)Fikes, Hart, and Nilsson}]{fikes1972learning}
Richard~E. Fikes, Peter~E. Hart, and Nils~J. Nilsson. 1972.
\newblock \href {https://doi.org/10.1016/0004-3702(72)90051-3} {Learning and executing generalized robot plans}.
\newblock \emph{Artificial Intelligence}, 3:251--288.

\bibitem[{Forouzandeh et~al.(2026)Forouzandeh, Peng, Moradi, Yu, and Jalili}]{forouzandeh2026learning}
Saman Forouzandeh, Wei Peng, Parham Moradi, Xinghuo Yu, and Mahdi Jalili. 2026.
\newblock \href {https://arxiv.org/abs/2512.18950} {Learning hierarchical procedural memory for {LLM} agents through {B}ayesian selection and contrastive refinement}.
\newblock In \emph{Proceedings of the 25th International Conference on Autonomous Agents and Multiagent Systems (AAMAS)}.

\bibitem[{Gallo et~al.(2014)Gallo, Kauffmann, Muscariello, Simonian, and Tanguy}]{gallo2014performance}
Massimo Gallo, Bruno Kauffmann, Luca Muscariello, Alain Simonian, and Christian Tanguy. 2014.
\newblock \href {https://doi.org/10.1016/j.peva.2013.10.004} {Performance evaluation of the random replacement policy for networks of caches}.
\newblock \emph{Performance Evaluation}, 72:16--36.

\bibitem[{Gao et~al.(2026)Gao, Chen, Wang, Guo, Pang, Liu, Shen, and Cheng}]{gao2026skillaudit}
Haowen Gao, Haoran Chen, Can Wang, Shasha Guo, Liang Pang, Zhaoyang Liu, Huawei Shen, and Xueqi Cheng. 2026.
\newblock \href {https://arxiv.org/abs/2606.14239} {{SelfEvoSkill}: Paired execution audits for skill revision without external outcome supervision}.
\newblock arXiv preprint arXiv:2606.14239.
\newblock Version 1 (12 Jun 2026) titled ``SkillAudit: Ground-Truth-Free Skill Evolution via Paired Trajectory Auditing''.

\bibitem[{Garivier and Kaufmann(2016)}]{garivier2016optimal}
Aur{\'e}lien Garivier and Emilie Kaufmann. 2016.
\newblock \href {https://proceedings.mlr.press/v49/garivier16a.html} {Optimal best arm identification with fixed confidence}.
\newblock In \emph{Proceedings of the 29th Annual Conference on Learning Theory (COLT)}, volume~49 of \emph{Proceedings of Machine Learning Research}, pages 998--1027. PMLR.

\bibitem[{Gnedin et~al.(2007)Gnedin, Hansen, and Pitman}]{gnedin2007notes}
Alexander Gnedin, Ben Hansen, and Jim Pitman. 2007.
\newblock \href {https://doi.org/10.1214/07-PS092} {Notes on the occupancy problem with infinitely many boxes: General asymptotics and power laws}.
\newblock \emph{Probability Surveys}, 4:146--171.

\bibitem[{Good(1953)}]{good1953population}
I.~J. Good. 1953.
\newblock \href {https://doi.org/10.1093/biomet/40.3-4.237} {The population frequencies of species and the estimation of population parameters}.
\newblock \emph{Biometrika}, 40(3--4):237--264.

\bibitem[{Good and Toulmin(1956)}]{good1956number}
I.~J. Good and G.~H. Toulmin. 1956.
\newblock \href {https://doi.org/10.1093/biomet/43.1-2.45} {The number of new species, and the increase in population coverage, when a sample is increased}.
\newblock \emph{Biometrika}, 43(1--2):45--63.

\bibitem[{Gratch and DeJong(1992)}]{gratch1992composer}
Jonathan Gratch and Gerald DeJong. 1992.
\newblock {COMPOSER}: A probabilistic solution to the utility problem in speed-up learning.
\newblock In \emph{Proceedings of the Tenth National Conference on Artificial Intelligence (AAAI)}, pages 235--240. AAAI Press.

\bibitem[{Greiner(1996)}]{greiner1996palo}
Russell Greiner. 1996.
\newblock \href {https://doi.org/10.1016/0004-3702(95)00040-2} {{PALO}: A probabilistic hill-climbing algorithm}.
\newblock \emph{Artificial Intelligence}, 84(1--2):177--208.

\bibitem[{Hall and Heyde(1980)}]{hall1980martingale}
Peter Hall and Christopher~C. Heyde. 1980.
\newblock \emph{Martingale Limit Theory and Its Application}.
\newblock Academic Press, New York.

\bibitem[{Hao et~al.(2025)Hao, Sukhbaatar, Su, Li, Hu, Weston, and Tian}]{hao2025training}
Shibo Hao, Sainbayar Sukhbaatar, DiJia Su, Xian Li, Zhiting Hu, Jason Weston, and Yuandong Tian. 2025.
\newblock \href {https://arxiv.org/abs/2412.06769} {Training large language models to reason in a continuous latent space}.
\newblock In \emph{Conference on Language Modeling (COLM)}.

\bibitem[{Heathcote et~al.(2000)Heathcote, Brown, and Mewhort}]{heathcote2000power}
Andrew Heathcote, Scott Brown, and D.~J.~K. Mewhort. 2000.
\newblock \href {https://doi.org/10.3758/BF03212979} {The power law repealed: The case for an exponential law of practice}.
\newblock \emph{Psychonomic Bulletin \& Review}, 7(2):185--207.

\bibitem[{Honda and Takemura(2010)}]{honda2010asymptotically}
Junya Honda and Akimichi Takemura. 2010.
\newblock An asymptotically optimal bandit algorithm for bounded support models.
\newblock In \emph{Proceedings of the 23rd Conference on Learning Theory (COLT)}, pages 67--79.

\bibitem[{Howard et~al.(2021)Howard, Ramdas, McAuliffe, and Sekhon}]{howard2021time}
Steven~R. Howard, Aaditya Ramdas, Jon McAuliffe, and Jasjeet Sekhon. 2021.
\newblock \href {https://doi.org/10.1214/20-AOS1991} {Time-uniform, nonparametric, nonasymptotic confidence sequences}.
\newblock \emph{The Annals of Statistics}, 49(2):1055--1080.

\bibitem[{Huang et~al.(2026)Huang, Zeng, Jing, Wei, and An}]{huang2026conditional}
Jianguo Huang, Hao Zeng, Bingyi Jing, Hongxin Wei, and Bo~An. 2026.
\newblock \href {https://arxiv.org/abs/2601.22790} {Conditional performance guarantee for large reasoning models}.
\newblock arXiv preprint arXiv:2601.22790.

\bibitem[{Hutter(2021)}]{hutter2021learning}
Marcus Hutter. 2021.
\newblock \href {https://arxiv.org/abs/2102.04074} {Learning curve theory}.
\newblock arXiv preprint arXiv:2102.04074.

\bibitem[{{IBM Granite Team}(2025)}]{ibm2025granite}
{IBM Granite Team}. 2025.
\newblock \href {https://huggingface.co/ibm-granite/granite-3.3-8b-instruct} {Granite 3.3 {8B} instruct}.
\newblock Model card.

\bibitem[{Iglehart(1972)}]{iglehart1972extreme}
Donald~L. Iglehart. 1972.
\newblock \href {https://doi.org/10.1214/aoms/1177692642} {Extreme values in the {GI/G/1} queue}.
\newblock \emph{The Annals of Mathematical Statistics}, 43(2):627--635.

\bibitem[{Jelenkovi{\'c}(1999)}]{jelenkovic1999asymptotic}
Predrag~R. Jelenkovi{\'c}. 1999.
\newblock \href {https://doi.org/10.1214/aoap/1029962750} {Asymptotic approximation of the move-to-front search cost distribution and least-recently used caching fault probabilities}.
\newblock \emph{The Annals of Applied Probability}, 9(2):430--464.

\bibitem[{Jiang et~al.(2023)Jiang, Sablayrolles, Mensch, Bamford, Chaplot, de~las Casas, Bressand, Lengyel, Lample, Saulnier, Lavaud, Lachaux, Stock, Le~Scao, Lavril, Wang, Lacroix, and El~Sayed}]{jiang2023mistral}
Albert~Q. Jiang, Alexandre Sablayrolles, Arthur Mensch, Chris Bamford, Devendra~Singh Chaplot, Diego de~las Casas, Florian Bressand, Gianna Lengyel, Guillaume Lample, Lucile Saulnier, L{\'e}lio~Renard Lavaud, Marie-Anne Lachaux, Pierre Stock, Teven Le~Scao, Thibaut Lavril, Thomas Wang, Timoth{\'e}e Lacroix, and William El~Sayed. 2023.
\newblock \href {https://arxiv.org/abs/2310.06825} {Mistral 7{B}}.
\newblock \emph{Preprint}, arXiv:2310.06825.

\bibitem[{Karlin(1967)}]{karlin1967central}
Samuel Karlin. 1967.
\newblock \href {https://doi.org/10.1512/iumj.1968.17.17020} {Central limit theorems for certain infinite urn schemes}.
\newblock \emph{Journal of Mathematics and Mechanics}, 17(4):373--401.

\bibitem[{Kasai et~al.(2023)Kasai, Sakaguchi, Takahashi, Bras, Asai, Yu, Radev, Smith, Choi, and Inui}]{kasai2023realtime}
Jungo Kasai, Keisuke Sakaguchi, Yoichi Takahashi, Ronan~Le Bras, Akari Asai, Xinyan Yu, Dragomir Radev, Noah~A. Smith, Yejin Choi, and Kentaro Inui. 2023.
\newblock {RealTime QA}: What's the answer right now?
\newblock In \emph{Advances in Neural Information Processing Systems (NeurIPS), Datasets and Benchmarks Track}, volume~36.

\bibitem[{Kaufmann et~al.(2016)Kaufmann, Capp{\'e}, and Garivier}]{kaufmann2016complexity}
Emilie Kaufmann, Olivier Capp{\'e}, and Aur{\'e}lien Garivier. 2016.
\newblock \href {https://jmlr.org/papers/v17/kaufman16a.html} {On the complexity of best-arm identification in multi-armed bandit models}.
\newblock \emph{Journal of Machine Learning Research}, 17(1):1--42.

\bibitem[{Khosravi and Huo(2026)}]{khosravi2026conformal}
Hamed Khosravi and Xiaoming Huo. 2026.
\newblock \href {https://arxiv.org/abs/2605.20270} {Conformal selective acting: Anytime-valid risk control for {RLVR}-trained {LLMs}}.
\newblock arXiv preprint arXiv:2605.20270.

\bibitem[{Kim et~al.(2025)Kim, Garg, Peng, and Garg}]{kim2025correlated}
Elliot Kim, Avi Garg, Kenny Peng, and Nikhil Garg. 2025.
\newblock \href {https://arxiv.org/abs/2506.07962} {Correlated errors in large language models}.
\newblock In \emph{Proceedings of the 42nd International Conference on Machine Learning (ICML)}, volume 267 of \emph{Proceedings of Machine Learning Research}, pages 30038--30066. PMLR.

\bibitem[{Kolobov et~al.(2019)Kolobov, Peres, Lu, and Horvitz}]{kolobov2019staying}
Andrey Kolobov, Yuval Peres, Cheng Lu, and Eric~J. Horvitz. 2019.
\newblock \href {https://proceedings.neurips.cc/paper_files/paper/2019/file/ad13a2a07ca4b7642959dc0c4c740ab6-Paper.pdf} {Staying up to date with online content changes using reinforcement learning for scheduling}.
\newblock In \emph{Advances in Neural Information Processing Systems (NeurIPS)}, volume~32.

\bibitem[{Korf(1985)}]{korf1985macro}
Richard~E. Korf. 1985.
\newblock \href {https://doi.org/10.1016/0004-3702(85)90012-8} {Macro-operators: A weak method for learning}.
\newblock \emph{Artificial Intelligence}, 26(1):35--77.

\bibitem[{Kwon et~al.(2023)Kwon, Li, Zhuang, Sheng, Zheng, Yu, Gonzalez, Zhang, and Stoica}]{kwon2023efficient}
Woosuk Kwon, Zhuohan Li, Siyuan Zhuang, Ying Sheng, Lianmin Zheng, Cody~Hao Yu, Joseph~E. Gonzalez, Hao Zhang, and Ion Stoica. 2023.
\newblock \href {https://doi.org/10.1145/3600006.3613165} {Efficient memory management for large language model serving with {PagedAttention}}.
\newblock In \emph{Proceedings of the 29th Symposium on Operating Systems Principles (SOSP)}, pages 611--626.

\bibitem[{Lai(1981)}]{lai1981asymptotic}
Tze~Leung Lai. 1981.
\newblock \href {https://doi.org/10.1214/aos/1176345398} {Asymptotic optimality of invariant sequential probability ratio tests}.
\newblock \emph{The Annals of Statistics}, 9(2):318--333.

\bibitem[{Lai(1998)}]{lai1998information}
Tze~Leung Lai. 1998.
\newblock \href {https://doi.org/10.1109/18.737522} {Information bounds and quick detection of parameter changes in stochastic systems}.
\newblock \emph{IEEE Transactions on Information Theory}, 44(7):2917--2929.

\bibitem[{Laird et~al.(1986)Laird, Rosenbloom, and Newell}]{laird1986chunking}
John~E. Laird, Paul~S. Rosenbloom, and Allen Newell. 1986.
\newblock \href {https://doi.org/10.1007/BF00116249} {Chunking in {Soar}: The anatomy of a general learning mechanism}.
\newblock \emph{Machine Learning}, 1(1):11--46.

\bibitem[{Lee et~al.(2025)Lee, Qi, Neeley, Liu, and Jeong}]{lee2025consol}
Jaeyeon Lee, Guantong Qi, Matthew~Brady Neeley, Zhandong Liu, and Hyun-Hwan Jeong. 2025.
\newblock \href {https://arxiv.org/abs/2503.17587} {{ConSol}: Sequential probability ratio testing to find consistent {LLM} reasoning paths efficiently}.
\newblock arXiv preprint arXiv:2503.17587.

\bibitem[{Li et~al.(2025)Li, Xi, Li, Chen, Chen, Song, Niu, Wang, Yang, Tang, Yu, Zhao, Wang, Liu, Lin, Wang, Huo, Chen, Chen, Li, Tao, Lai, Wu, Tang, Wang, Fan, Zhang, Zhang, Yan, Yang, Xu, Xu, Chen, Wang, Yang, Zhang, Xu, Chen, and Xiong}]{li2025memos}
Zhiyu Li, Chenyang Xi, Chunyu Li, Ding Chen, Boyu Chen, Shichao Song, Simin Niu, Hanyu Wang, Jiawei Yang, Chen Tang, Qingchen Yu, Jihao Zhao, Yezhaohui Wang, Peng Liu, Zehao Lin, Pengyuan Wang, Jiahao Huo, Tianyi Chen, Kai Chen, and 20 others. 2025.
\newblock \href {https://arxiv.org/abs/2507.03724} {{MemOS}: A memory {OS} for {AI} system}.
\newblock arXiv preprint arXiv:2507.03724.

\bibitem[{Li{\v{s}}ka et~al.(2022)Li{\v{s}}ka, Ko{\v{c}}isk{\'y}, Gribovskaya, Terzi, Sezener, Agrawal, de~Masson~d'Autume, Scholtes, Zaheer, Young, Gilsenan-McMahon, Austin, Blunsom, and Lazaridou}]{liska2022streamingqa}
Adam Li{\v{s}}ka, Tom{\'a}{\v{s}} Ko{\v{c}}isk{\'y}, Elena Gribovskaya, Tayfun Terzi, Eren Sezener, Devang Agrawal, Cyprien de~Masson~d'Autume, Tim Scholtes, Manzil Zaheer, Susannah Young, Ellen Gilsenan-McMahon, Sophia Austin, Phil Blunsom, and Angeliki Lazaridou. 2022.
\newblock {StreamingQA}: A benchmark for adaptation to new knowledge over time in question answering models.
\newblock In \emph{Proceedings of the 39th International Conference on Machine Learning (ICML)}, volume 162 of \emph{Proceedings of Machine Learning Research}, pages 13604--13622.

\bibitem[{Liu et~al.(2023)Liu, Xia, Wang, and Zhang}]{liu2023evalplus}
Jiawei Liu, Chunqiu~Steven Xia, Yuyao Wang, and Lingming Zhang. 2023.
\newblock Is your code generated by {ChatGPT} really correct? {R}igorous evaluation of large language models for code generation.
\newblock In \emph{Advances in Neural Information Processing Systems}.
\newblock ArXiv:2305.01210.

\bibitem[{Liu et~al.(2024)Liu, Guo, Hu, Jiayang, Zhang, Qiu, and Zhang}]{liu2024can}
Tengxiao Liu, Qipeng Guo, Xiangkun Hu, Cheng Jiayang, Yue Zhang, Xipeng Qiu, and Zheng Zhang. 2024.
\newblock \href {https://arxiv.org/abs/2411.01855} {Can language models learn to skip steps?}
\newblock In \emph{Advances in Neural Information Processing Systems (NeurIPS)}, volume~37.

\bibitem[{Liu et~al.(2026)Liu, Atalar, Dai, Zuo, Wang, Lui, Chen, and Joe-Wong}]{liu2026semantic}
Xutong Liu, Baran Atalar, Xiangxiang Dai, Jinhang Zuo, Siwei Wang, John C.~S. Lui, Wei Chen, and Carlee Joe-Wong. 2026.
\newblock \href {https://arxiv.org/abs/2508.07675} {Semantic caching for low-cost {LLM} serving: From offline learning to online adaptation}.
\newblock In \emph{Proceedings of IEEE INFOCOM 2026 -- IEEE Conference on Computer Communications}.

\bibitem[{Logan(1988)}]{logan1988toward}
Gordon~D. Logan. 1988.
\newblock \href {https://doi.org/10.1037/0033-295X.95.4.492} {Toward an instance theory of automatization}.
\newblock \emph{Psychological Review}, 95(4):492--527.

\bibitem[{Lorden(1971)}]{lorden1971procedures}
G.~Lorden. 1971.
\newblock \href {https://doi.org/10.1214/aoms/1177693055} {Procedures for reacting to a change in distribution}.
\newblock \emph{The Annals of Mathematical Statistics}, 42(6):1897--1908.

\bibitem[{Madaan et~al.(2023)Madaan, Tandon, Gupta, Hallinan, Gao, Wiegreffe, Alon, Dziri, Prabhumoye, Yang, Gupta, Majumder, Hermann, Welleck, Yazdanbakhsh, and Clark}]{madaan2023self}
Aman Madaan, Niket Tandon, Prakhar Gupta, Skyler Hallinan, Luyu Gao, Sarah Wiegreffe, Uri Alon, Nouha Dziri, Shrimai Prabhumoye, Yiming Yang, Shashank Gupta, Bodhisattwa~Prasad Majumder, Katherine Hermann, Sean Welleck, Amir Yazdanbakhsh, and Peter Clark. 2023.
\newblock Self-refine: Iterative refinement with self-feedback.
\newblock In \emph{Advances in Neural Information Processing Systems}.
\newblock ArXiv:2303.17651.

\bibitem[{Mansoor et~al.(2026)Mansoor, Ahmad, and Yoon}]{mansoor2026risk}
Muhammad Mansoor, Tahir Ahmad, and Yeo-Chan Yoon. 2026.
\newblock \href {https://arxiv.org/abs/2607.04281} {Risk-constrained freshness-aware semantic caching for open-web retrieval-augmented {LLMs}}.
\newblock arXiv preprint arXiv:2607.04281.

\bibitem[{Martina et~al.(2014)Martina, Garetto, and Leonardi}]{martina2014unified}
Valentina Martina, Michele Garetto, and Emilio Leonardi. 2014.
\newblock \href {https://doi.org/10.1109/INFOCOM.2014.6848145} {A unified approach to the performance analysis of caching systems}.
\newblock In \emph{Proceedings of IEEE INFOCOM 2014 -- IEEE Conference on Computer Communications}, pages 2040--2048.

\bibitem[{McAllester and Schapire(2000)}]{mcallester2000convergence}
David~A. McAllester and Robert~E. Schapire. 2000.
\newblock \href {https://www.learningtheory.org/colt2000/papers/McAllesterSchapire.pdf} {On the convergence rate of {Good-Turing} estimators}.
\newblock In \emph{Proceedings of the Thirteenth Annual Conference on Computational Learning Theory (COLT)}, pages 1--6.

\bibitem[{Michaud et~al.(2023)Michaud, Liu, Girit, and Tegmark}]{michaud2023quantization}
Eric~J. Michaud, Ziming Liu, Uzay Girit, and Max Tegmark. 2023.
\newblock \href {https://doi.org/10.52202/075280-1248} {The quantization model of neural scaling}.
\newblock In \emph{Advances in Neural Information Processing Systems (NeurIPS)}, volume~36, pages 28699--28722.

\bibitem[{Minton(1990)}]{minton1990quantitative}
Steven Minton. 1990.
\newblock \href {https://doi.org/10.1016/0004-3702(90)90059-9} {Quantitative results concerning the utility of explanation-based learning}.
\newblock \emph{Artificial Intelligence}, 42(2--3):363--391.

\bibitem[{Mitchell et~al.(1986)Mitchell, Keller, and Kedar-Cabelli}]{mitchell1986explanation}
Tom~M. Mitchell, Richard~M. Keller, and Smadar~T. Kedar-Cabelli. 1986.
\newblock \href {https://doi.org/10.1007/BF00116250} {Explanation-based generalization: A unifying view}.
\newblock \emph{Machine Learning}, 1(1):47--80.

\bibitem[{Mo et~al.(2026)Mo, Zhao, and Geng}]{mo2026stable}
Yunxiang Mo, Donghao Zhao, and Hejia Geng. 2026.
\newblock \href {https://arxiv.org/abs/2609.09989} {Stable answers, unfinished reasoning: Why self-consensus is not a safe early-exit signal}.
\newblock arXiv preprint arXiv:2609.09989.

\bibitem[{Molloy(2021)}]{molloy2021misspecified}
Timothy~L. Molloy. 2021.
\newblock \href {https://doi.org/10.1109/TAC.2020.2985975} {Misspecified and asymptotically minimax robust quickest change diagnosis}.
\newblock \emph{IEEE Transactions on Automatic Control}, 66(2):857--864.

\bibitem[{Molloy and Ford(2017)}]{molloy2017misspecified}
Timothy~L. Molloy and Jason~J. Ford. 2017.
\newblock \href {https://doi.org/10.1109/TSP.2017.2740202} {Misspecified and asymptotically minimax robust quickest change detection}.
\newblock \emph{IEEE Transactions on Signal Processing}, 65(21):5730--5742.

\bibitem[{Moustakides(1986)}]{moustakides1986optimal}
George~V. Moustakides. 1986.
\newblock \href {https://doi.org/10.1214/aos/1176350164} {Optimal stopping times for detecting changes in distributions}.
\newblock \emph{The Annals of Statistics}, 14(4):1379--1387.

\bibitem[{Munford(1980)}]{munford1980control}
A.~G. Munford. 1980.
\newblock \href {https://doi.org/10.2307/2346899} {A control chart based on cumulative scores}.
\newblock \emph{Journal of the Royal Statistical Society. Series C (Applied Statistics)}, 29(3):252--258.

\bibitem[{Newell and Rosenbloom(1981)}]{newell1981mechanisms}
Allen Newell and Paul~S. Rosenbloom. 1981.
\newblock Mechanisms of skill acquisition and the law of practice.
\newblock In John~R. Anderson, editor, \emph{Cognitive Skills and Their Acquisition}, pages 1--55. Lawrence Erlbaum Associates, Hillsdale, NJ.

\bibitem[{Ohannessian and Dahleh(2012)}]{ohannessian2012rare}
Mesrob~I. Ohannessian and Munther~A. Dahleh. 2012.
\newblock \href {https://proceedings.mlr.press/v23/ohannessian12.html} {Rare probability estimation under regularly varying heavy tails}.
\newblock In \emph{Proceedings of the 25th Annual Conference on Learning Theory (COLT)}, volume~23 of \emph{Proceedings of Machine Learning Research}, pages 21.1--21.24. PMLR.

\bibitem[{Orlitsky et~al.(2016)Orlitsky, Suresh, and Wu}]{orlitsky2016optimal}
Alon Orlitsky, Ananda~Theertha Suresh, and Yihong Wu. 2016.
\newblock \href {https://doi.org/10.1073/pnas.1607774113} {Optimal prediction of the number of unseen species}.
\newblock \emph{Proceedings of the National Academy of Sciences}, 113(47):13283--13288.

\bibitem[{Ouyang et~al.(2026)Ouyang, Yan, Hsu, Chen, Jiang, Wang, Han, Le, Daruki, Tang, Tirumalashetty, Lee, Rofouei, Lin, Han, Lee, and Pfister}]{ouyang2026reasoningbank}
Siru Ouyang, Jun Yan, I-Hung Hsu, Yanfei Chen, Ke~Jiang, Zifeng Wang, Rujun Han, Long~T. Le, Samira Daruki, Xiangru Tang, Vishy Tirumalashetty, George Lee, Mahsan Rofouei, Hangfei Lin, Jiawei Han, Chen-Yu Lee, and Tomas Pfister. 2026.
\newblock \href {https://arxiv.org/abs/2509.25140} {{ReasoningBank}: Scaling agent self-evolving with reasoning memory}.
\newblock In \emph{International Conference on Learning Representations (ICLR)}.

\bibitem[{Page(1954)}]{page1954continuous}
E.~S. Page. 1954.
\newblock \href {https://doi.org/10.1093/biomet/41.1-2.100} {Continuous inspection schemes}.
\newblock \emph{Biometrika}, 41(1--2):100--115.

\bibitem[{Park et~al.(2023)Park, O'Brien, Cai, Morris, Liang, and Bernstein}]{park2023generative}
Joon~Sung Park, Joseph~C. O'Brien, Carrie~J. Cai, Meredith~Ringel Morris, Percy Liang, and Michael~S. Bernstein. 2023.
\newblock \href {https://doi.org/10.1145/3586183.3606763} {Generative agents: Interactive simulacra of human behavior}.
\newblock In \emph{Proceedings of the 36th Annual ACM Symposium on User Interface Software and Technology (UIST)}. ACM.

\bibitem[{Podkopaev and Ramdas(2022)}]{podkopaev2022tracking}
Aleksandr Podkopaev and Aaditya Ramdas. 2022.
\newblock \href {https://arxiv.org/abs/2110.06177} {Tracking the risk of a deployed model and detecting harmful distribution shifts}.
\newblock In \emph{International Conference on Learning Representations (ICLR)}.

\bibitem[{Pu et~al.(2026)Pu, Song, and Zhao}]{pu2026skillops}
Hongji Pu, Xinyuan Song, and Liang Zhao. 2026.
\newblock \href {https://arxiv.org/abs/2605.13716} {{SkillOps}: Managing {LLM} agent skill libraries as self-maintaining software ecosystems}.
\newblock arXiv preprint arXiv:2605.13716.

\bibitem[{Ramdas et~al.(2023)Ramdas, Gr{\"u}nwald, Vovk, and Shafer}]{ramdas2023game}
Aaditya Ramdas, Peter Gr{\"u}nwald, Vladimir Vovk, and Glenn Shafer. 2023.
\newblock \href {https://doi.org/10.1214/23-STS894} {Game-theoretic statistics and safe anytime-valid inference}.
\newblock \emph{Statistical Science}, 38(4):576--601.

\bibitem[{Reynolds and Stoumbos(1999)}]{reynolds1999cusum}
Marion~R. Reynolds, Jr. and Zachary~G. Stoumbos. 1999.
\newblock \href {https://doi.org/10.1080/00224065.1999.11979900} {A {CUSUM} chart for monitoring a proportion when inspecting continuously}.
\newblock \emph{Journal of Quality Technology}, 31(1):87--108.

\bibitem[{Richter et~al.(2025)Richter, He, Minervini, and Kusner}]{richter2025auditing}
Leo Richter, Xuanli He, Pasquale Minervini, and Matt~J. Kusner. 2025.
\newblock \href {https://proceedings.iclr.cc/paper_files/paper/2025/hash/7af140f516c9f57050f7359bf53bc8f0-Abstract-Conference.html} {An auditing test to detect behavioral shift in language models}.
\newblock In \emph{International Conference on Learning Representations (ICLR)}.

\bibitem[{Sadhuka et~al.(2025)Sadhuka, Prinster, Fannjiang, Scalia, Berger, Regev, and Wang}]{sadhuka2025evaluator}
Shuvom Sadhuka, Drew Prinster, Clara Fannjiang, Gabriele Scalia, Bonnie Berger, Aviv Regev, and Hanchen Wang. 2025.
\newblock \href {https://arxiv.org/abs/2512.03109} {{E-valuator}: Reliable agent verifiers with sequential hypothesis testing}.
\newblock arXiv preprint arXiv:2512.03109.

\bibitem[{Schroeder et~al.(2026)Schroeder, Desai, Cuadron, Chu, Liu, Zhao, Krusche, Kemper, Zaharia, and Gonzalez}]{schroeder2026vcache}
Luis~Gaspar Schroeder, Aditya Desai, Alejandro Cuadron, Kyle Chu, Shu Liu, Mark Zhao, Stephan Krusche, Alfons Kemper, Matei Zaharia, and Joseph~E. Gonzalez. 2026.
\newblock \href {https://arxiv.org/abs/2502.03771} {{vCache}: Verified semantic prompt caching}.
\newblock In \emph{International Conference on Learning Representations (ICLR)}.

\bibitem[{Schuster et~al.(2022)Schuster, Fisch, Gupta, Dehghani, Bahri, Tran, Tay, and Metzler}]{schuster2022confident}
Tal Schuster, Adam Fisch, Jai Gupta, Mostafa Dehghani, Dara Bahri, Vinh~Q. Tran, Yi~Tay, and Donald Metzler. 2022.
\newblock \href {https://arxiv.org/abs/2207.07061} {Confident adaptive language modeling}.
\newblock In \emph{Advances in Neural Information Processing Systems (NeurIPS)}, volume~35.

\bibitem[{Scrivens(2026)}]{scrivens2026limits}
Arsenios Scrivens. 2026.
\newblock \href {https://arxiv.org/abs/2603.28650} {Information-theoretic limits of safety verification for self-improving systems}.
\newblock arXiv preprint arXiv:2603.28650.

\bibitem[{Shafer(2021)}]{shafer2021testing}
Glenn Shafer. 2021.
\newblock \href {https://doi.org/10.1111/rssa.12647} {Testing by betting: A strategy for statistical and scientific communication}.
\newblock \emph{Journal of the Royal Statistical Society Series A: Statistics in Society}, 184(2):407--431.

\bibitem[{Shawn(2026)}]{shawn2026pace}
Zayx Shawn. 2026.
\newblock \href {https://arxiv.org/abs/2606.08106} {{PACE}: Anytime-valid acceptance tests for self-evolving agents}.
\newblock arXiv preprint arXiv:2606.08106.

\bibitem[{Shin et~al.(2024)Shin, Ramdas, and Rinaldo}]{shin2024edetectors}
Jaehyeok Shin, Aaditya Ramdas, and Alessandro Rinaldo. 2024.
\newblock \href {https://doi.org/10.51387/23-NEJSDS51} {E-detectors: A nonparametric framework for sequential change detection}.
\newblock \emph{The New England Journal of Statistics in Data Science}, 2(2):229--260.

\bibitem[{Shinn et~al.(2023)Shinn, Cassano, Gopinath, Narasimhan, and Yao}]{shinn2023reflexion}
Noah Shinn, Federico Cassano, Ashwin Gopinath, Karthik Narasimhan, and Shunyu Yao. 2023.
\newblock Reflexion: Language agents with verbal reinforcement learning.
\newblock In \emph{Advances in Neural Information Processing Systems}.
\newblock ArXiv:2303.11366.

\bibitem[{Siegmund(1985)}]{siegmund1985sequential}
David Siegmund. 1985.
\newblock \href {https://doi.org/10.1007/978-1-4757-1862-1} {\emph{Sequential Analysis: Tests and Confidence Intervals}}.
\newblock Springer Series in Statistics. Springer, New York.

\bibitem[{Simsek(2026)}]{simsek2026when}
Baris Simsek. 2026.
\newblock \href {https://arxiv.org/abs/2604.12007} {When to forget: A memory governance primitive}.
\newblock arXiv preprint arXiv:2604.12007.

\bibitem[{Singh and Yu(2025)}]{singh2025validation}
Pragya Singh and Stanley Yu. 2025.
\newblock \href {https://ceur-ws.org/Vol-4162/paper4.pdf} {Validation-gated {H}ebbian learning for adaptive agent memory}.
\newblock In \emph{Proceedings of the First Workshop on Knowledge Graphs and Agentic Systems Interplay (NORA 2025), co-located with NeurIPS 2025}, volume 4162 of \emph{CEUR Workshop Proceedings}, pages 42--54.

\bibitem[{Sui et~al.(2025)Sui, Chuang, Wang, Zhang, Zhang, Yuan, Liu, Wen, Zhong, Zou, Chen, and Hu}]{sui2025stop}
Yang Sui, Yu-Neng Chuang, Guanchu Wang, Jiamu Zhang, Tianyi Zhang, Jiayi Yuan, Hongyi Liu, Andrew Wen, Shaochen Zhong, Na~Zou, Hanjie Chen, and Xia Hu. 2025.
\newblock \href {https://arxiv.org/abs/2503.16419} {Stop overthinking: A survey on efficient reasoning for large language models}.
\newblock \emph{Transactions on Machine Learning Research}.

\bibitem[{Sun et~al.(2025)Sun, Chen, Luo, Zhang, and Li}]{sun2025beyond}
Changzhi Sun, Xiangyu Chen, Jixiang Luo, Dell Zhang, and Xuelong Li. 2025.
\newblock \href {https://arxiv.org/abs/2512.21567} {Beyond heuristics: A decision-theoretic framework for agent memory management}.
\newblock arXiv preprint arXiv:2512.21567.

\bibitem[{Suzgun et~al.(2026)Suzgun, Yuksekgonul, Bianchi, Jurafsky, and Zou}]{suzgun2026dynamic}
Mirac Suzgun, Mert Yuksekgonul, Federico Bianchi, Dan Jurafsky, and James Zou. 2026.
\newblock \href {https://doi.org/10.18653/v1/2026.eacl-long.333} {{Dynamic Cheatsheet}: Test-time learning with adaptive memory}.
\newblock In \emph{Proceedings of the 19th Conference of the {E}uropean Chapter of the {A}ssociation for {C}omputational {L}inguistics (Volume 1: Long Papers)}, pages 7080--7106. Association for Computational Linguistics.

\bibitem[{Taatgen and Lee(2003)}]{taatgen2003production}
Niels~A. Taatgen and Frank~J. Lee. 2003.
\newblock \href {https://doi.org/10.1518/hfes.45.1.61.27224} {Production compilation: A simple mechanism to model complex skill acquisition}.
\newblock \emph{Human Factors}, 45(1):61--76.

\bibitem[{Tadepalli and Natarajan(1996)}]{tadepalli1996formal}
Prasad Tadepalli and Balas~K. Natarajan. 1996.
\newblock \href {https://arxiv.org/abs/cs/9605105} {A formal framework for speedup learning from problems and solutions}.
\newblock \emph{Journal of Artificial Intelligence Research}, 4:445--475.

\bibitem[{Tartakovsky et~al.(2014)Tartakovsky, Nikiforov, and Basseville}]{tartakovsky2014sequential}
Alexander Tartakovsky, Igor Nikiforov, and Mich{\`e}le Basseville. 2014.
\newblock \href {https://doi.org/10.1201/b17279} {\emph{Sequential Analysis: Hypothesis Testing and Changepoint Detection}}.
\newblock Chapman and Hall/CRC.

\bibitem[{{Team OLMo}(2025)}]{olmo2025olmo3}
{Team OLMo}. 2025.
\newblock {OLMo} 3.
\newblock Allen Institute for AI. Model card: \url{https://huggingface.co/allenai/Olmo-3-7B-Instruct}.

\bibitem[{Thomas et~al.(2015)Thomas, Theocharous, and Ghavamzadeh}]{thomas2015hcpi}
Philip Thomas, Georgios Theocharous, and Mohammad Ghavamzadeh. 2015.
\newblock \href {https://proceedings.mlr.press/v37/thomas15.html} {High confidence policy improvement}.
\newblock In \emph{Proceedings of the 32nd International Conference on Machine Learning (ICML)}, volume~37 of \emph{Proceedings of Machine Learning Research}, pages 2380--2388. PMLR.

\bibitem[{Thomas et~al.(2019)Thomas, Castro~da Silva, Barto, Giguere, Brun, and Brunskill}]{thomas2019seldonian}
Philip~S. Thomas, Bruno Castro~da Silva, Andrew~G. Barto, Stephen Giguere, Yuriy Brun, and Emma Brunskill. 2019.
\newblock \href {https://doi.org/10.1126/science.aag3311} {Preventing undesirable behavior of intelligent machines}.
\newblock \emph{Science}, 366(6468):999--1004.

\bibitem[{Trivedi et~al.(2022)Trivedi, Balasubramanian, Khot, and Sabharwal}]{trivedi2022musique}
Harsh Trivedi, Niranjan Balasubramanian, Tushar Khot, and Ashish Sabharwal. 2022.
\newblock \href {https://doi.org/10.1162/tacl_a_00475} {{MuSiQue}: Multihop questions via single-hop question composition}.
\newblock \emph{Transactions of the Association for Computational Linguistics}, 10:539--554.

\bibitem[{Ville(1939)}]{ville1939etude}
Jean Ville. 1939.
\newblock \href {https://www.numdam.org/item/THESE_1939__218__1_0/} {\emph{{\'E}tude critique de la notion de collectif}}.
\newblock Gauthier-Villars, Paris.

\bibitem[{Wald(1945)}]{wald1945sequential}
Abraham Wald. 1945.
\newblock \href {https://doi.org/10.1214/aoms/1177731118} {Sequential tests of statistical hypotheses}.
\newblock \emph{The Annals of Mathematical Statistics}, 16(2):117--186.

\bibitem[{Wald and Wolfowitz(1948)}]{wald1948optimum}
Abraham Wald and Jacob Wolfowitz. 1948.
\newblock \href {https://doi.org/10.1214/aoms/1177730197} {Optimum character of the sequential probability ratio test}.
\newblock \emph{The Annals of Mathematical Statistics}, 19(3):326--339.

\bibitem[{Wang et~al.(2024{\natexlab{a}})Wang, Xie, Jiang, Mandlekar, Xiao, Zhu, Fan, and Anandkumar}]{wang2024voyager}
Guanzhi Wang, Yuqi Xie, Yunfan Jiang, Ajay Mandlekar, Chaowei Xiao, Yuke Zhu, Linxi Fan, and Anima Anandkumar. 2024{\natexlab{a}}.
\newblock \href {https://arxiv.org/abs/2305.16291} {{Voyager}: An open-ended embodied agent with large language models}.
\newblock \emph{Transactions on Machine Learning Research}.

\bibitem[{Wang et~al.(2024{\natexlab{b}})Wang, Xin, Zheng, Li, Liu, Cao, Huang, Xiong, Shi, Xie, Yin, Li, Liao, and Liang}]{wang2024legoprover}
Haiming Wang, Huajian Xin, Chuanyang Zheng, Lin Li, Zhengying Liu, Qingxing Cao, Yinya Huang, Jing Xiong, Han Shi, Enze Xie, Jian Yin, Zhenguo Li, Heng Liao, and Xiaodan Liang. 2024{\natexlab{b}}.
\newblock \href {https://arxiv.org/abs/2310.00656} {{LEGO-Prover}: Neural theorem proving with growing libraries}.
\newblock In \emph{International Conference on Learning Representations (ICLR)}.

\bibitem[{Wang et~al.(2026)Wang, Suresh, Zhang, More, Jurayj, Van~Durme, Farajtabar, Khashabi, and Nalisnick}]{wang2026conformal}
Xi~Wang, Anushri Suresh, Alvin Zhang, Rishi More, William Jurayj, Benjamin Van~Durme, Mehrdad Farajtabar, Daniel Khashabi, and Eric Nalisnick. 2026.
\newblock \href {https://arxiv.org/abs/2602.03814} {{Conformal Thinking}: Risk control for reasoning on a compute budget}.
\newblock In \emph{Proceedings of the 43rd International Conference on Machine Learning (ICML)}.

\bibitem[{Wang et~al.(2023)Wang, Wei, Schuurmans, Le, Chi, Narang, Chowdhery, and Zhou}]{wang2023self}
Xuezhi Wang, Jason Wei, Dale Schuurmans, Quoc Le, Ed~Chi, Sharan Narang, Aakanksha Chowdhery, and Denny Zhou. 2023.
\newblock Self-consistency improves chain of thought reasoning in language models.
\newblock In \emph{International Conference on Learning Representations}.
\newblock ArXiv:2203.11171.

\bibitem[{Wang et~al.(2024{\natexlab{c}})Wang, Neubig, and Fried}]{wang2024trove}
Zhiruo Wang, Graham Neubig, and Daniel Fried. 2024{\natexlab{c}}.
\newblock \href {https://proceedings.mlr.press/v235/wang24az.html} {{T}ro{VE}: Inducing verifiable and efficient toolboxes for solving programmatic tasks}.
\newblock In \emph{Proceedings of the 41st International Conference on Machine Learning (ICML)}, volume 235 of \emph{Proceedings of Machine Learning Research}, pages 51177--51191. PMLR.

\bibitem[{Wang et~al.(2025{\natexlab{a}})Wang, Gandhi, Neubig, and Fried}]{wang2025inducing}
Zora~Zhiruo Wang, Apurva Gandhi, Graham Neubig, and Daniel Fried. 2025{\natexlab{a}}.
\newblock \href {https://arxiv.org/abs/2504.06821} {Inducing programmatic skills for agentic tasks}.
\newblock arXiv preprint arXiv:2504.06821.

\bibitem[{Wang et~al.(2025{\natexlab{b}})Wang, Mao, Fried, and Neubig}]{wang2025agent}
Zora~Zhiruo Wang, Jiayuan Mao, Daniel Fried, and Graham Neubig. 2025{\natexlab{b}}.
\newblock \href {https://proceedings.mlr.press/v267/wang25bx.html} {{Agent Workflow Memory}}.
\newblock In \emph{Proceedings of the 42nd International Conference on Machine Learning (ICML)}, volume 267 of \emph{Proceedings of Machine Learning Research}, pages 63897--63911. PMLR.

\bibitem[{Waudby-Smith and Ramdas(2024)}]{waudbysmith2024estimating}
Ian Waudby-Smith and Aaditya Ramdas. 2024.
\newblock \href {https://doi.org/10.1093/jrsssb/qkad009} {Estimating means of bounded random variables by betting}.
\newblock \emph{Journal of the Royal Statistical Society Series B: Statistical Methodology}, 86(1):1--27.

\bibitem[{Wu and Yang(2016)}]{wu2016quickest}
Jingxian Wu and Jing Yang. 2016.
\newblock \href {https://arxiv.org/abs/1601.06868} {Quickest change detection with mismatched post-change models}.
\newblock arXiv preprint arXiv:1601.06868.

\bibitem[{Wu and Canedo(2026)}]{wu2026invalidation}
Michael Wu and Arquimedes Canedo. 2026.
\newblock \href {https://arxiv.org/abs/2609.00243} {Invalidation contracts for cross-episode agent memory}.
\newblock arXiv preprint arXiv:2609.00243.

\bibitem[{Wu et~al.(2025)Wu, Yin, Kang, Zhang, Xu, Chen, and Zhang}]{wu2025sgm}
Xuening Wu, Shenqin Yin, Yanlan Kang, Xinhang Zhang, Qianya Xu, Zeping Chen, and Wenqiang Zhang. 2025.
\newblock \href {https://arxiv.org/abs/2510.10232} {{SGM}: A statistical {G}{\"o}del machine for risk-controlled recursive self-modification}.
\newblock arXiv preprint arXiv:2510.10232.

\bibitem[{Xia et~al.(2025)Xia, Leong, Wang, Li, and Li}]{xia2025tokenskip}
Heming Xia, Chak~Tou Leong, Wenjie Wang, Yongqi Li, and Wenjie Li. 2025.
\newblock \href {https://doi.org/10.18653/v1/2025.emnlp-main.165} {{T}oken{S}kip: Controllable chain-of-thought compression in {LLM}s}.
\newblock In \emph{Proceedings of the 2025 Conference on Empirical Methods in Natural Language Processing}, pages 3351--3363. Association for Computational Linguistics.

\bibitem[{Xu et~al.(2025)Xu, Hu, Zhang, Yu, Tang, Song, Duan, Ai, and Shi}]{xu2025sedm}
Haoran Xu, Jiacong Hu, Ke~Zhang, Lei Yu, Yuxin Tang, Xinyuan Song, Yiqun Duan, Lynn Ai, and Bill Shi. 2025.
\newblock \href {https://arxiv.org/abs/2509.09498} {{SEDM}: Scalable self-evolving distributed memory for agents}.
\newblock arXiv preprint arXiv:2509.09498.

\bibitem[{Xu et~al.(2021)Xu, Mei, and Moustakides}]{xu2021optimum}
Qunzhi Xu, Yajun Mei, and George~V. Moustakides. 2021.
\newblock \href {https://doi.org/10.1109/TIT.2021.3074961} {Optimum multi-stream sequential change-point detection with sampling control}.
\newblock \emph{IEEE Transactions on Information Theory}, 67(11):7627--7636.

\bibitem[{Xu et~al.(2024)Xu, Karampatziakis, and Mineiro}]{xu2024active}
Ziyu Xu, Nikos Karampatziakis, and Paul Mineiro. 2024.
\newblock \href {https://proceedings.neurips.cc/paper_files/paper/2024/hash/6eb05d8bc6bd7bb6868c64b5802125bd-Abstract-Conference.html} {Active, anytime-valid risk controlling prediction sets}.
\newblock In \emph{Advances in Neural Information Processing Systems (NeurIPS)}, volume~37.

\bibitem[{Yang et~al.(2025)Yang, Li, Yang et~al.}]{yang2025qwen3}
An~Yang, Anfeng Li, Baosong Yang, et~al. 2025.
\newblock \href {https://arxiv.org/abs/2505.09388} {{Qwen3} technical report}.
\newblock arXiv:2505.09388.

\bibitem[{Yang et~al.(2024{\natexlab{a}})Yang, Yu, Zhang, Cao, Xu, Zhang, Gonzalez, and Cui}]{yang2024buffer}
Ling Yang, Zhaochen Yu, Tianjun Zhang, Shiyi Cao, Minkai Xu, Wentao Zhang, Joseph~E. Gonzalez, and Bin Cui. 2024{\natexlab{a}}.
\newblock \href {https://proceedings.neurips.cc/paper_files/paper/2024/hash/cde328b7bf6358f5ebb91fe9c539745e-Abstract-Conference.html} {{Buffer of Thoughts}: Thought-augmented reasoning with large language models}.
\newblock In \emph{Advances in Neural Information Processing Systems (NeurIPS)}, volume~37.

\bibitem[{Yang et~al.(2024{\natexlab{b}})Yang, Gribovskaya, Kassner, Geva, and Riedel}]{yang2024do}
Sohee Yang, Elena Gribovskaya, Nora Kassner, Mor Geva, and Sebastian Riedel. 2024{\natexlab{b}}.
\newblock \href {https://doi.org/10.18653/v1/2024.acl-long.550} {Do large language models latently perform multi-hop reasoning?}
\newblock In \emph{Proceedings of the 62nd Annual Meeting of the Association for Computational Linguistics (Volume 1: Long Papers)}, pages 10210--10229. Association for Computational Linguistics.

\bibitem[{Yao et~al.(2023{\natexlab{a}})Yao, Yu, Zhao, Shafran, Griffiths, Cao, and Narasimhan}]{yao2023tree}
Shunyu Yao, Dian Yu, Jeffrey Zhao, Izhak Shafran, Thomas~L. Griffiths, Yuan Cao, and Karthik Narasimhan. 2023{\natexlab{a}}.
\newblock Tree of thoughts: Deliberate problem solving with large language models.
\newblock In \emph{Advances in Neural Information Processing Systems}.
\newblock ArXiv:2305.10601.

\bibitem[{Yao et~al.(2023{\natexlab{b}})Yao, Zhao, Yu, Du, Shafran, Narasimhan, and Cao}]{yao2023react}
Shunyu Yao, Jeffrey Zhao, Dian Yu, Nan Du, Izhak Shafran, Karthik Narasimhan, and Yuan Cao. 2023{\natexlab{b}}.
\newblock {ReAct}: Synergizing reasoning and acting in language models.
\newblock In \emph{International Conference on Learning Representations}.
\newblock ArXiv:2210.03629.

\bibitem[{Yates et~al.(2017)Yates, Ciblat, Yener, and Wigger}]{yates2017age}
Roy~D. Yates, Philippe Ciblat, Aylin Yener, and Mich{\`e}le Wigger. 2017.
\newblock \href {https://doi.org/10.1109/ISIT.2017.8006506} {Age-optimal constrained cache updating}.
\newblock In \emph{Proceedings of the IEEE International Symposium on Information Theory (ISIT)}, pages 141--145.

\bibitem[{Yu et~al.(2026{\natexlab{a}})Yu, Fan, Wu, Zheng, Zhou, and Li}]{yu2026comem}
Chengxin Yu, Zhaoxin Fan, Faguo Wu, Hongwei Zheng, Yun Zhou, and Zhiyu Li. 2026{\natexlab{a}}.
\newblock \href {https://arxiv.org/abs/2609.15009} {{CoMem}: Collective-individual memory synergy for evolutionary multi-agent systems}.
\newblock arXiv preprint arXiv:2609.15009.

\bibitem[{Yu et~al.(2026{\natexlab{b}})Yu, Zeng, Zhu, Huang, Zeng, and Jing}]{yu2026anytime}
Chengyao Yu, Hao Zeng, Youxin Zhu, Jianguo Huang, Huajun Zeng, and Bingyi Jing. 2026{\natexlab{b}}.
\newblock \href {https://arxiv.org/abs/2601.22446} {Anytime safe {PAC} efficient reasoning}.
\newblock In \emph{Proceedings of the 43rd International Conference on Machine Learning (ICML)}.

\bibitem[{Yu et~al.(2024)Yu, Xu, Weston, and Kulikov}]{yu2024distilling}
Ping Yu, Jing Xu, Jason Weston, and Ilia Kulikov. 2024.
\newblock \href {https://arxiv.org/abs/2407.06023} {Distilling {System 2} into {System 1}}.
\newblock arXiv preprint arXiv:2407.06023.

\bibitem[{Yuan et~al.(2024)Yuan, Chen, Wang, Fung, Peng, and Ji}]{yuan2024craft}
Lifan Yuan, Yangyi Chen, Xingyao Wang, Yi~R. Fung, Hao Peng, and Heng Ji. 2024.
\newblock {CRAFT}: Customizing {LLMs} by creating and retrieving from specialized toolsets.
\newblock In \emph{International Conference on Learning Representations}.
\newblock ArXiv:2309.17428.

\bibitem[{Zeng et~al.(2025)Zeng, Huang, Jing, Wei, and An}]{zeng2025provable}
Hao Zeng, Jianguo Huang, Bingyi Jing, Hongxin Wei, and Bo~An. 2025.
\newblock \href {https://arxiv.org/abs/2510.09133} {On the provable performance guarantee of efficient reasoning models}.
\newblock arXiv preprint arXiv:2510.09133.

\bibitem[{Zhang et~al.(2025{\natexlab{a}})Zhang, Fu, Wang, Wan, Yu, and Yan}]{zhang2025gmemory}
Guibin Zhang, Muxin Fu, Kun Wang, Guancheng Wan, Miao Yu, and Shuicheng Yan. 2025{\natexlab{a}}.
\newblock \href {https://doi.org/10.52202/085713-0439} {{G-Memory}: Tracing hierarchical memory for {Multi-Agent} systems}.
\newblock In \emph{Advances in Neural Information Processing Systems (NeurIPS)}, volume~38, pages 14587--14617.

\bibitem[{Zhang et~al.(2026{\natexlab{a}})Zhang, Hu, Lu, Lange, and Clune}]{zhang2025darwin}
Jenny Zhang, Shengran Hu, Cong Lu, Robert Lange, and Jeff Clune. 2026{\natexlab{a}}.
\newblock \href {https://arxiv.org/abs/2505.22954} {Darwin {G}{\"o}del machine: Open-ended evolution of self-improving agents}.
\newblock In \emph{International Conference on Learning Representations (ICLR)}.

\bibitem[{Zhang et~al.(2025{\natexlab{b}})Zhang, Lin, Hou, Feng, and Li}]{zhang2025adaptthink}
Jiajie Zhang, Nianyi Lin, Lei Hou, Ling Feng, and Juanzi Li. 2025{\natexlab{b}}.
\newblock \href {https://doi.org/10.18653/v1/2025.emnlp-main.184} {{A}dapt{T}hink: Reasoning models can learn when to think}.
\newblock In \emph{Proceedings of the 2025 Conference on Empirical Methods in Natural Language Processing}, pages 3716--3730. Association for Computational Linguistics.

\bibitem[{Zhang et~al.(2026{\natexlab{b}})Zhang, Hu, Upasani, Ma, Hong, Kamanuru, Rainton, Wu, Ji, Li, Thakker, Zou, and Olukotun}]{zhang2026agentic}
Qizheng Zhang, Changran Hu, Shubhangi Upasani, Boyuan Ma, Fenglu Hong, Vamsidhar Kamanuru, Jay Rainton, Chen Wu, Mengmeng Ji, Hanchen Li, Urmish Thakker, James Zou, and Kunle Olukotun. 2026{\natexlab{b}}.
\newblock \href {https://arxiv.org/abs/2510.04618} {{Agentic Context Engineering}: Evolving contexts for self-improving language models}.
\newblock In \emph{International Conference on Learning Representations (ICLR)}.
\newblock Code: \url{https://github.com/ace-agent/ace}.

\bibitem[{Zhang et~al.(2025{\natexlab{c}})Zhang, Wornow, Wan, and Olukotun}]{zhang2025agentic}
Qizheng Zhang, Michael Wornow, Gerry Wan, and Kunle Olukotun. 2025{\natexlab{c}}.
\newblock \href {https://arxiv.org/abs/2506.14852} {{Agentic Plan Caching}: Test-time memory for fast and cost-efficient {LLM} agents}.
\newblock In \emph{Advances in Neural Information Processing Systems (NeurIPS)}, volume~38.

\bibitem[{Zhang et~al.(2026{\natexlab{c}})Zhang, Wang, Zhou, Liao, Feng, Li, Zheng, Zhang, Wen, Li, Xiong, Qi, Tang, and Wen}]{zhang2026memrl}
Shengtao Zhang, Jiaqian Wang, Ruiwen Zhou, Junwei Liao, Yuchen Feng, Zhuo Li, Yujie Zheng, Weinan Zhang, Ying Wen, Zhiyu Li, Feiyu Xiong, Yutao Qi, Bo~Tang, and Muning Wen. 2026{\natexlab{c}}.
\newblock \href {https://arxiv.org/abs/2601.03192} {{MemRL}: Self-evolving agents via runtime reinforcement learning on episodic memory}.
\newblock arXiv preprint arXiv:2601.03192.

\bibitem[{Zhang et~al.(2026{\natexlab{d}})Zhang, Cui, Wang, Li, Qiu, Zhu, and He}]{zhang2026ratchet}
Xing Zhang, Yanwei Cui, Guanghui Wang, Ziyuan Li, Wei Qiu, Bing Zhu, and Peiyang He. 2026{\natexlab{d}}.
\newblock \href {https://arxiv.org/abs/2605.22148} {{Ratchet}: How reliable must an {LLM} judge be to retire a skill?}
\newblock arXiv preprint arXiv:2605.22148.

\bibitem[{Zhang et~al.(2026{\natexlab{e}})Zhang, Wu, Yu, Wu, and Wang}]{zhang2026liveevo}
Yaolun Zhang, Yiran Wu, Yijiong Yu, Qingyun Wu, and Huazheng Wang. 2026{\natexlab{e}}.
\newblock \href {https://arxiv.org/abs/2602.02369} {{Live-Evo}: Online evolution of agentic memory from continuous feedback}.
\newblock arXiv preprint arXiv:2602.02369.

\bibitem[{Zhao et~al.(2024)Zhao, Huang, Xu, Lin, Liu, and Huang}]{zhao2024expel}
Andrew Zhao, Daniel Huang, Quentin Xu, Matthieu Lin, Yong-Jin Liu, and Gao Huang. 2024.
\newblock \href {https://doi.org/10.1609/aaai.v38i17.29936} {{ExpeL}: {LLM} agents are experiential learners}.
\newblock \emph{Proceedings of the AAAI Conference on Artificial Intelligence}, 38(17):19632--19642.

\bibitem[{Zhong et~al.(2024)Zhong, Guo, Gao, Ye, and Wang}]{zhong2024memorybank}
Wanjun Zhong, Lianghong Guo, Qiqi Gao, He~Ye, and Yanlin Wang. 2024.
\newblock \href {https://doi.org/10.1609/aaai.v38i17.29946} {{MemoryBank}: Enhancing large language models with long-term memory}.
\newblock \emph{Proceedings of the AAAI Conference on Artificial Intelligence}, 38(17):19724--19731.

\bibitem[{Zhong et~al.(2023)Zhong, Wu, Manning, Potts, and Chen}]{zhong2023mquake}
Zexuan Zhong, Zhengxuan Wu, Christopher~D. Manning, Christopher Potts, and Danqi Chen. 2023.
\newblock {MQuAKE}: Assessing knowledge editing in language models via multi-hop questions.
\newblock In \emph{Proceedings of the 2023 Conference on Empirical Methods in Natural Language Processing (EMNLP)}, pages 15686--15702.

\bibitem[{Zhou et~al.(2026{\natexlab{a}})Zhou, Rong, and Lin}]{anonymous2026certification}
Jianghong Zhou, Hao Rong, and Chen Lin. 2026{\natexlab{a}}.
\newblock Certification as a rate limit on self-improving {LLM} agents.
\newblock Under review.

\bibitem[{Zhou et~al.(2026{\natexlab{b}})Zhou, Rong, Lin, and Zhang}]{anonymous2026verifiers}
Jianghong Zhou, Hao Rong, Chen Lin, and Daoan Zhang. 2026{\natexlab{b}}.
\newblock More verifiers are not always better: Robust shared memory for multi-agent {LLM} systems.
\newblock Under review.

\bibitem[{Zhu et~al.(2023)Zhu, Sheng, Zheng, Barrett, Jordan, and Jiao}]{zhu2023towards}
Banghua Zhu, Ying Sheng, Lianmin Zheng, Clark Barrett, Michael~I. Jordan, and Jiantao Jiao. 2023.
\newblock \href {https://proceedings.neurips.cc/paper_files/paper/2023/hash/b914a8fcea5c176cf1ed75c762ce27fd-Abstract-Conference.html} {Towards optimal caching and model selection for large model inference}.
\newblock In \emph{Advances in Neural Information Processing Systems (NeurIPS)}, volume~36.

\bibitem[{Zhu et~al.(2026)Zhu, Li, Zhang, Liu, and Yang}]{zhu2026helamem}
Jinchang Zhu, Jindong Li, Cheng Zhang, Jiahong Liu, and Menglin Yang. 2026.
\newblock \href {https://doi.org/10.18653/v1/2026.acl-long.625} {{HeLa-Mem}: {Hebbian} learning and associative memory for {LLM} agents}.
\newblock In \emph{Proceedings of the 64th Annual Meeting of the {A}ssociation for {C}omputational {L}inguistics (Volume 1: Long Papers)}, pages 13757--13769. Association for Computational Linguistics.

\bibitem[{Zrnic and Cand{\`e}s(2024)}]{zrnic2024active}
Tijana Zrnic and Emmanuel~J. Cand{\`e}s. 2024.
\newblock \href {https://proceedings.mlr.press/v235/zrnic24a.html} {Active statistical inference}.
\newblock In \emph{Proceedings of the 41st International Conference on Machine Learning (ICML)}, volume 235 of \emph{Proceedings of Machine Learning Research}, pages 62993--63010. PMLR.

\end{thebibliography}

\appendix
\section{Proofs for \cref{sec:hebbian}}
\label{app:hebbian}

\paragraph{Setting.} Outcomes $Y_1,Y_2,\dots$ are independent, with law $P_0$ before an unknown change time $t_c$ and $P_1$ from $t_c$ on. A score rule uses a bounded score $X=f(Y)$ with $\bar x=\operatorname{ess\,sup}X$ and $\bar b=\operatorname{ess\,sup}(-X)$, and runs $S_0=0$, $S_n=\max\{0,S_{n-1}+X_n\}$, $\tau_h=\inf\{n:S_n\ge h\}$. We assume $\mu_0\defeq\E_0X<0<P_0(X>0)$. The ARL is $A(h)=\E_\infty\tau_h$. Lorden's worst-case delay $\bar D(h)=\sup_{t_c}\operatorname{ess\,sup}\E_{t_c}[(\tau_h-t_c+1)^+\mid Y_{<t_c}]$ equals the zero-start delay $D(h)=\E_1\tau_h$. The reason is that $s\mapsto S_n(s)$ is monotone, so the worst pre-change state is $S=0$. We write $W_n=\sum_{k\le n}X_k$, $\varphi_j(\theta)=\ln\E_je^{\theta X}$, and $\theta_j>0$ for the positive root of $\varphi_j$.

\paragraph{Hebbian rules and the exact level.} Start from $c_0=1$, update $c\gets\operatorname{clip}_{[0,1]}(c+\eta_+\1\{Y=1\}-\eta_-\1\{Y=0\})$, and retire at $c<\tau$. With $S=(1-c)/\eta_-$ and $\rho=\eta_+/\eta_-$, the upper clip is the reflection at $0$. The lower clip can bind only on a step that retires under both descriptions. So the rule is \eqref{eq:scorerule} with score $+1$ on failure, $-\rho$ on success, and retirement at $S>(1-\tau)/\eta_-$. For $\eta_+=\eta_-=\eta$ ($\rho=1$), $S$ is integer and the retirement time is $\tau_H$ with $H=\lfloor(1-\tau)/\eta\rfloor+1$. A first-passage recursion ($t_0=1/q$, $t_k=1/q+\lambda t_{k-1}$, $\lambda=(1-q)/q$) gives the exact run length
\[
\E_q\tau_H=\frac{1}{q(\lambda-1)}\Bigl[\frac{\lambda(\lambda^H-1)}{\lambda-1}-H\Bigr]
\]
for $q\ne\tfrac12$, and $\E_{1/2}\tau_H=H(H+1)$.
This agrees with an exact Markov-chain solve to $10^{-46}$ relative error. Floating-point implementations of ``$c<\tau$'' can retire one failure early (e.g.\ $\eta=0.05,\tau=0.9$), so we recommend an integer counter.

\paragraph{Proof of \cref{rem:implied}.} By convexity, $\varphi_0$ has a unique positive root $\theta_0$ and $\varphi_0'(\theta_0)>0$. Then $\tilde P_0(dy)=e^{\theta_0f(y)}P_0(dy)$ is a probability measure with $\ln(d\tilde P_0/dP_0)=\theta_0f$. Multiplying by $\theta_0>0$ commutes with $\max\{0,\cdot\}$, so $\theta_0S_n$ is Page's log-likelihood CUSUM for $P_0\to\tilde P_0$ with threshold $\theta_0h$. Its Lorden optimality for that change is due to \citet{moustakides1986optimal}. It is the likelihood-ratio CUSUM of no other alternative because $\ln(dQ/dP_0)=\kappa f$ with $E_0e^{\kappa f}=1$ forces $\kappa=\theta_0$. For the $\pm1$ rule $e^{\theta_0}=(1-q_0)/q_0$, so $\tilde q=q_0e^{\theta_0}=1-q_0$. More generally, the rule ``$+a$ on failure, $-b$ on success'' is the exact CUSUM along the curve $\{(q_0,q_1):\rho(q_0,q_1)=b/a\}$.

\begin{lemma}[Page cycles and Lundberg bounds]
\label{lem:page}
Let $N_h=\inf\{n\ge1:W_n\le0\text{ or }W_n\ge h\}$, $p_h=\P(W_{N_h}\ge h)$ and $N_-=\inf\{n\ge1:W_n\le0\}$. (a) If $\E N_h<\infty$ and $p_h>0$, then $\E\tau_h=\E N_h/p_h$. (b) If $\E X<0$ with adjustment coefficient $\theta$, and $c_\theta=\E[1-e^{\theta X};X<0]$, then $c_\theta e^{-\theta(h+\bar x)}\le p_h\le e^{-\theta h}$ and $\P(\sup_nW_n\ge x)\le e^{-\theta x}$. (c) If $\E X<0$, then $\E N_-=1/\P(\sup_n W_n=0)<\infty$.
\end{lemma}
\begin{proof}
(a) Until $N_h$ the walk and $S$ coincide. On a downward exit $S$ restarts at $0$, so $\tau_h$ is a geometric($p_h$) sum of i.i.d.\ cycle lengths, and Wald's identity applies \citep{page1954continuous,siegmund1985sequential}. (b) $e^{\theta W_n}$ is a martingale bounded up to $N_h$, so $1=\E e^{\theta W_{N_h}}$. Split into upward exits (value in $[e^{\theta h},e^{\theta(h+\bar x)})$) and downward exits (value ${\le}1$, and ${=}e^{\theta X_1}$ on $\{X_1<0\}$). The maximal inequality follows from optional stopping at the first passage above $x$. (c) Duality: $\{W_1>0,\dots,W_n>0\}$ is the event that $n$ is a strict ascending ladder epoch of the reversed walk. The number of such epochs is geometric with mean $1/\P(\sup W=0)$.
\end{proof}

\paragraph{Proof of \cref{thm:trichotomy}.} \emph{ARL.} By \cref{lem:page}(a)--(c) under $P_0$, $e^{\theta_0h}\le A(h)\le(\E_0N_-/c_0)e^{\theta_0(h+\bar x)}$, so $\theta_0h\le\ln A\le\theta_0h+\Delta_0$ with an explicit constant $\Delta_0$.
\emph{(i) $\mu_1>0$.} Since $S_n\ge W_n$, $\tau_h$ is at most the first passage of $W$ above $h$, and Wald gives $D\le(h+\bar x)/\mu_1$. Conversely $\mu_1D=\E_1W_{\tau_h}\ge h-\E_1|\inf_nW_n|\ge h-1/\beta_1$, where $\beta_1>0$ solves $\E_1e^{-\beta_1X}=1$ (maximal inequality for $-W$). Substituting $h=(\ln A+O(1))/\theta_0$ gives $D=\ln A/(\theta_0\mu_1)+O(1)$.
\emph{(ii) $\mu_1=0$.} $S_n^2-n\sigma_1^2$ is a supermartingale, because $\E[\max(0,s+X)^2]\le s^2+\sigma_1^2$. Hence $D\ge h^2/\sigma_1^2$. Wald's second identity on a Page cycle, together with $p_h\ge\E_1X^-/(h+\bar x)$, gives $D\le[(h+\bar x)^2+\bar b^2(h+\bar x)/\E_1X^-]/\sigma_1^2$.
\emph{(iii) $\mu_1<0$.} Apply the ARL argument under $P_1$ with $\theta_1$: $e^{\theta_1h}\le D\le(\E_1N_-/c_1)e^{\theta_1(h+\bar x)}$. Since $e^{\theta_1h}=(e^{\theta_0h})^{\gamma}$, this gives $D\asymp A^\gamma$. Moreover $\gamma<1$ iff $\varphi_1(\theta_0)>0$, which holds when the score law under $P_1$ stochastically dominates that under $P_0$. For Bernoulli rules this is equivalent to $q_1>q_0$. If $P_1(X>0)=0$ the rule never alarms after the change.
\emph{Efficiency.} Because $\ln(d\tilde P_0/dP_0)=\theta_0f$,
\begin{align*}
\theta_0\mu_1&=\E_1\ln\tfrac{d\tilde P_0}{dP_0}\\
&=\E_1\ln\tfrac{dP_1}{dP_0}-\E_1\ln\tfrac{dP_1}{d\tilde P_0}\\
&=\KL(P_1\|P_0)-\KL(P_1\|\tilde P_0).
\end{align*}
Dividing the optimal delay $\ln A/\KL(P_1\|P_0)$ \citep{lorden1971procedures} by $\ln A/(\theta_0\mu_1)$ gives \eqref{eq:eff}. Equality holds iff $\tilde P_0=P_1$, i.e.\ $\theta_0f=\ln dP_1/dP_0$. \qed

\paragraph{Proof of \cref{cor:lr}.} The Bernoulli log-likelihood ratio is $\ln(q_1/q_0)>0$ on a failure and $\ln\frac{1-q_1}{1-q_0}<0$ on a success. A rule with scores $(+1,-\rho)$ is proportional to it iff $\rho=\rho(q_0,q_1)$, and the retirement threshold in log-likelihood units is $\ln(q_1/q_0)\,(1-\tau)/\eta_-$. The symmetric rule is optimal iff $q_1=1-q_0$.

\paragraph{Unknown degradation.}
\begin{proposition}[First-order minimax tuning]
\label{prop:minimax}
Let $q_1\in[q_-,q_+]$ with $q_0<q_-$. For every detector with ARL $\ge A$, the worst-case delay over this interval is at least $(1+o(1))\ln A/\KL(q_-\|q_0)$, and the likelihood-ratio rule tuned at $q_-$ attains it.
\end{proposition}
\begin{proof}
The lower bound is Lorden's bound at $q_1=q_-$ \citep{lorden1971procedures}. The rule tuned at $q_-$ has scores equal to the log-likelihood ratio of $q_-$ against $q_0$, so $\theta_0=1$, and its post-change drift $\mu_1(q_1)$ increases in $q_1$ because the failure score is positive and the success score negative. By \cref{thm:trichotomy}(i) its delay is $\ln A/\mu_1(q_1)+O(1)\le\ln A/\mu_1(q_-)+O(1)$, and $\mu_1(q_-)=\KL(q_-\|q_0)$.
\end{proof}
\begin{proposition}[Unclipped counters; restates \cref{rem:unclipped}]
\label{prop:unclipped}
Consider an unclipped counter $C_n=C_{n-1}+\1\{Y_n=1\}-\1\{Y_n=0\}$ started at $c_0$ and deleted at $C\le0$ (ExpeL: $c_0=2$), or the majority rule ``flag when harmful $\ge$ helpful'' (ACE's released code). (a)~$A=\infty$, since false deletion has probability $(q_0/(1-q_0))^{c_0}<1$, but no worst-case delay is finite: for $q_1>\tfrac12$ the delay after a change equals the credit banked before it divided by $2q_1-1$. (b)~Under partial staleness ($q_1<\tfrac12$) the stale macro is never flagged (ACE) or deleted (ExpeL) with probability at least $1-2q_1$ (majority rule) or $1-(q_1/(1-q_1))^{c_0}$, for every change time (averaged over the pre-change history).
\end{proposition}
\paragraph{Proof of \cref{prop:unclipped}.} (a) Without a cap, a $\pm1$ walk with up-probability $1-q_0>\tfrac12$ started at $c_0\ge1$ hits $0$ with probability $\lambda_0^{-c_0}<1$, so $A=\infty$. Given the pre-change history on $\{\tau\ge t_c\}$, the post-change walk starts at $C_{t_c-1}$ with drift $-(2q_1-1)$. Wald's identity gives the conditional delay $C_{t_c-1}/(2q_1-1)$, and the all-success history makes $C_{t_c-1}=c_0+t_c-1$ arbitrarily large, so $\bar D=\infty$. (b) For $q_1<\tfrac12$, the probability that a walk from $c$ never hits $0$ is $h_1(c)=1-\lambda_1^{-c}$. The process $h_1(C_n)\1\{\tau>n\}$ is a $P_\infty$-submartingale, so $\P_{t_c}(\tau=\infty)$ is nondecreasing in $t_c$. Its value at $t_c=1$ is $1-\lambda_1^{-c_0}$, and for the majority rule it is $(1-q_1)h_1(1)=1-2q_1$. Exact dynamic programming reproduces these values to 8 digits.

\paragraph{Time decay is a time-to-live.} Suppose confidence also decays as $c\gets(1-\lambda)c$ at every query, and the template is used with probability $\phi$ per query. Then any run of $k^\star=\lfloor\ln\tau/\ln(1-\lambda)\rfloor+1$ consecutive queries that do not use the template retires the macro, whatever its reliability. The expected waiting time for such a run is $((1-\phi)^{-k^\star}-1)/\phi$ queries, i.e.\ $(1-\phi)^{-k^\star}-1$ uses. For $\lambda=10^{-3}$ and $\tau=0.85$ we get $k^\star=163$, so a perfectly reliable macro used by $1\%$ of queries is retired after at most $4.15$ uses on average.

\paragraph{The negative-drift regime in prior work: a correction.} \citet{wu2016quickest} study Page's CUSUM run with a mismatched post-change density $g$. Their eq.~(26) defines it by the restarted recursion $\tilde C_{n+1}=\max(1,\tilde C_n)\,\tilde\Lambda_{n+1}$, i.e.\ \eqref{eq:scorerule} with $f=\ln(g/f_0)$, for which $\theta_0=1$ and $\tilde P_0=g$. Their Thm.~2 states that if $\KL(P_1\|P_0)<\KL(P_1\|g)$ (their $D_{10}<\tilde D_{11}$), the average detection delay is infinite for every false-alarm level below one, and they stress that the statement is not asymptotic. This is our regime (iii), and there the delay is finite whenever $P_1(f>0)>0$: $D(h)\le(\E_1N_-/c_1)e^{\theta_1(h+\bar x)}$ (proof of \cref{thm:trichotomy}). The proof of their Thm.~2 writes the CUSUM stopping time as a minimum over restart times, $\tilde\tau_c=\min_v\tau_v(A)$ (their eq.~(77)), where $\tau_v(A)$ is the one-sided test started at $v$ and never restarted, and their eq.~(78) then equates $\E_k(\min_v\tau_v(A)-k)^+$ with $\min_v\E_k(\tau_v(A)-k)^+$. Their Lemma~6 is correct: each $\tau_v(A)$ never stops with positive probability, so its mean delay is infinite. But in general only $\E\min\le\min\E$ holds, and here the inequality is strict: the CUSUM alarms at the first success of a geometric number of Page cycles (\cref{lem:page}(a)), each of finite mean length. \emph{Counterexample.} The symmetric rule with $q_0=0.05$ is the CUSUM with $g$ the failure rate $0.95$. At $q_1=0.25$, $\KL(P_1\|P_0)=0.225<\KL(P_1\|g)=1.697$, so their condition holds; their regularity conditions (their (5) and (47)) hold for i.i.d.\ data by the strong law. A single test $\tau_v(A)$ with threshold $h=4$ never stops with probability $1-3^{-4}=0.988$, yet the CUSUM's exact Lorden delays are $D(4)=232$ and $D(6)=2172$ (zero start is the worst case), and Monte Carlo gives $231.9$ and $2160$ ($2\times10^5$ and $4\times10^4$ runs). Under their own Bayesian criterion, with the change at a geometric($0.1$) time, $\E[\tilde\tau_c-\nu\mid\tilde\tau_c\ge\nu]=230.8$ and $2170.8$. What their theorem does capture is the first-order statement $D/\ln A\to\infty$ in this regime; \cref{thm:trichotomy}(iii) gives its rate, $D\asymp A^{\gamma}$ (\texttt{checks/wuyang\_check.py} in our code release).

\section{Proofs for \cref{sec:limit}}
\label{app:limit}

\paragraph{Assumptions.}
\begin{itemize}
\item \textbf{(A1) Demand.} Queries of template $i$ form independent Poisson processes of rate $\pi_i$, with $\pi_1\ge\pi_2\ge\cdots$ and $\sum_i\pi_i=1$.
\item \textbf{(A2) Audits.} Each query is audited independently with probability $p$. An audit yields \emph{one} verifier outcome, of the candidate that the policy chose to evaluate on that query, revealed after answering. Nothing else about the candidates' correctness is observed; in particular, the policy does not see whether an unaudited shadow answer agrees with the live one. (Under exact-match scoring such agreement is informative, and \cref{thm:app-agree} shows that it can speed up certification.)
\item \textbf{(A3) Changes.} Facts change at the points of an independent Poisson process of rate $\nu$ per template. Changes may or may not be announced to the policy. A macro formed in an earlier epoch never counts as coverage.
\item \textbf{(A4) Outcomes.} In environment $\mathsf E_r$ the audited outcomes of the current epoch are i.i.d.\ Bernoulli$(r)$, whichever candidate they evaluate, and independent of everything else, denoted $\mathcal G$ (arrivals, audits, change times, other templates and epochs, internal randomization).
\item \textbf{(A5) Policies.} Policies are non-anticipating: a decision at time $t$ depends only on $\mathcal G$-information up to $t$ and on outcomes revealed before $t$.
\end{itemize}
\paragraph{Precise form of \cref{thm:limit}.} Assume (A1)--(A5) and that in every epoch the audited outcomes are i.i.d.\ Bernoulli$(r_1)$, whichever candidate they evaluate (environment $\mathsf E_{r_1}$, the least-favorable good case). (i)~For every demand law $\pi$ and all $\nu,p,\delta$, every policy that is $\delta$-safe in the sense of \cref{def:safety} has $\liminf\cov\ge(1-\eps_\delta)F_\pi(\Th)$ almost surely, whether or not changes are announced and however many candidates it forms, where $\eps_\delta=O\bigl(\sqrt{\ln\ln(1/\delta)/\ln(1/\delta)}\bigr)$ depends only on $(\delta,r_0,r_1)$. (ii)~With announced changes, the SPRT that discards all evidence and restarts at each change is $\delta$-safe and has $\cov\le(1+\ln(r_1/r_0)/\ln(1/\delta))F_\pi(\Th)$. Hence $\cov^\star=(1+o(1))F_\pi(\Th)$ as $\delta\to0$, uniformly in $\pi,\nu,p$. The main text calls the environment ``good memories are only barely good''; the converse needs no other restriction on the policy than non-anticipation (A5) and the information structure (A2).

Notation: $x=\nu/p$, $\rho_i=\pi_i/(\pi_i+x)$, $a_\pm$ the log-likelihood-ratio increments on success and failure, $w=a_++a_-$, $\KL=\KL(r_1\|r_0)$, $L=\ln(1/\delta)$, $\nstar=L/\KL$, $\Th=x\nstar$ and $\beta=1-1/s$. Define $M_{\rm fix}(n,x)=\sum_i\pi_i(1-\rho_i^n)$.

\begin{lemma}[Exact stationary gap]
\label{lem:exact}
If the policy folds when the $N_e$-th outcome of epoch $e$ is observed, with $(N_e)$ i.i.d.\ copies of $N$ independent of arrivals, audits and changes, then the long-run unfolded fraction of template $i$ is $u_i=1-\E[\rho_i^N]$ a.s.\ and in mean, and $\cov=\sum_i\pi_iu_i$.
\end{lemma}
\begin{proof}
Epochs are i.i.d.\ renewal cycles of mean length $1/\nu$ with $\pi_i/\nu$ queries on average. A query at time $t$ of an epoch is unfolded iff fewer than $N$ audited queries preceded it. By the Mecke formula, $\E U_e=\pi_i\E\int_0^\ell\P(\Poi(p\pi_it)<N)\,dt$ with $\ell\sim\mathrm{Exp}(\nu)$. The count $\Poi(p\pi_iE)$ with $E\sim\mathrm{Exp}(\nu)$ is geometric with $\P(=k)=(1-\rho_i)\rho_i^k$, so $\P(\Poi(p\pi_iE)<N)=1-\E\rho_i^N$. Apply renewal--reward per template and sum, using Fatou for the lower bound and the superposition property for the upper bound.
\end{proof}

\begin{lemma}[Zipf sums]
\label{lem:zipf}
For $\pi_i=i^{-s}/\zeta(s)$, $s>1$, real $n\ge1$ and $x>0$,
\[
K_{s,n}x^{\beta}-nx\le M_{\rm fix}(n,x)\le K_{s,n}x^{\beta},
\]
where $K_{s,n}=\Gamma(\tfrac1s)\,C_{s,n}$ and $C_{s,n}=\zeta(s)^{-1/s}\Gamma(n+1-\frac1s)/((s-1)\Gamma(n))$. For integer $n$ the latter equals the occupancy form $\frac{\zeta(s)^{-1/s}}{s}\sum_{k<n}\Gamma(k+1-\frac1s)/k!$. Likewise $\ksz\Th^{\beta}-\Th\le F_\pi(\Th)\le\ksz\Th^\beta$ with $\ksz=\Gamma(\frac1s)/((s-1)\zeta(s)^{1/s})$.
\end{lemma}
\begin{proof}
The summand $g(\pi)=\pi(1-(\pi/(\pi+x))^n)$ is nondecreasing in $\pi$ and bounded by $nx$. Hence the sum lies between $\int_1^\infty$ and $\int_0^\infty$ of $g(t^{-s}/\zeta(s))\,dt$, and the two differ by at most $nx$. Substituting $y=x\zeta(s)t^s$ and integrating by parts turns $\int_0^\infty$ into a Beta integral, which gives $K_{s,n}x^\beta$. The identity $\sum_{k<n}\Gamma(k+a)/k!=\Gamma(n+a)/(a\Gamma(n))$ follows by induction. The bounds on $F_\pi$ follow in the same way with $g(\pi)=\pi(1-e^{-\Th/\pi})\le\Th$.
\end{proof}

\begin{lemma}[From Poisson to exponential]
\label{lem:poisexp}
For $n\ge1$ and $y\ge0$, $(1-C_\ast/n)(1-e^{-ny})\le1-(1+y)^{-n}\le1-e^{-ny}$ with $C_\ast=1.1047$. Consequently $(1-C_\ast/n)F_\pi(nx)\le M_{\rm fix}(n,x)\le F_\pi(nx)$. The function $F_\pi$ is concave and nondecreasing with $F_\pi(0)=0$. For random $N$, $1-\E\rho^N\le1-\rho^{\E N}$ by Jensen's inequality.
\end{lemma}

\begin{proof}[Proof of \cref{thm:limit}: converse]
Fix template $i$ and let $\mathcal H_e$ be the history up to the start of epoch $e$. We bound the expected unfolded count of one epoch given $\mathcal H_e$, and then average over epochs with a martingale strong law.

\emph{One epoch.} Condition on $\mathcal H_e$. Let $T_1<T_2<\cdots$ be the audit times of template $i$ from the epoch start (continued past its end), $Y_k$ the audited outcomes, $\sigma$ the first fold event of the epoch, of a candidate formed in any epoch ($\sigma=\infty$ if there is none), and $N=\#\{k:T_k<\sigma\}$. By (A5), $N$ is a stopping time for $\mathcal F_m=\sigma(\mathcal H_e,\mathcal G,Y_1,\dots,Y_m)$, and $\sigma$ is $\mathcal F_N$-measurable. By (A4), $d\P_{r_1}/d\P_{r_0}|_{\mathcal F_m}=e^{W_m}$, where $W_m$ is the log-likelihood-ratio walk; this holds whichever candidates the policy forms or audits, because every audit reveals one outcome with the same law. Fix an integer $n\le\nstar$ and $c\ge0$. Let $Q^{<n}$ be the number of the epoch's queries preceded by fewer than $n$ of its audits, and let $R$ be the number of those that arrive after the \emph{first} fold event $\sigma$. Every query answered by a candidate of the epoch arrives after $\sigma$, so the unfolded count satisfies $U\ge Q^{<n}-R$. Moreover, $R$ is a function of $\sigma$ and $\mathcal G$, hence $\mathcal F_N$-measurable even if the policy later unfolds and refolds, and $R>0$ implies $N\le n-1$.
\begin{itemize}
\item \emph{Low evidence.} Wald's likelihood-ratio identity for the $\mathcal F_N$-measurable $R\ge0$ gives $\E_{r_1}[R;W_N\le c,N<\infty]\le e^c\E_{r_0}[R]$. Under $r_0$ all candidates are bad, so by \cref{def:safety} the epoch has a fold event with probability at most $\delta$ given $\mathcal H_e$. Given $\mathcal F_\sigma$, the strong Markov property of the Poisson arrival, audit and change processes bounds the mean number of queries between $\sigma$ and the earlier of the $n$-th audit and the epoch end by $\min(n/p,\pi_i/\nu)$. Hence $\E_{r_0}[R\mid\mathcal H_e]\le\delta\min(n/p,\pi_i/\nu)$.
\item \emph{High evidence.} On $\{W_N>c,R>0\}$ we have $\max_{k\le n-1}W_k>c$. Since $R\le Q^{<n}$ and $Q^{<n}$ is $\mathcal G$-measurable, $\E_{r_1}[R;W_N>c\mid\mathcal H_e]\le\E Q^{<n}\cdot\beta_{n-1}(c)$, where $\beta_m(c)=\P_{r_1}(\max_{k\le m}W_k\ge c)$.
\end{itemize}
Since $\E Q^{<n}=\frac{\pi_i}{\nu}(1-\rho_i^n)$ by the computation in \cref{lem:exact}, the unfolded count $U_e$ of epoch $e$ satisfies $\E[U_e\mid\mathcal H_e]\ge\frac{\pi_i}{\nu}\,b_i$, where
\[
b_i\defeq(1-\beta_{n-1}(c))(1-\rho_i^n)-e^c\delta\min\bigl(1,\tfrac{n\nu}{p\pi_i}\bigr).
\]

\emph{Many epochs.} Let $Q_e$ be the number of queries of template $i$ in epoch $e$. The $Q_e$ are i.i.d.\ geometric with mean $\pi_i/\nu$ whatever the policy does, so $\E Q_e^2<\infty$. The differences $D_e=U_e-\E[U_e\mid\mathcal H_e]$ are martingale differences with $\E D_e^2\le\E Q_1^2$, because $0\le U_e\le Q_e$. Hence $\sum_eD_e/e$ converges a.s.\ by the $L^2$ martingale convergence theorem, and Kronecker's lemma gives $\frac1K\sum_{e\le K}D_e\to0$ a.s.\ \citep[Thm.~2.18]{hall1980martingale}. With the ordinary strong law for $(Q_e)$, $\liminf_K\sum_{e\le K}U_e/\sum_{e\le K}Q_e\ge b_i$ a.s. The last, incomplete epoch contributes a vanishing share because $Q_K/K\to0$ a.s., so the long-run unfolded fraction satisfies $\liminf u_i\ge b_i$ a.s.

\emph{Choice of $n$ and $c$.} Take $n-1=\lfloor(1-\eps)\nstar\rfloor$ and $c=(1-\eps/2)L$, with $\eps\to0$ and $\eps\sqrt{L}/\sqrt{\ln L}\to\infty$. Then $\beta_{n-1}(c)\to0$ by Azuma's maximal inequality for $W$ (the mean of $W_{n-1}$ is at most $(1-\eps)L$), and $e^c\delta=\delta^{\eps/2}\to0$. \Cref{lem:poisexp} and concavity give $1-\rho_i^n\ge(1-C_\ast/n)(1-\eps)(1-e^{-\Th/\pi_i})$, and $\min(1,\Th/\pi_i)\le\frac{e}{e-1}(1-e^{-\Th/\pi_i})$. Hence $\liminf u_i\ge(1-\eps_\delta)(1-e^{-\Th/\pi_i})$ with $\eps_\delta=O(\sqrt{\ln L/L})$, uniformly in $i$, $\pi$, $\nu$ and $p$. Summing over templates with Fatou's lemma for series gives $\liminf\cov\ge(1-\eps_\delta)F_\pi(\Th)$ a.s.

\emph{Announcements and candidates.} The argument used announcements only through $\mathcal F_m$, which contains the change times through $\mathcal G$. A policy that does not observe changes is non-anticipating for a smaller filtration, so the bound holds for it a fortiori. The number of candidates entered only through the likelihood ratio, which (A2) and (A4) make independent of which candidate is audited.
\end{proof}

\begin{proof}[Proof of \cref{thm:limit}: achievability]
At each announced change the policy discards all candidates' evidence and restarts one SPRT, so every fold event of an epoch is a crossing of that epoch's test.
Under $r\le r_0$, $\E e^{Z}\le1$ for each increment of the SPRT statistic. So $e^{\Lambda_n}$ is a nonnegative supermartingale, and Ville's inequality gives $\P(\sup_n\Lambda_n\ge L)\le\delta$ under arbitrary drift below $r_0$ and adaptive audit schedules. Under $r\ge r_1$, Wald's identity with overshoot $<a_+$ gives $\E N\le(L+a_+)/\KL$. By \cref{lem:exact}, Jensen's inequality and \cref{lem:poisexp}, $\cov_{\rm SPRT}\le F_\pi(x\E N)\le(1+a_+/L)F_\pi(\Th)$, where the last step uses concavity of $F_\pi$.
\end{proof}

The per-template gap of the SPRT can be bracketed exactly, which sharpens the Jensen step above and makes the limit quantitative at practical $\delta$.

\begin{lemma}[Finite-$\delta$ bracket]
\label{lem:exact-gap}
Let $Z$ be the SPRT increment, $a_+=\ln(r_1/r_0)$ its value on a success, and let $\xi_i>0$ solve
\begin{equation}
\E_{r_1}e^{\xi_iZ}=1+\nu/(p\pi_i).
\label{eq:xi}
\end{equation}
If every epoch runs the SPRT at level $\delta$, the unfolded fraction $u_i$ of template $i$ (so $\cov=\sum_i\pi_iu_i$) satisfies
\[
1-\delta^{\xi_i}\le u_i\le1-\delta^{\xi_i}e^{-\xi_ia_+}.
\]
Every per-epoch $\delta$-safe certifier that folds at a stopping time of the audited outcomes has $u_i\ge1-\delta^{\xi_i/(1+\xi_i)}$.
As $\nu/(p\pi_i)\to0$ with $\Th/\pi_i$ fixed, $\xi_i\ln(1/\delta)\to\Th/\pi_i$, and all three bounds tend to $1-e^{-\Th/\pi_i}$.
\end{lemma}
\begin{proof}[Proof of \cref{lem:exact-gap}]
Write $L=\ln(1/\delta)$ and $\sigma^2=\mathrm{Var}_{r_1}Z$. The right side of \cref{eq:xi} is $1/\rho_i$, and $N$ is the SPRT sample size in audited outcomes.
By \cref{lem:exact}, $u_i=1-\E\rho_i^N$, since audited outcomes of template $i$ arrive at rate $p\pi_i$ and the epoch ends at rate $\nu$. The function $\psi(\xi)=\ln\E_{r_1}e^{\xi Z}$ is convex with $\psi(0)=0$, $\psi'(0)=\KL>0$ and $\psi(\xi)\to\infty$, so \cref{eq:xi} has a unique root. Hence $M_n=\rho_i^ne^{\xi_i\Lambda_n}$ is a mean-one $\P_{r_1}$-martingale, as in Wald's fundamental identity \citep{siegmund1985sequential}.
\emph{Bracket.} Before the crossing $M_n\le e^{\xi_iL}$, and at the crossing $L\le\Lambda_N<L+a_+$. Since $N<\infty$ a.s., dominated convergence gives $\E M_N=1$, and the bounds on $\Lambda_N$ give $\delta^{\xi_i}e^{-\xi_ia_+}\le\E\rho_i^N\le\delta^{\xi_i}$.
\emph{Converse.} Write $\rho_i^N=M_N^{1/(1+\xi_i)}(\rho_i^Ne^{-\Lambda_N})^{\xi_i/(1+\xi_i)}$. By Hölder's inequality with exponents $1+\xi_i$ and $\frac{1+\xi_i}{\xi_i}$, $\E_{r_1}[\rho_i^N;N<\infty]$ is at most $\E_{r_1}[M_N;N<\infty]^{1/(1+\xi_i)}$ times $\E_{r_1}[\rho_i^Ne^{-\Lambda_N};N<\infty]^{\xi_i/(1+\xi_i)}$. The first factor is $\le1$ by Fatou's lemma. By change of measure the second expectation is $\E_{r_0}[\rho_i^N;N<\infty]$, the probability of folding a bad candidate within the epoch, which is $\le\delta$.
\emph{Limit.} Expanding $\psi(\xi)=\KL\xi+\sigma^2\xi^2/2+O(\xi^3)$ and $\ln(1+y)=y+O(y^2)$ in \cref{eq:xi} gives $\xi_iL=\frac\Th{\pi_i}\bigl(1+O(\frac{\nu}{p\pi_i}(1+\frac{\sigma^2}{\KL^2}))\bigr)$, and $\xi_ia_+\to0$.
\end{proof}
The bracket is two-sided and sharper than the Jensen bound $u_i\le1-\rho_i^{\E N}$ used for \cref{thm:limit}.
For $(r_0,r_1)=(0.85,0.95)$, $\delta=10^{-2}$ and $\Th/\pi_i=0.908$, exact dynamic programming gives $u_i=0.5610\in[0.5557,0.5643]$. Here $\xi_i=0.1762$, the converse bound is $0.4984$, the first-order value $1-e^{-\Th/\pi_i}$ is $0.5966$, and the Jensen bound is $0.5998$.

\paragraph{Universality.} Suppose $G(\theta)=\sum_{i:\pi_i\le\theta}\pi_i$ is regularly varying with index $1-\alpha\in(0,1)$. Writing $F_\pi(\Th)=1-e^{-\Th}+\int_\Th^\infty G(\Th/y)e^{-y}dy$ and applying Potter's bounds with dominated convergence gives $F_\pi(\Th)\sim\Gamma(\alpha)G(\Th)$ as $\Th\to0$. For Zipf demand, $\alpha=1/s$.

\paragraph{A finite-size law for cold starts.} The steady state of \cref{thm:limit} is reached only after about $1/\nu$ queries. The following law interpolates between a cold start and the drift floor. It is stated for a fixed certification sample size $n$, not for sequentially tested (safe) folding; extending it to safe folding is open.

\begin{theorem}[One law for cold start and drift]
\label{thm:fss}
Let demand be Zipf$(s)$, start all templates cold at time $0$, let facts change at rate $\nu$, and fold after $n$ audited uses. The probability that the query at time $T$ is not answered in one step satisfies
$\cov(T,\nu)=(pT)^{-(1-\frac1s)}\,\Phi_{s,n}(\nu T)\,(1+o(1))$ with relative error $O(m^{1/s}\ln(1/m))$, $m=\max(1/(pT),\nu/p)$, where $\Phi_{s,n}(y)=C_{s,n}\bigl[e^{-y}+y^{1-\frac1s}\gamma_{\rm inc}(\tfrac1s,y)\bigr]$, $C_{s,n}$ is as in \cref{lem:zipf}, and $\gamma_{\rm inc}$ is the lower incomplete gamma function.
\end{theorem}
At $y=\nu T=0$ this is the static amortization law $\cov(T)\sim C_{s,n}(pT)^{-(1-1/s)}$, a corollary of \citet{karlin1967central} and \citet[Prop.~7]{ohannessian2012rare} summed over $r<n$. It is a power law of practice \citep{newell1981mechanisms,heathcote2000power} whose exponent $1-1/s$, shared by memorization learning curves \citep{hutter2021learning,michaud2023quantization}, is set by the task distribution. As $y\to\infty$ it tends to $\Gamma(\tfrac1s)C_{s,n}(\nu/p)^{1-1/s}=\ksz\Th^{1-1/s}(1+O(1/n))$ with $\Th=n\nu/p$, the steady state of \cref{cor:zipf}. \Cref{app:natural} tests the static branch on natural template streams.

\begin{proof}[Proof of \cref{thm:fss}]
With changes at rate $\nu$ from a cold start at $0$, the age of the current epoch at time $T$ is $A_T=\min(T,E_\nu)$, and the audited count of template $i$ given $A_T$ is $\Poi(p\pi_iA_T)$. For audited exposure $a$ (i.e.\ $a=pA_T$), the sum $S_n(a)=\sum_i\pi_i\P(\Poi(\pi_ia)<n)$ satisfies $|S_n(a)-C_{s,n}a^{-\beta}|\le2n/a$. The summand is unimodal in $i$ with maximum at most $n/a$, and its integral is $C_{s,n}a^{-\beta}$. Averaging $(pA_T)^{-\beta}$ gives $p^{-\beta}\E A_T^{-\beta}=(pT)^{-\beta}[e^{-y}+y^{\beta}\gamma_{\rm inc}(1-\beta,y)]$ with $y=\nu T$, which is the stated $\Phi_{s,n}$. The error bound $O(m^{1/s}\ln(1/m))$ with $m=\max(1/(pT),\nu/p)$ follows by integrating $\min(2n/a,1)$ against the law of $pA_T$.
\end{proof}

\section{Certifying utility by betting}\label{app:utility}
\providecommand{\KLinf}{\KL_{\inf}}

The SPRT of \cref{sec:limit} tests reliability against $r_0=\rstar$ (\cref{eq:breakeven}), which depends on the live accuracy $a$. Here we certify the macro's \emph{net utility relative to live} on paired shadow evaluations instead, so $a$ never enters the null. The tools are testing by betting \citep{shafer2021testing,ramdas2023game,waudbysmith2024estimating} with inverse-propensity audits, as in \citet{xu2024active} and \citet{yu2026anytime}.

\paragraph{Setup.}
Fix a template and a frozen candidate. At shadow evaluation $t=1,2,\dots$ both paths run on the same instance. Live has cost $C^L_t$ and correctness $Y^L_t\in\{0,1\}$; the macro has $C^M_t$ and $Y^M_t$. The paired utility of folding is
\begin{equation}
U_t=C^L_t-C^M_t+L\,(Y^M_t-Y^L_t).
\label{eq:app-util}
\end{equation}
Given a known $B$ with $U_t\ge-B$, put $Z_t=U_t/B\in[-1,M]$. In the paired Bernoulli model ($C^L\equiv m$, $C^M\equiv1$), let $\eta=(m-1)/L<1$, $D_t=Y^M_t-Y^L_t$ and $B=L(1-\eta)$. Then $Z_t=(\eta+D_t)/(1-\eta)$ and $\E Z_t=(r-\rstar)/(1-\eta)$.

Correctness is revealed only on audited instances ($A_t=1$). Let $\mathcal F_{t-1}$ be everything that happened before $t$, observed or not, and let $W_t$ be pre-audit side information (costs, both answers, a judge score). Put $\mathcal H_t=\mathcal F_{t-1}\vee\sigma(W_t)$, and let $\mathcal G_{t-1}\subseteq\mathcal F_{t-1}$ be the system's records. We assume:
\begin{itemize}
\item[(A1)] the audit probability $p_t$ is $\mathcal H_t$-measurable and computable, and $p_t\ge\pi_t$ on $\{p_t>0\}$ for a $\mathcal G_{t-1}$-measurable floor $\pi_t\in(0,1]$;
\item[(A2)] $\P(A_t=1\mid\mathcal H_t,Z_t)=p_t$;
\item[(A3)] a computable surrogate $g_t\in[-1,\bar G]$ satisfies $Z_t=g_t$ on $\{p_t=0\}$.
\end{itemize}
The estimator $\hat Z_t=g_t+A_t(Z_t-g_t)/p_t$ (with $0/0\defeq0$) is observable. Taking $g\equiv0$ gives IPW, and $p\equiv1$ gives full information. The candidate is inadmissible under
\begin{equation}
H_0:\ \E[Z_t\mid\mathcal F_{t-1}]\le0\ \text{ a.s.\ for all }t.
\label{eq:app-null}
\end{equation}
$H_0$ leaves arbitrary the joint law of $(C^L,C^M,Y^L,Y^M)$, including the live--macro error dependence, and its dependence on the past. Drifting live accuracy, shared difficulty and adaptive query selection are therefore all allowed.

\paragraph{Test.}
Let $\kappa_t=(1+\bar G(1-\pi_t))/\pi_t$, which is $1/\pi_t$ for IPW. Take $\mathcal G_{t-1}$-measurable bets $\lambda^{(j)}_t\in[0,1/\kappa_t]$ and weights $w_j\ge0$ with $\sum_jw_j\le1$. Fold at $\tau=\inf\{n:K_n\ge1/\delta\}$, where
\begin{equation}
K_n=\sum_jw_j\prod_{t\le n}\bigl(1+\lambda^{(j)}_t\hat Z_t\bigr).
\label{eq:app-wealth}
\end{equation}
The \emph{grid mixture} at audit rate $p$ uses $\lambda^{(j)}=p\,c_j$, $c_j=(1+\eps)^{-j}$, $w_j=1/(j(j+1))$ and $\eps=1/\ln(1/\delta)$.

\begin{theorem}[Anytime validity]\label{thm:app-valid}
Under (A1)--(A3) and $H_0$, $(K_n)$ is a nonnegative supermartingale with $K_0\le1$. Hence $\P(\exists n:K_n\ge1/\delta)\le\delta$ for every schedule of shadow evaluations and every stopping or abandonment rule.
\end{theorem}
\begin{proof}
\emph{Positivity.} If $A_t=0$, then $\hat Z_t=g_t\ge-1\ge-\kappa_t$. If $A_t=1$, then $\hat Z_t=g_t(1-1/p_t)+Z_t/p_t\ge\bar G-(1+\bar G)/p_t$. This bound increases in $p_t$, so it is $\ge-\kappa_t$ because $p_t\ge\pi_t$. Hence every factor is $\ge1-\lambda^{(j)}_t\kappa_t\ge0$.
\emph{Unbiasedness.} On $\{p_t>0\}$, (A2) gives $\E[\hat Z_t\mid\mathcal H_t,Z_t]=Z_t$. On $\{p_t=0\}$, $\hat Z_t=g_t=Z_t$ by (A3). Since $\mathcal F_{t-1}\subseteq\mathcal H_t$, $\E[\hat Z_t\mid\mathcal F_{t-1}]=\E[Z_t\mid\mathcal F_{t-1}]\le0$.
\emph{Supermartingale.} Because $\lambda^{(j)}_t$ is $\mathcal F_{t-1}$-measurable, each product in \cref{eq:app-wealth} is a nonnegative supermartingale, and so is their weighted sum, by monotone convergence.
\emph{Ville.} Let $\sigma=\inf\{n:K_n\ge1/\delta\}$. Optional stopping at $\sigma\wedge n$ gives $1\ge\E K_{\sigma\wedge n}\ge\delta^{-1}\P(\sigma\le n)$. Let $n\to\infty$ \citep{ville1939etude}.
\end{proof}
\begin{remark}
Validity has three limits.
\begin{itemize}
\item Bets may not use $W_t$.
\item (A2) excludes outcome-dependent labelling, such as labels that arrive through user complaints.
\item Drift is covered only inside $H_0$, so folded macros must still be monitored.
\end{itemize}
Audits can also be triggered by disagreement. Audit only when the two answers differ, and set $g_t=(C^L_t-C^M_t)/B$ on agreement, which (A3) allows only under exact-match scoring (condition~(E) of \cref{app:agree}). Then $p_t\in\{0,1\}$ and $\kappa_t=1$, so the test keeps full-information power at a label cost of $\P(\text{answers differ})\,\E\tau$.
Unlike a sign test on discordant pairs \citep{shawn2026pace}, concordant pairs here carry the deterministic gain $\eta/(1-\eta)$. A macro slightly \emph{less} accurate than live is therefore still certifiable when $r>\rstar$.
\end{remark}

\paragraph{The information constant.}
For a law $P$ of $Z$ on $[-1,M]$ with mean $\mu$, let $g_P(c)=\E_P\ln(1+cZ)$ and $g^*(P)=\max_{c\in[0,1]}g_P(c)$, attained at $c^*$. Let $\KLinf(P)=\inf\{\KL(P\|Q):\E_QZ\le0\}$ over laws $Q$ on $[-1,M]$.

\begin{lemma}[Growth equals information]\label{lem:app-klinf}
\emph{(The dual form of $\KLinf$ is due to \citet{honda2010asymptotically}; we include a short proof for bounded $Z$.)}
$g^*(P)=\KLinf(P)$, and $g^*>0$ iff $\mu>0$. If $\mu>0$, the infimum is attained by $Q^*(dz)=P(dz)/(1+c^*z)+w\,\delta_{-1}(dz)$ with $w=1-\E_P[(1+c^*Z)^{-1}]\ge0$.
\end{lemma}
\begin{proof}
\emph{($\le$).} Let $\E_QZ\le0$ and $\ell=dP/dQ$. For every $c\in[0,1]$, Jensen gives $\E_P\ln\frac{1+cZ}{\ell}\le\ln\E_Q[(1+cZ)\1_{\ell>0}]\le\ln(1+c\,\E_QZ)\le0$.
\emph{($\ge$).} If $\mu\le0$, both sides are $0$. Otherwise $c^*>0$.
If $c^*<1$, differentiation under $\E$ is legitimate because $|Z/(1+cZ)|\le\max(\frac1c,\frac1{1-c})$. The first-order condition $\E_P[Z/(1+c^*Z)]=0$ then shows that $w=0$ and that $Q^*$ has mass $1$ and mean $0$.
If $c^*=1$, then $P(\{-1\})=0$, and the left derivative $\E_P[Z/(1+Z)]=w$ is $\ge0$, so $Q^*$ again has mass $1$ and mean $0$.
In both cases $\KL(P\|Q^*)=\E_P\ln(1+c^*Z)=g^*$.
\end{proof}
Since $Q^*$ lives on $\mathrm{supp}P\cup\{-1\}$, in the paired Bernoulli model $\KLinf$ equals the information of the three-point model. Locally, $\KLinf=\frac{\mu^2}{2\sigma^2}(1+O(\mu))$ with $\sigma^2=\mathrm{Var}(Z)$.

\paragraph{Observation models.} A lower bound on certification time depends on what the rule observes. Let $U$ be an independent randomization of the rule, and let $\xi_1,\xi_2,\dots$ be i.i.d.\ uniform audit coins, independent of $U$ and of all shadow outcomes.
\begin{itemize}
\item[(O1)] \emph{Audited utilities only.} After $t$ shadow evaluations the rule's records are $\mathcal R_t=\sigma(U,(A_s,A_sZ_s)_{s\le t})$, and $A_t=\1\{\xi_t\le p_t\}$, where $p_t$ is a fixed (law-independent) function of the records $\mathcal R_{t-1}$. Unaudited shadow outputs, including whether the two answers agree, are neither used nor allowed to trigger audits.
\item[(O2)] \emph{Audited utilities and agreement.} The records also contain the agreement indicators $G_s$, and $A_t=\1\{\xi_t\le p_t(G_t)\}$, where $p_t(\cdot)$ is a fixed function of the records before $t$; audits may thus depend on $G_t$ but not on $Z_t$ (\cref{app:agree}).
\end{itemize}
The IPW certificate ($g\equiv0$) with audits that do not depend on $W_t$ is an (O1) rule. So is our restart-only certificate, which updates its wealth only on audited pairs.

\begin{theorem}[Certification time and optimality]\label{thm:app-rate}
Let $Z_t$ be i.i.d.\ $P$ with $\mu>0$.
\emph{(a)} Suppose audits are i.i.d.\ Bernoulli$(p)$ and independent of $(Z_t)$. Under the grid mixture, the label count $N_p$ has the law of the full-information stopping time $\tau_1$, $\E\tau_p=\E\tau_1/p$, and
\[
\E N_p\le\frac{\ln\frac1\delta+2\ln\ln\frac1\delta+O(1)}{\KLinf(P)},
\]
where $O(1)$ depends only on $c^*$ and $M$.
\emph{(b)} Consider a rule of type (O1) with $p_t\le\bar p$, folding at a stopping time $\tau$ of $(\mathcal R_t)$. Let $\P_Q(\tau<\infty)\le\delta$ for every i.i.d.\ law $Q$ of $(Z_t)$ on $[-1,M]$ with $\E_QZ\le0$, and let $\P_P(\tau<\infty)=1$. Then
\begin{gather*}
\E_P\textstyle\sum_{t\le\tau}A_t\ge\ln(1/\delta)/\KLinf(P),\\
\E_P\tau\ge\ln(1/\delta)/\bigl(\bar p\,\KLinf(P)\bigr).
\end{gather*}
Hence, among (O1) rules, $\E\tau_p=(1+o(1))\ln(1/\delta)/(p\,\KLinf(P))$ is optimal as $\delta\to0$. Here $P$ is the law of the \emph{unweighted} $Z_t$. The audit rate enters only through the factor $1/p$ (proof of (a)); expressing the rate through the law of the weighted increment and dividing by $p$ again would count $p$ twice. Part (b) does not extend beyond (O1): a rule that also sees whether the answers agree can be faster (\cref{prop:app-agree-example,thm:app-agree}).
\end{theorem}
\begin{proof}
(a) With $\lambda^{(j)}=pc_j$, the factor $1+c_jA_tZ_t$ equals $1$ on unaudited steps. The wealth therefore moves only at audits, whose outcomes are i.i.d.\ $P$ and independent of the Geometric$(p)$ gaps. So $N_p\overset d=\tau_1$, and Wald's identity gives $\E\tau_p=\E N_p/p$.
Pick $j^*$ with $c_{j^*}\in[c^*/(1+\eps),c^*]$. By concavity, $g(c_{j^*})\ge g^*/(1+\eps)$. Since $K_n\ge w_{j^*}\prod_{t\le n}(1+c_{j^*}Z_t)$, Wald's identity with overshoot $\le\ln(1+M)$ gives
\begin{align*}
\E\tau_1\le{}&\tfrac{1+\eps}{g^*}\bigl[\ln\tfrac1\delta+\ln\bigl(j^*(j^*+1)\bigr)\\
&+\ln(1+M)\bigr].
\end{align*}
Finally, $j^*\le1+2\ln(1/c^*)\ln(1/\delta)$.

(b) Take $Q^*$ from \cref{lem:app-klinf}, so that $dP/dQ^*=1+c^*z$ on $\mathrm{supp}P$, and put $\ell=\ln(1+c^*Z)$. Under an i.i.d.\ law $Q$ of $(Z_t)$ the records $\mathcal R_n$ have, with respect to the law of $U$ times a fixed dominating measure, the density $\prod_{t\le n}p_t^{A_t}(1-p_t)^{1-A_t}\,q(Z_t)^{A_t}$, where $q$ is the density of $Q$. Under (O1) the audit factors are the same functions of the records under $P$ and $Q^*$, because the coins $\xi_t$ and the map $\mathcal R_{t-1}\mapsto p_t$ do not depend on the law; they cancel. The likelihood ratio of $\mathcal R_n$ is therefore $\exp\sum_{t\le n}A_t\ell(Z_t)$. (This is the step that fails outside (O1): if the records contain $W_t$, or audits depend on $W_t$, the ratio also contains the likelihood ratio of $W_t$, whose law may differ between $P$ and every null.) Since $Z_t$ and $\xi_t$ are independent of $\mathcal R_{t-1}$ and of each other, $\E_P[A_t\ell(Z_t)\mid\mathcal R_{t-1}]=p_tg^*$.
By optional stopping at $\tau\wedge n$, the KL divergence between the laws of the records up to $\tau\wedge n$ is $g^*\E_P\sum_{t\le\tau\wedge n}A_t$. Data processing on $\{\tau\le n\}$, whose $Q^*$-probability is $\le\delta$, together with $\mathrm{kl}(x,y)\ge x\ln\frac1y-h(x)$, bounds this below by $x_n\ln\frac1\delta-h(x_n)$, where $x_n=\P_P(\tau\le n)\to1$. Now let $n\to\infty$ \citep{kaufmann2016complexity,garivier2016optimal}.
Finally, $\E_P\sum_{t\le\tau}A_t=\E_P\sum_{t\le\tau}p_t\le\bar p\,\E_P\tau$.
\end{proof}

\paragraph{Why the plug-in reliability test fails.}
The SPRT is a Kelly bet: $e^{\Lambda_n}=\prod_{k\le n}(1+b(Y_k-r_0))$ with $b=\frac{r_1-r_0}{r_0(1-r_0)}$. With $r_0=\rstar$ it is the betting test on $(Y^M-\rstar)/\rstar$, and it is valid only if $a$ is known.

\begin{proposition}[Plug-in tests are not level $\delta$; restates \cref{prop:plugin}]\label{prop:app-plugin}
Let $\hat a=k/n_a$ with $k\sim\mathrm{Bin}(n_a,a)$, estimated from independent live runs. Run the one-sided SPRT of \cref{sec:limit} with $r_0=\hat a-\eta$ and $r_1=\hat a$ on i.i.d.\ macro outcomes at the break-even $r=\rstar$. Let $E=\{\eta<\hat a<1\}$ and
\[
r_{\rm mid}=\ln\tfrac{1-r_0}{1-r_1}\Big/\ln\tfrac{r_1(1-r_0)}{r_0(1-r_1)}\in(r_0,r_1).
\]
Then:
\begin{itemize}
\item[(i)] $\P(\text{fold})\ge\P(E,r_{\rm mid}<\rstar)\ge\P(\eta<\hat a\le\rstar)$;
\item[(ii)] $\P(\text{fold})\ge\E\bigl[\1_E\1_{\{r_0<\rstar\}}(\delta r_0/r_1)^{\theta}\bigr]$.
\end{itemize}
In (ii), $\theta=0$ if $\rstar\ge r_{\rm mid}$. Otherwise $\theta\in(0,1)$ solves $\rstar(r_1/r_0)^\theta+(1-\rstar)(\frac{1-r_1}{1-r_0})^\theta=1$.
For large $n_a$, the middle term of (i) is $\approx\Phi\bigl(-\frac\eta2\sqrt{n_a/(a(1-a))}\bigr)$.
\end{proposition}
\begin{proof}
Condition on $\hat a\in E$. The increments $X_k$ of $\Lambda$ are i.i.d.\ with mean $\KL(\rstar\|r_0)-\KL(\rstar\|r_1)$. This mean is affine in $\rstar$ and vanishes at $r_{\rm mid}$. If $\rstar\ge r_{\rm mid}$, the drift is nonnegative and the steps are non-degenerate, so $\limsup\Lambda_n=\infty$ and the test folds a.s. On $E$, $\hat a\le\rstar$ implies $r_{\rm mid}<r_1\le\rstar$, which proves (i).
For (ii), let $r_0<\rstar<r_{\rm mid}$. The map $\theta\mapsto\E e^{\theta X}$ is strictly convex. It equals $1$ at $0$ with negative slope and equals $1+b(\rstar-r_0)>1$ at $1$, so the root $\theta$ exists and $e^{\theta\Lambda_n}$ is a martingale. At the fold time $\sigma$, $\Lambda_\sigma\le\ln\frac1\delta+\ln\frac{r_1}{r_0}$; on $\{\sigma=\infty\}$, $\Lambda_n\to-\infty$. Bounded convergence therefore gives $1=\E[e^{\theta\Lambda_\sigma};\sigma<\infty]\le(r_1/(\delta r_0))^\theta\,\P(\sigma<\infty\mid\hat a)$.
\end{proof}

\begin{table}[h]
\centering\small\setlength{\tabcolsep}{3pt}
\begin{tabular}{@{}lccccccc@{}}
\toprule
$n_a$ & 10 & 20 & 50 & 100 & 200 & 500 & 1000\\
\midrule
$\delta=0.05$ & .28 & .34 & .30 & .25 & .19 & .11 & .081\\
$\delta=0.01$ & .27 & .33 & .26 & .20 & .13 & .058 & .030\\
bound (i) & .26 & .32 & .23 & .20 & .10 & .025 & .003\\
\bottomrule
\end{tabular}
\caption{Exact false-fold probability of the plug-in SPRT for a macro at break-even, computed by dynamic programming ($a=0.9$, $(m-1)/L=0.05$, horizon 2000 uses).}
\label{tab:app-plugin}
\end{table}
\Cref{tab:app-plugin} shows that the excess over $\delta$ persists long after the middle term of (i) drops below $\delta$. At $n_a=1000$ and $\delta=0.01$, bound (ii) gives $0.035$ and the exact value is $0.034$.
Macros slightly below break-even behave similarly. At shortfall $0.04$ with $n_a\le20$ live labels, the rate is $0.1$--$0.4$ across $a\in[0.5,0.9]$ and $\eta\in[0.05,0.2]$.
On the full \textsc{KQA-Stream} tables (Qwen3-8B, 30 streams) both certificates stay far below $\delta$ on below-break-even candidates. The plug-in SPRT folds $0.32\%$ of them (477 of 147,349 encounters) and the betting test $0.017\%$ (33 of 190,187), while counter rules fold all of them. The failure of the plug-in test is a boundary phenomenon: it needs candidates just below break-even and live accuracy estimated from few labels. Most spurious candidates in our streams are far below break-even: the ones that counters fold fall short by a median of $0.68$.

\paragraph{Pairing.}
Write $P_1\le_{\rm cx}P_2$ if $\E\phi(Z_1)\le\E\phi(Z_2)$ for all convex $\phi$.
\begin{theorem}[Pairing gain]\label{thm:app-pairing}
If $P_1\le_{\rm cx}P_2$ on $[-1,M]$, then $\KLinf(P_1)\ge\KLinf(P_2)$, strictly if $P_1\ne P_2$ and $\mu>0$.
Now take the paired Bernoulli model with marginals $(a,r)$ fixed and $r>\rstar$. Then $\KLinf$ is strictly increasing in $\gamma=\mathrm{Cov}(Y^L,Y^M)$. Pairing therefore certifies faster than an unpaired design (live outcome from an independent instance) iff $\gamma>0$, that is, when live and macro fail on the same instances.
\end{theorem}
\begin{proof}
For $c<1$, $z\mapsto\ln(1+cz)$ is concave and bounded, so $g_{P_1}(c)\ge g_{P_2}(c)$. Let $c\uparrow1$ and maximize.
For strictness, Strassen's theorem gives a coupling $Z_2=Z_1+\xi$ with $\E[\xi\mid Z_1]=0$ and $\xi\ne0$ with positive probability. At $c=c^*(P_2)$ the map $\ln(1+cz)$ is strictly concave where finite, so conditional Jensen is strict.
Raising $\gamma$ by $\Delta$ moves mass $\Delta$ from each of $D=\pm1$ to $D=0$. This changes $\E\phi(D)$ by $-\Delta[\phi(1)+\phi(-1)-2\phi(0)]\le0$ for convex $\phi$, and $Z$ is affine in $D$.
\end{proof}
At $a=r=0.9$ and $\eta=0.05$, pairing raises $\KLinf$ by $1.85\times$ at correlation $0.5$ and by $7.56\times$ at correlation $1$. Near break-even, it beats even the SPRT that \emph{knows} $a$ iff the correlation exceeds $\sqrt{a(1-a)}/(2\sqrt{\rstar(1-\rstar)})$, which is $0.42$ here.

\paragraph{The scaling law with $\KLinf$.}
\begin{corollary}[Coverage gap with $\KLinf$]\label{cor:app-scaling}
Assume (A1)--(A3) and (A5) of \cref{app:limit}, and replace (A4) by: in every epoch the normalized paired utilities $Z_t=U_t/B$ of the audited shadow evaluations are i.i.d.\ from a finitely supported law $P_Z$ on $[-1,M]$ with positive mean, whichever candidate they evaluate, and independent of $\mathcal G$. Let
\[
\Th_U\defeq\frac{\nu\,\ln(1/\delta)}{p\,\KLinf(P_Z)}.
\]
(i)~Every policy of type (O1) that is $\delta$-safe in the sense of \cref{def:safety}, with ``bad'' meaning $\E Z\le0$, has $\liminf\cov\ge(1-\eps_\delta)F_\pi(\Th_U)$ almost surely, with $\eps_\delta=O\bigl(\sqrt{\ln\ln(1/\delta)/\ln(1/\delta)}\bigr)$ depending only on $(\delta,P_Z)$. (ii)~With announced changes, the grid-mixture betting test of \cref{thm:app-rate}(a), restarted at each change with all earlier evidence discarded, is $\delta$-safe and has $\cov\le\bigl(1+(2\ln\ln\frac1\delta+O(1))/\ln\frac1\delta\bigr)F_\pi(\Th_U)$. Thus \cref{thm:limit} holds with $\KLinf(P_Z)$ in place of $\KL(r_1\|r_0)$, and the live accuracy $a$ need not be known.
\end{corollary}
\begin{proof}
(i) Repeat the converse proof of \cref{thm:limit} with three substitutions. \emph{The null environment.} Let $Q^*$ be the least-favorable null of \cref{lem:app-klinf}; it has mean $0$, so in the environment where the epoch's audited utilities are i.i.d.\ $Q^*$ every candidate is bad, and $\delta$-safety bounds the probability of a fold event in the epoch by $\delta$ given $\mathcal H_e$. \emph{The likelihood ratio.} By \cref{lem:app-klinf}, $dP_Z/dQ^*=1+c^*z>0$ on $\mathrm{supp}P_Z$, so $P_Z\ll Q^*$. Under (O1) the records add to the audited utilities only the audit coins and the internal randomization, whose laws do not depend on the environment, exactly as in the proof of \cref{thm:app-rate}(b). Hence $d\P_{P_Z}/d\P_{Q^*}$ restricted to $\mathcal F_m$ is $e^{W_m}$ with $W_m=\sum_{k\le m}\ell(Z_k)$ and $\ell(z)=\ln(1+c^*z)$, whichever candidates the policy forms or audits. Wald's likelihood-ratio identity then gives the low-evidence bound $\E_{P_Z}[R;W_N\le c,N<\infty]\le e^c\E_{Q^*}[R]$ as before. \emph{The drift.} Under $P_Z$ the increments $\ell(Z_k)$ are i.i.d., bounded (the support is finite and $1+c^*z>0$ on it) and have mean $g^*=\KLinf(P_Z)$. With $\nstar=\ln(1/\delta)/\KLinf(P_Z)$, $n-1=\lfloor(1-\eps)\nstar\rfloor$ and $c=(1-\eps/2)\ln(1/\delta)$, Azuma's maximal inequality gives $\beta_{n-1}(c)\to0$, and the rest of the proof (one epoch, many epochs, choice of $n$ and $c$) applies verbatim with $\Th$ replaced by $\Th_U$.
(ii) After each change the test restarts from wealth $1$ on the epoch's own evidence, so \cref{thm:app-valid} and Ville's inequality bound the probability of a fold of a null candidate in the epoch by $\delta$, which is \cref{def:safety}. The wealth moves only at audits (\cref{thm:app-rate}(a)), so the test folds at the $N_e$-th audited outcome of epoch $e$, where the $N_e$ are i.i.d., independent of arrivals, audits and changes, and have the law of $N_p$ in \cref{thm:app-rate}(a). \Cref{lem:exact} gives $u_i=1-\E\rho_i^{N}$, and \cref{lem:poisexp} gives $u_i\le1-\rho_i^{\E N}\le1-e^{-x\E N/\pi_i}$ with $x=\nu/p$. By \cref{thm:app-rate}(a), $x\E N\le(1+\eta_\delta)\Th_U$ with $\eta_\delta=(2\ln\ln\frac1\delta+O(1))/\ln\frac1\delta$. Since $F_\pi$ is concave and nondecreasing with $F_\pi(0)=0$, $F_\pi((1+\eta)\Th_U)\le(1+\eta)F_\pi(\Th_U)$; summing $\pi_iu_i$ gives the bound.
\end{proof}
Pairing, disagreement-triggered audits and cost-aware utilities thus all act through one constant, $\KLinf(P_Z)$, in the certification load. Under Zipf demand, a pairing gain $\rho_g=\KLinf^{\rm pair}/\KLinf^{\rm unpair}$ lowers the coverage gap by the factor $\rho_g^{1-1/s}$.

\subsection{Agreement-aware certification}\label{app:agree}
Shadow mode runs both paths, so the system sees both answers before any audit. When the answers agree, their correctness labels agree, provided scoring depends on the answer alone:
\begin{itemize}
\item[(E)] \emph{Exact-match scoring.} There are $G_t\in\{0,1\}$ and $z_{0,t}$, observed before the audit decision (functions of $W_t$), such that $G_t=1$ implies $Z_t=z_{0,t}$.
\end{itemize}
(E) holds if correctness is a deterministic function of the instance and the final answer, $Y=\phi_t(\text{answer})$ (exact match to a gold answer, or execution against fixed tests), if $G_t=1$ exactly when the two final answers are identical as inputs to $\phi_t$, and if both costs are observed; then $G_t=1$ forces $Y^M_t=Y^L_t$, and $z_{0,t}=(C^L_t-C^M_t)/B$. It fails for stochastic or rubric-based judges, for judges that read more than the final answer, and whenever a path's score depends on anything other than its answer. \textsc{KQA-Stream} is such a case: a macro answer counts as correct only if the query was also routed to the right template, so on a misrouted query whose macro answer equals a correct live answer, $G_t=1$ but $Y^M_t=0\ne Y^L_t=1$. The certificates of this subsection are therefore not valid for the \textsc{KQA-Stream} labels as defined; they are valid for a variant that scores a macro by its answer alone.

\begin{proposition}[Agreement breaks the (O1) bound]\label{prop:app-agree-example}
Assume (E) in the paired model with constant costs, so $z_0=\eta/(1-\eta)$. Let $V_t=z_0$ if $G_t=1$ and $V_t=-1$ otherwise. For every $\lambda\in[0,1]$, $\prod_{s\le t}(1+\lambda V_s)$ is a test supermartingale under $H_0$, for any audit mechanism. If $G_t$ is i.i.d.\ with $\P(G_t=1)=g$, the best constant bet grows the log-wealth by $\mathrm{kl}\bigl(g\,\|\,\frac1{1+z_0}\bigr)$ per shadow run if $g>\frac1{1+z_0}$, and by $0$ otherwise. If $g=1$, the bet $\lambda=1$ folds after $\lceil\ln(1/\delta)/\ln(1+z_0)\rceil$ shadow runs and uses no labels, whereas every (O1) rule needs $\E_P\tau\ge\ln(1/\delta)/(p\ln(1+z_0))$.
\end{proposition}
\begin{proof}
By (E), $V_t=Z_t$ if $G_t=1$, and $V_t=-1\le Z_t$ otherwise. So $V_t\le Z_t$ pointwise, and $\E[V_t\mid\mathcal F_{t-1}]\le\E[Z_t\mid\mathcal F_{t-1}]\le0$ whatever the audits do. The factors satisfy $1+\lambda V_t\ge1-\lambda\ge0$, and $V_t$ is observed without a label. For i.i.d.\ $G_t$, $V_t$ has the two-point law $g\delta_{z_0}+(1-g)\delta_{-1}$. By \cref{lem:app-klinf} the best constant bet grows at rate $\KLinf$ of this law, and its least-favorable null lives on $\{z_0,-1\}$. A null law $q\delta_{z_0}+(1-q)\delta_{-1}$ has mean $qz_0-(1-q)\le0$ iff $q\le1/(1+z_0)$, and $\mathrm{kl}(g\|q)$ decreases in $q$ on $[0,g]$, which gives the rate. If $g=1$, the wealth is $(1+z_0)^n$. For (O1) rules, $P_Z=\delta_{z_0}$ and $\KLinf(\delta_{z_0})=\ln(1+z_0)$, so \cref{thm:app-rate}(b) applies.
\end{proof}
At $\eta=\delta=0.05$ and $p=0.1$ the agreement bet folds after $59$ shadow runs, while (O1) rules need at least $584$ (the grid-mixture IPW certificate needs $910$). Under the least-favorable null $q=1/(1+z_0)$ with $Z=-1$ on disagreement, the bet $\lambda=1$ folds with probability $q^{59}=0.048\le\delta$. The rule is powerless unless agreement is nearly certain ($g>0.95$ here), because an unaudited disagreement is charged the worst case $-1$. The certificate below also bets on audited disagreements, with inverse-propensity weights.

\paragraph{The agreement-aware certificate.} Assume (E), (A1), (A2), and $p_t\ge\pi_t>0$ on $\{G_t=0\}$ (agreements need no audits). Let
\[
\mathcal C_t=\{(\lambda,u):\lambda\ge0,u\ge-1,\lambda+u\le\pi_t(1+u)\}.
\]
For $\mathcal F_{t-1}$-measurable $(\lambda_t,u_t)\in\mathcal C_t$, the \emph{agreement-aware factor} is
\[
F_t=\begin{cases}1+\lambda_tz_{0,t}, & G_t=1,\\ 1+u_t+A_t(\lambda_tZ_t-u_t)/p_t, & G_t=0.\end{cases}
\]
Equivalently, $F_t=1+\lambda_tX_t$ with $X_t=z_{0,t}$ on agreement and $X_t=v_t+A_t(Z_t-v_t)/p_t$ on disagreement, where $v_t=u_t/\lambda_t$ (for $\lambda_t>0$): an inverse-propensity estimator whose surrogate is exact on agreement and the constant $v_t$ on disagreement. The \emph{conservative} increment
\[
X^0_t=A_tZ_t+(1-A_t)\bigl[G_tz_{0,t}-(1-G_t)\bigr]
\]
(audited: $Z_t$; unaudited agreement: $z_{0,t}$; unaudited disagreement: $-1$, or any known lower bound on $Z_t$) is bet on with $\lambda\in[0,1]$ and no weights. As before, $K_n=\sum_kw_k\prod_{t\le n}F^{(k)}_t$ with $\sum_kw_k\le1$, and the certificate folds when $K_n\ge1/\delta$.

For the rate, take the i.i.d.\ model: $(G_t,Z_t)$ are i.i.d.\ with law $P$ satisfying (E) with a constant $z_0\ge0$, $g=P(G=1)$, $Z$ given $G=0$ has a finitely supported law $P_{\rm dis}$ on $[-1,M]$, and audits are i.i.d.\ Bernoulli$(p)$, independent of everything else. A law $Q$ of $(G,Z)$ that satisfies (E) with the same $z_0$ is \emph{structural}; write $q=Q(G=1)$ and $R$ for the law of $Z$ given $G=0$. Define
\[
D^*(P;p)=\inf\bigl\{\mathrm{kl}(g\|q)+p(1-g)\KL(P_{\rm dis}\|R)\bigr\},
\]
over structural $Q$ with $qz_0+(1-q)\E_RZ\le0$. The first term is the information in the agreement frequency, the second the information in the audited disagreements. Let $\Gamma(\lambda,u)=\E_P\ln F$ be the growth rate of the constant bet $(\lambda,u)\in\mathcal C$ (with $\pi_t=p$; for $p=1$, where $\Gamma$ does not depend on $u$, we also require $u\le0$, so that $\mathcal C$ is compact), and call $(\lambda,u)$ \emph{strictly feasible} if $\lambda>0$, $u>-1$ and $\lambda+u<p(1+u)$.

\begin{theorem}[Agreement-aware certificate]\label{thm:app-agree}
\emph{(i) Validity.} Under (E), (A1), (A2), $p_t\ge\pi_t>0$ on $\{G_t=0\}$ and $H_0$, $(K_n)$ is a test supermartingale, so the certificate folds with probability at most $\delta$. The conservative certificate is a test supermartingale under (E) and $H_0$ alone, for any audit mechanism, including outcome-dependent audits.

\emph{(ii) Information.} In the i.i.d.\ model, $D^*(P;p)=\max_{\mathcal C}\Gamma$; $D^*>0$ iff $\E_PZ>0$; some maximizer has $u^*\le0$; and
\begin{align*}
D^*\ge\max\Bigl\{&p\,\KLinf(P_Z),\ \KLinf(P_{X^0}),\\
&\mathrm{kl}\bigl(g\,\|\,\tfrac1{1+z_0}\bigr)\1_{\{g>\frac1{1+z_0}\}}\Bigr\}.
\end{align*}

\emph{(iii) Converse.} Consider a rule of type (O2) with $p_t(0)\le p$ that folds at a stopping time $\tau$ of its records. Suppose $\P_Q(\tau<\infty)\le\delta$ for every i.i.d.\ structural $Q$ with $\E_QZ\le0$, and $\P_P(\tau<\infty)=1$. Then $\E_P\tau\ge\ln(1/\delta)/D^*(P;p)$.

\emph{(iv) Achievability.} For any finite set of strictly feasible $(\lambda_k,u_k)$ with $u_k\le0$ and weights $w_k$, the mixture has
\[
\E_P\tau\le\min_{k:\Gamma_k>0}\frac{\ln\frac1\delta+\ln\frac1{w_k}+\ln\frac{1+M}p}{\Gamma(\lambda_k,u_k)}.
\]
With grids refined as $\delta\to0$ and $\ln(1/w_k)=o(\ln\frac1\delta)$, $\E_P\tau=(1+o(1))\ln(1/\delta)/D^*(P;p)$. The conservative certificate with the grid mixture of \cref{thm:app-rate}(a) (with $p=1$) has $\E_P\tau\le(\ln\frac1\delta+2\ln\ln\frac1\delta+O(1))/\KLinf(P_{X^0})$.
\end{theorem}
\begin{proof}
\emph{(i)} Nonnegativity. Since $\pi_t\le1$ and $1+u\ge0$, $\lambda\le\pi_t(1+u)-u\le1$, so $1+\lambda z_{0,t}\ge1-\lambda\ge0$ on agreement. On disagreement without audit, $F_t=1+u_t\ge0$. With audit, $Z_t\ge-1$ gives $F_t\ge1+u_t-(\lambda_t+u_t)/p_t$. This is $\ge1+u_t\ge0$ if $\lambda_t+u_t\le0$; otherwise it increases in $p_t$ and is $\ge1+u_t-(\lambda_t+u_t)/\pi_t\ge0$ because $(\lambda_t,u_t)\in\mathcal C_t$.
Mean. On $\{G_t=1\}$, (E) gives $F_t=1+\lambda_tZ_t$. On $\{G_t=0\}$, (A2) and $p_t>0$ give $\E[F_t\mid\mathcal H_t,Z_t]=1+u_t+\lambda_tZ_t-u_t=1+\lambda_tZ_t$. Since $\lambda_t,u_t$ are $\mathcal F_{t-1}$-measurable and $\lambda_t\ge0$, $\E[F_t\mid\mathcal F_{t-1}]=1+\lambda_t\E[Z_t\mid\mathcal F_{t-1}]\le1$. The mixture and Ville's inequality follow as in \cref{thm:app-valid}.
For the conservative certificate, $X^0_t=Z_t$ if $A_t=1$, $X^0_t=Z_t$ if $A_t=0$ and $G_t=1$ (by (E)), and $X^0_t=-1\le Z_t$ otherwise. So $X^0_t\le Z_t$ pointwise, whatever the audits depend on, $X^0_t$ is observable, and $1+\lambda X^0_t\ge1-\lambda\ge0$.

\emph{(ii)} Let $O=(G,A,AZ)$. Because $A$ is independent Bernoulli$(p)$ under $P$ and under every structural $Q$, and $AZ=Az_0$ on $\{G=1\}$, the chain rule gives $\KL(P_O\|Q_O)=\mathrm{kl}(g\|q)+p(1-g)\KL(P_{\rm dis}\|R)$. So $D^*$ is the infimum of $\KL(P_O\|Q_O)$ over structural nulls.
\emph{Weak duality.} The computation in (i) with $p_t=p$ gives $\E_QF=1+\lambda\E_QZ\le1$ for every $(\lambda,u)\in\mathcal C$ and structural null $Q$. As in \cref{lem:app-klinf}, Jensen then gives $\Gamma(\lambda,u)=\E_P\ln F\le\KL(P_O\|Q_O)$. Hence $\sup_{\mathcal C}\Gamma\le D^*$.
If $\E_PZ\le0$, $P$ is itself a structural null, so $D^*=0=\Gamma(0,0)$. Let $\E_PZ>0$. The set $\mathcal C$ is a compact polygon, and $\Gamma$ is concave and upper semicontinuous with values in $[-\infty,\infty)$, so it has a maximizer $(\lambda^*,u^*)$. For $\lambda\le p$, $(\lambda,0)\in\mathcal C$ and $\frac{d}{d\lambda}\Gamma(\lambda,0)|_{\lambda=0}=\E_PZ>0$, so $\Gamma^*\defeq\Gamma(\lambda^*,u^*)>0$ and $\lambda^*>0$.
\emph{Case $g=1$.} Then $\Gamma=\ln(1+\lambda z_0)$, which is maximal at $(1,-1)\in\mathcal C$ with value $\ln(1+z_0)$. The structural law with $q=1/(1+z_0)$ and $R=\delta_{-1}$ has mean $0$ and $\KL(P_O\|Q_O)=\ln(1+z_0)$, so $D^*=\Gamma^*$ and $u^*=-1$.
\emph{Case $g<1$.} Since $\Gamma^*$ is finite and $P$ has finite support, all factors are positive at $(\lambda^*,u^*)$ on the support of $P_O$, and $\Gamma$ is differentiable nearby, with $F_d(z)\defeq1+u+(\lambda z-u)/p$,
\begin{align*}
\partial_\lambda\Gamma&=\frac{gz_0}{1+\lambda z_0}+(1-g)\,\E_{P_{\rm dis}}\frac Z{F_d},\\
\partial_u\Gamma&=(1-g)(1-p)\Bigl[\frac1{1+u}-\E_{P_{\rm dis}}\frac1{F_d}\Bigr].
\end{align*}
The constraint $\lambda\ge0$ is inactive. Write the others as $c_1=1+u\ge0$ and $c_2=p(1+u)-u-\lambda\ge0$. The KKT conditions for a concave function on a polytope give $\nu_1,\nu_2\ge0$ with $\nu_ic_i=0$, $\partial_\lambda\Gamma=\nu_2$ and $\partial_u\Gamma=\nu_2(1-p)-\nu_1$. If $p<1$, $\Gamma=-\infty$ at $u=-1$, so $\nu_1=0$. If $p=1$, $\partial_u\Gamma\equiv0$; take the maximizer with $u^*=0$, where $c_1>0$, so $\nu_1=0$ and the constraint $u\le0$ gets multiplier $0$. If $\nu_2>0$, then $c_2=0$, so $F_d(-1)=0$ and $P_{\rm dis}(\{-1\})=0$.
Define a measure $Q^*_O$ on the values of $O$: mass $g/(1+\lambda^*z_0)$ on agreement, $(1-g)(1-p)/(1+u^*)$ on an unaudited disagreement, and $(1-g)pP_{\rm dis}(z)/F_d(z)+p\nu_2\1_{\{z=-1\}}$ on an audited disagreement with $Z=z$. On the support of $P_O$, $dP_O/dQ^*_O=F^*$, and the added atom lies outside it.
\emph{Mass.} Since $F-1$ is linear in $(\lambda,u)$, $\lambda\partial_\lambda\Gamma+u\partial_u\Gamma=\E_P[1-1/F]=1-(Q^*_O(\text{all})-p\nu_2)$. By the KKT conditions the left side equals $\nu_2(\lambda^*+u^*(1-p))$, and this is $p\nu_2$ because $\nu_2>0$ forces $\lambda^*=p(1+u^*)-u^*$. So $Q^*_O$ has mass $1$.
\emph{Structure.} For $p<1$, the $u$-condition with $\nu_1=0$ reads $(1-g)/(1+u^*)-(1-g)\E_{P_{\rm dis}}[1/F_d]=\nu_2$, that is, $Q^*_O(\text{unaudited disagreement})/(1-p)=Q^*_O(\text{audited disagreement})/p\eqqcolon1-q^*$. With the mass identity, the agreement mass is $q^*$. So $Q^*_O$ is the observation law (the audit coin is Bernoulli$(p)$ under both laws) of the structural $Q^*$ with $Q^*(G=1)=q^*$ and $R^*=Q^*_O(\text{audited disagreement},\cdot)/(p(1-q^*))$. For $p=1$ there are no unaudited disagreements, and $1-q^*$ is the audited disagreement mass.
\emph{Mean.} $\E_{Q^*}Z=q^*z_0+(1-q^*)\E_{R^*}Z=\partial_\lambda\Gamma-\nu_2=0$.
So $Q^*$ is a structural null and $D^*\le\KL(P_O\|Q^*_O)=\E_P\ln F^*=\Gamma^*\le D^*$.
\emph{Sign of $u^*$.} For $p<1$, the structure identity gives $1+u^*=(1-g)/(1-q^*)$, and $q^*=g/(1+\lambda^*z_0)\le g$ because $z_0\ge0$. Hence $u^*\le0$. For $p=1$, $u$ does not enter $\Gamma$, and $u^*=0$ is a maximizer.
\emph{Lower bounds.} Under a structural null $Q$, the pair $(A,AZ)$ is a function of $O$ with law Bernoulli$(p)\otimes Q_Z$ and $\E_QZ\le0$, so data processing gives $\KL(P_O\|Q_O)\ge p\KL(P_Z\|Q_Z)\ge p\KLinf(P_Z)$. Likewise $X^0$ is a function of $O$ with $\E_QX^0\le\E_QZ\le0$, which gives $\KLinf(P_{X^0})$. Finally $G$ is a function of $O$, and $\E_RZ\ge-1$ forces $q\le1/(1+z_0)$, which gives the $\mathrm{kl}$ term. Take the infimum over $Q$.

\emph{(iii)} Let $Q^*$ be the structural null from (ii). The records after $n$ runs are $(U,(G_t,A_t,A_tZ_t)_{t\le n})$. Their density under an i.i.d.\ structural law is a product of $Q(G_t)$, of the audit factors $p_t(G_t)^{A_t}(1-p_t(G_t))^{1-A_t}$, and of the density of $R$ at $Z_t$ when $A_t=1$ and $G_t=0$; on $\{G_t=1\}$, $A_tZ_t=A_tz_0$ carries no further information. The audit factors are the same under $P$ and $Q^*$. The log-likelihood ratio is therefore $\sum_{t\le n}\ell_t$ with
\[
\ell_t=\ln\frac{P(G_t)}{Q^*(G_t)}+A_t(1-G_t)\ln\frac{dP_{\rm dis}}{dR^*}(Z_t),
\]
which is bounded because the supports are finite and $R^*>0$ on $\mathrm{supp}P_{\rm dis}$. Its conditional mean under $P$ given the records before $t$ is $\mathrm{kl}(g\|q^*)+(1-g)p_t(0)\KL(P_{\rm dis}\|R^*)\le D^*$. Optional stopping at $\tau\wedge n$ and data processing on $\{\tau\le n\}$ give $D^*\E_P\tau\ge x_n\ln(1/\delta)-h(x_n)$ with $x_n=\P_P(\tau\le n)\to1$, as in \cref{thm:app-rate}(b).

\emph{(iv)} For a strictly feasible $(\lambda_k,u_k)$ with $u_k\le0$, the factors lie between a positive constant and $(1+M)/p$, since $1+u_k(1-1/p)+\lambda_kM/p\le(1+M)/p$. The log-wealth $S_n$ of this bet has i.i.d.\ bounded increments of mean $\Gamma_k>0$. Since $K_n\ge w_ke^{S_n}$, $\tau$ is at most the first $n$ with $S_n\ge\ln(1/(\delta w_k))$, and Wald's identity with overshoot at most $\ln((1+M)/p)$ gives the bound. For the limit, take $\theta\in(0,1)$ and a maximizer with $u^*\le0$. The point $\theta(\lambda^*,u^*)$ has $\lambda=\theta\lambda^*>0$, $u=\theta u^*\in(-1,0]$ and $c_2=(1-\theta)p+\theta c_2(\lambda^*,u^*)\ge(1-\theta)p>0$, so it is strictly feasible, and $\Gamma(\theta\lambda^*,\theta u^*)\ge\theta\Gamma^*+(1-\theta)\Gamma(0,0)=\theta\Gamma^*$ by concavity. $\Gamma$ is continuous on the strictly feasible set, so fine grids contain points with $\Gamma\ge\theta^2\Gamma^*$; let $\theta\to1$. The conservative certificate is \cref{thm:app-rate}(a) applied to the i.i.d.\ $X^0_t\in[-1,M]$ with $p=1$.
\end{proof}

\paragraph{Implementation.} A replay policy implements the certificate as follows. The agreement indicator is $G_t=1$ iff both final answers are non-empty and identical under the scorer's own comparison, and (E) must hold for the benchmark's labels. On a shadow run with live cost $c_t$, put $z_{0,t}=(c_t-1)/B$ (clipped to $[-1,1]$ as for $Z_t$), and on an audited disagreement compute $Z_t$ as in the IPW certificate. The bets are $\lambda=c(\pi(1-s)+s)$ and $u=-s$, where $\pi$ is the audit floor, $c$ runs over the grid $c_j=(1+\eps)^{-j}$ of \cref{thm:app-rate}(a) with weights $1/(j(j+1))$ (truncated at $c_j\ge1/200$), and $s$ runs over $\{0,0.01,0.03,0.1,0.3,0.6,0.9,0.97,0.99\}$ with weight $1/2$ for $s=0$ and $1/16$ for each other value. Every grid point is strictly feasible, because $\lambda+u-\pi(1+u)=-(1-c)(\pi(1-s)+s)<0$. The factors are $1+\lambda z_{0,t}$ on agreement (audited or not), $1-s$ on an unaudited disagreement, and $1-s+(\lambda Z_t+s)/p_t$ on an audited disagreement. The wealth is updated on every shadow run, not only on audited ones. The slice $s=0$ is the IPW certificate with the exact surrogate $z_{0,t}$ on agreement. By concavity, $g\ln(1+cpz_0)\ge pg\ln(1+cz_0)$, so the growth of its bet $c$ is at least that of the IPW bet $c$. The bound of \cref{thm:app-rate}(a) therefore holds for the mixture with an extra $\ln2$ in the numerator.

\paragraph{Numerical checks.} On 274 random structural laws ($p\in\{0.05,0.1,0.3,0.7,1\}$), $\max_{\mathcal C}\Gamma$ and a direct convex program over structural nulls agree to $3\cdot10^{-12}$. The constructed $Q^*$ has mass $1$, mean $0$ and the structural ratio to within $5\cdot10^{-8}$, $u^*\le10^{-8}$, and $D^*$ never falls below the three lower bounds in (ii). The median of $D^*/(p\KLinf(P_Z))$ is only $1.02$: agreement adds little information unless it is frequent. In the paired model with $p=0.1$ the ratio is $1.00$ for live accuracy $0.38$ and macro reliability $0.97$ (as on \textsc{KQA-Stream}), $1.18$ for $a=r=0.9$ with correlated errors, and $5.2$ when the answers agree with probability $0.99$. In $200$ simulated runs per setting, the grid above certified in $1.07$, $1.03$ and $0.29$ times the time of the IPW grid mixture at $\delta=10^{-2}$, and in $1.02$, $0.88$ and $0.23$ times at $\delta=10^{-6}$. The conservative certificate has rate $0$ in the first two settings and $0.027$ (against $D^*=0.030$) in the third. Under five structural nulls, including the least-favorable one, the certificate folded in $2.4$--$3.3\%$ of $4{,}000$ runs at $\delta=0.05$ (horizon $4{,}000$).

\subsection{Changes, carried-over evidence and the lifetime ledger}\label{app:ledger}
\Cref{def:safety} counts every fold event of a bad candidate in an epoch, including candidates formed in earlier epochs. We separate what is proved for the \emph{idealized} certificate from what holds for the two implemented ledgers: without restarts and with restarts (the restart-only certificate of \cref{sec:certify}).

\paragraph{Idealized.} Changes are announced, all certification evidence is discarded at each change, and one test per template (SPRT, or the grid mixture of \cref{thm:app-rate}) restarts in each epoch at level $\delta$; if several candidates are tested within an epoch, they share $\delta$, e.g.\ through a per-epoch ledger. Every fold in an epoch then rests on that epoch's evidence alone, Ville's inequality gives \cref{def:safety} for every fold event, and \cref{thm:limit}(ii) and \cref{cor:app-scaling} describe this certificate.

\paragraph{Ledger with restarts (restart only).} The replay notifies the policy of every edit that changes the gold answer of some instance routed to a template; epochs of a template are the intervals between such edits, within which the macro's reliability is constant. At each notification, a candidate under test discards its wealth and its shadow statistics and restarts its test with the next ledger index, so the lifetime ledger still sums to $\delta$. Folded macros are not reset; they are monitored by the LR-CUSUM. By \cref{prop:app-ledger}(iv) this certificate satisfies \cref{def:safety} for these epochs \emph{if its own null is constant within them}. The null $\E Z\le0$ also depends on live reasoning, whose accuracy and cost can change without any answer change; across such a silent change only \cref{prop:app-ledger}(iii) applies (\cref{app:r4-silent} measures how often this happens and evaluates a variant that also restarts on live-trace changes). It needs to know which templates an edit touches; a coarser change log that also flags edits without effect on a template's answers triggers extra restarts, which cost coverage but not safety, because every restart consumes a fresh index.

\paragraph{Ledger without restarts.} Here the policy is not told when an edit occurs. A candidate keeps its wealth (or its SPRT statistic) across edits; it is neither reset nor discarded. The $j$-th candidate of a template, counted over the whole stream, is tested at $\delta_j=6\delta/(\pi^2j^2)$ with seven fixed bets. For a candidate $j$ and its shadow evaluations $t=1,2,\dots$, let $\zeta_j$ be the first $t$ with $\E[Z_t\mid\mathcal F_{t-1}]>0$ ($\zeta_j=\infty$ if there is none). Call a fold of candidate $j$ at evaluation $t<\zeta_j$ a fold \emph{on null evidence}: the candidate has been at or below break-even at every evaluation so far.

\begin{proposition}[What the lifetime ledger guarantees]\label{prop:app-ledger}
Let each candidate's wealth be a mixture of test supermartingales, as in \cref{thm:app-valid}, started at $1$ at its formation and compared with $1/\delta_j$.
\begin{itemize}
\item[(i)] \emph{Lifetime.} With probability at least $1-\delta$, over the whole stream no candidate of the template folds on null evidence. The expected number of such folds per template is at most $\delta$.
\item[(ii)] \emph{Fresh candidates.} Given the history at the start of an epoch, the candidates formed from then on fold on null evidence with total probability at most $\sum_{j>J}\delta_j\le\delta$, where $J$ is the number of earlier candidates.
\item[(iii)] \emph{Carried-over candidates.} Let candidate $j$ be unfolded at a stopping time $c$ with wealth $K_c$, and let $e\ge c$ be a stopping time such that the null holds at its evaluations in $[c,e)$. Then it folds in $[c,e)$ with conditional probability at most $\min\{1,\delta_jK_c\}$ given $\mathcal F_c$. Equality holds (with $e=\infty$) if, after $c$, its wealth is a martingale that tends to $0$ unless it reaches $1/\delta_j$, and it reaches $1/\delta_j$ without overshoot.
\item[(iv)] \emph{Per epoch.} If the library holds at most one candidate per template, the probability that an epoch contains a fold of a candidate that is bad in that epoch is at most $\sum_{j>J}\delta_j+\delta_JK_c\le\delta+\delta_J(K_c-1)^+$, where candidate $J$ is carried over with wealth $K_c$ at the epoch start ($\delta_JK_c\defeq0$ if there is none, in particular if $J=0$). The ledger satisfies \cref{def:safety} if, at each change, carried-over candidates are discarded or restart their wealth at $1$ with the next ledger index (a restarted test then counts as a fresh candidate in (ii)); otherwise it does not in general.
\end{itemize}
\end{proposition}
\begin{proof}
(i) Fix $j$ and condition on the history at the formation of candidate $j$. Since $\E[Z_t\mid\mathcal F_{t-1}]$ is $\mathcal F_{t-1}$-measurable, $\{\zeta_j>t\}\in\mathcal F_{t-1}$. Let $\tilde K_t$ be the wealth with factors replaced by $1$ from evaluation $\zeta_j$ on. Then $\E[\tilde K_t\mid\mathcal F_{t-1}]=\tilde K_{t-1}(\1_{\{\zeta_j>t\}}\E[F_t\mid\mathcal F_{t-1}]+\1_{\{\zeta_j\le t\}})\le\tilde K_{t-1}$, because $\E[F_t\mid\mathcal F_{t-1}]\le1$ whenever the null holds at $t$ (\cref{thm:app-valid}). A fold on null evidence implies $\sup_t\tilde K_t\ge1/\delta_j$, which has conditional probability at most $\delta_j$ by Ville's inequality. Sum over $j$; $\sum_j\delta_j=\delta$. Each candidate folds at most once, which gives the expectation bound.
(ii) The same argument, conditioned on the history at the epoch start, for $j>J$.
(iii) Stop the factors at $e$ and at the first non-null evaluation, as in (i). Given $\mathcal F_c$, the stopped wealth $(K_{t\wedge e})_{t\ge c}$ is a nonnegative supermartingale started at $K_c$; apply Ville's inequality to $K_{t\wedge e}/K_c$. In the equality case, the martingale stopped at $1/\delta_j$ is bounded by $1/\delta_j$, so bounded convergence gives $K_c=\E[K_\infty\mid\mathcal F_c]=\delta_j^{-1}\P(\text{fold after }c\mid\mathcal F_c)$.
(iv) A fold in the epoch of a candidate that is bad in the epoch is either a fold on null evidence of a candidate formed in the epoch, covered by (ii), or a fold of the carried-over candidate, covered by (iii) with $c$ the epoch start and $e$ its end. Previously folded macros cause no fold event. Add the bounds and use $\sum_{j\ge J}\delta_j\le\delta$.
\end{proof}
In the paired model the utility null $\E Z\le0$ is the reliability condition $r\le\rstar$, so ``bad'' in \cref{def:safety} (with $r_0=\rstar$) and ``null'' coincide, up to the online estimates of $L$ and $m$ that the certificate uses; since $\rstar=a-(m-1)/L$ moves with live reasoning, both can change inside an epoch of answer changes.

\paragraph{Numerical check of (iii).} We simulated a candidate with the deployed seven-bet mixture ($p=0.1$, $j=1$, $\delta=0.05$) that sees a good utility law for $60$ shadow runs and then a null law with mean exactly $0$. Of the candidates still unfolded at the change (mean wealth $4.4$), $11.1\%$ folded after it, against the bound $\E\min(1,\delta_1K_c)=13.4\%$ and $\delta=5\%$. After $120$ good runs the numbers are $22.8\%$ and $28.1\%$. So carried-over evidence can defeat per-epoch safety, as (iv) allows.

\paragraph{Restarts without change information.} The restarts above use the edit log. The next result shows when per-epoch safety is out of reach without change information; the one after it gives the guarantee of restarts triggered by the memory's own change detector, which needs none.

Under (A1)--(A3) of \cref{app:limit}, audits of template $i$ form a Poisson process of rate $\lambda_i\defeq p\pi_i$, independent of the changes. Let
\[
\alpha_i\defeq\frac{\lambda_i}{\lambda_i+\nu}
\]
be the probability that an epoch of template $i$ contains at least one audit. By memorylessness this holds given any history at the epoch start, and the number of audits in an epoch is geometric, $\P(K\ge k)=\alpha_i^k$. In the notation of \cref{lem:exact}, $\alpha_i=\rho_i$.

\begin{proposition}[Per-epoch safety without change information]\label{prop:app-nonotice-lb}
Let the audited outcomes of an epoch be i.i.d.\ from the epoch's law, whichever candidate they evaluate, and let a policy observe neither change times nor any signal of them, and fold only at audited outcomes of template $i$.
(i)~\emph{Upper bound.} For every epoch and every history before it, the probability that the epoch contains a fold of template $i$ is at most $\alpha_i$. So if $\alpha_i\le\delta$, \cref{def:safety} holds for template $i$ whatever the policy.
(ii)~\emph{Lower bound.} Suppose that, with positive probability, the policy reaches a history $h$ after which its next audited outcome of template $i$, whenever it arrives, folds a candidate if it lies in a set $F$. Then for every bad law $Q$, given that an epoch starts after $h$ and before that outcome, the probability that this epoch contains a fold of a bad candidate is at least $\alpha_iQ(F)$, and at least $Q(F)$ given in addition that the epoch contains an audit. The policy is therefore not $\delta$-safe if $\alpha_iQ(F)>\delta$.
Consequently, without change information, among certificates whose fold decisions for template $i$ are deterministic functions of its own audited outcomes,
(a)~every reliability certificate that folds with positive probability violates \cref{def:safety} at every level $\delta<\alpha_ir_0$;
(b)~every paired-utility certificate with $Z$ taking finitely many values in $[-1,M]$ that folds with positive probability violates it at every level $\delta<\alpha_i/(1+M)$.
\end{proposition}
\begin{proof}
(i) A fold of template $i$ in the epoch requires an audit of template $i$ in the epoch. Given the history at the epoch start $s$, the audits after $s$ form a Poisson process of rate $\lambda_i$ and the epoch length is exponential with rate $\nu$, independent of each other and of the history ((A1)--(A3)), so the first audit after $s$ precedes the epoch end with probability $\lambda_i/(\lambda_i+\nu)=\alpha_i$.
(ii) The policy's action at the next audited outcome is the same function of $h$ and of that outcome whether or not a change occurred, because it observes neither. Changes are the points of a Poisson process independent of the history, so an epoch starts between $h$ and the next audited outcome with positive probability, and \cref{def:safety} must hold for the histories at the start of such epochs; these histories consist of $h$ and the elapsed time. Given such a history, the next audited outcome is the first audit of the epoch, and it falls in the epoch with probability $\alpha_i$, by the argument of (i). Under $Q$ every candidate is bad in the epoch, and this outcome lies in $F$ with probability $Q(F)$, independently of its arrival time and of the history; on the intersection of the two events a bad candidate folds in the epoch.
For (a) and (b), outcomes take finitely many values, so some history $h$ just before the policy's first fold has positive probability; take $F$ the set of outcomes on which the policy folds after $h$, which is not empty and, by determinism, does not depend on when the next audit arrives. (a)~If $1\in F$, take $Q=\mathrm{Bernoulli}(r_0)$, so $Q(F)\ge r_0$; otherwise $0\in F$ and $Q=\mathrm{Bernoulli}(0)$ gives $Q(F)=1$. (b)~Take $z\in F$. If $z\le0$, the point mass at $z$ is bad and $Q(F)=1$. If $z>0$, the law with mass $1/(1+z)$ at $z$ and $z/(1+z)$ at $-1$ has mean $0$, so it is bad, and $Q(F)\ge1/(1+z)\ge1/(1+M)$.
\end{proof}
So, for a reliability certificate without change information, per-epoch safety is impossible on every template that is likely enough to be audited in an epoch, $\alpha_ir_0>\delta$, i.e.\ with more than $\delta/(r_0-\delta)$ audits per epoch on average, and automatic on every template with $\alpha_i\le\delta$; in between it depends on the rule. With $M=1$ (our clipped utilities) or $r_0=\tfrac12$, and $\delta=0.05$, the impossibility covers every template with more than one audit per nine epochs on average. The templates it leaves out are barely coverable even with change information: by \cref{lem:exact}, the restarted SPRT of \cref{thm:limit}(ii), which needs at least $n_{\min}=\lceil\ln(1/\delta)/\ln(r_1/r_0)\rceil\ge1$ audits in an epoch to fold, answers in one step a share at most $\E\rho_i^N\le\alpha_i^{n_{\min}}\le(\delta/r_0)^{n_{\min}}$ of the queries of a template with $\alpha_ir_0\le\delta$. What remains attainable without change information is safety for every fold that rests on evidence collected after the change (\cref{prop:app-detect}).

\paragraph{Exact value for a stationary rule.} The lower bound of (ii) is attained by a rule whose only chance to fold is the next audited outcome. A rule that, from the epoch start on, folds at the first audited outcome in $F$, whenever it comes, has an exact bad-fold probability: outcomes are i.i.d.\ within the epoch and independent of the audit times, so the audits whose outcome lies in $F$ form a Poisson process of rate $\lambda_iQ(F)$, and the first of them precedes the epoch end with probability
\[
\beta_i(F)=\frac{\lambda_iQ(F)}{\lambda_iQ(F)+\nu}\in\bigl[\alpha_iQ(F),\,\alpha_i\bigr].
\]
For fold-at-first-success under the bad law $\mathrm{Bernoulli}(r_0)$ this is $\lambda_ir_0/(\lambda_ir_0+\nu)$, so this rule is $\delta$-safe if and only if $\lambda_i/\nu\le\delta/(r_0(1-\delta))$, i.e.\ $0.0658$ at $r_0=0.8$ and $\delta=0.05$, slightly below the threshold $\delta/(r_0-\delta)=0.0667$ beyond which (a) excludes every rule. The two bounds of the proposition bracket the exact value, and (a) cannot be sharpened to $\beta_i$ for every rule, because a rule may stop folding after an audited failure.

\paragraph{Numerical check.} At $\lambda_i=0.01$, $\nu=1$, $\delta=0.05$ and $r_0=0.8$, $\alpha_i=0.0099<\delta$, so every audit-driven rule is $\delta$-safe there; the next-audit rule folds a bad candidate with probability $\alpha_ir_0=0.0079$, and fold-at-first-success with $\lambda_ir_0/(\lambda_ir_0+\nu)=0.0079$ (they differ in the fifth decimal). Monte Carlo over $2\times10^6$ epochs per setting (Poisson audits and changes, bad law $\mathrm{Bernoulli}(r_0)$) reproduces the exact value of fold-at-first-success at $49$ settings with $r_0\in\{0.5,0.8,0.95,1\}$ and $\lambda_i/\nu$ from $10^{-3}$ to $100$: $47$ of the $49$ 95\% intervals contain it, the largest relative deviation is $2.1\%$, and an event-driven simulator ($2\times10^5$ epochs) agrees on all $10$ rows it was run on. Every exact value lies between $\alpha_ir_0$ and $\alpha_i$, with equality to $\alpha_i$ at $r_0=1$, and the $\delta$-safety boundary falls where predicted: at $r_0=0.8$ the probability is $0.0234$ at $\lambda_i/\nu=0.03$ and $0.0500$ at $0.0658$. For the next-audit rule, Monte Carlo reproduces $\alpha_ir_0$ to within $3\cdot10^{-4}$ at five settings with $\alpha_i$ from $0.0099$ to $0.99$. For an SPRT without restarts ($r_1=0.95$, $r_0=0.5$, $\delta=0.05$), bad epochs that start one success from the threshold contain a bad fold with frequency $\alpha_ir_0$ up to Monte Carlo error (more for $\alpha_i\ge0.9$, through later audits), which exceeds $\delta$ from $\alpha_i=0.2$ on; in a stream of epochs the frequency stays below $\alpha_i$ at every tested $\alpha_i\in[0.02,0.98]$ (\texttt{checks/prop24\_check.py} in our code release).

\begin{proposition}[Detector-triggered restarts]\label{prop:app-detect}
Let the library hold at most one candidate per template, and let each candidate restart its test (wealth $1$, the next ledger index) at stopping times of the system's own records, for instance at the alarms of an LR-CUSUM on its audited shadow outcomes, or be discarded then. Call the interval from a candidate's last restart (or its formation) to its decision its \emph{test segment}. Then:
(i)~statement (i) of \cref{prop:app-ledger} holds;
(ii)~for every epoch and every history before it, the probability that the epoch contains a fold of a candidate that is bad in the epoch and whose test segment began in the epoch is at most $\sum_{j>J}\delta_j\le\delta$, where $J$ is the number of ledger indices used before the epoch;
(iii)~the segment running at the epoch start, with index $J$ and wealth $K_c$, folds a candidate that is bad in the epoch before its next restart with probability at most $\min\{1,\delta_JK_c\}$.
\end{proposition}
\begin{proof}
Restart times are stopping times and every restart starts a fresh mixture of test supermartingales at $1$ with a fresh index, so each test segment is a candidate in the sense of \cref{prop:app-ledger}. Apply (i) of that proposition to all segments, (ii) to the segments that begin in the epoch, and (iii) to the segment running at its start, with $e$ the earlier of its next restart and the epoch end.
\end{proof}
A restart costs coverage, never safety, because it consumes a fresh index. The carried-over risk of (iii) lasts until the detector's next alarm, so the detector bounds its duration by its detection delay when the change raises the candidate's failure rate. A change that leaves the macro's outcomes unchanged but improves live reasoning is invisible to a detector on macro outcomes; it can make a candidate useless without making it wrong. \Cref{app:acl2-nonotice} measures both effects on \textsc{KQA-Stream}. In a simulation of discard-on-alarm with the lifetime ledger (Bernoulli likelihood-ratio wealth, $r_1=0.95$, bad law at $r_0=0.5$, $\delta=0.05$, $\alpha_i\in\{0.5,0.9,0.97\}$), fresh segments fold in at most $0.0003$ of bad epochs (mean bound $\sum_{j>J}\delta_j=0.001$--$0.009$), and carried-over segments in $0.009$--$0.014$ (mean bound $\E\min\{1,\delta_JK_c\}=0.015$--$0.052$), as (ii) and (iii) require.

\paragraph{What the restart-only certificate attains.} For the $j$-th test of a template (restarts included) the threshold is $\ln(1/\delta_j)=\ln(1/\delta)+2\ln j+\ln(\pi^2/6)$. With i.i.d.\ audited utilities of law $P$ and the seven bets $\mathcal B$ with weights $1/7$, the best bet in $\mathcal B$ and Wald's identity (factors in $[0.1,1.9]$ because the code clips $Z$ to $[-1,1]$) give an expected label count of at most
\[
\frac{\ln\frac1\delta+2\ln j+\ln\frac{\pi^2}6+\ln7+\ln1.9}{g_{\mathcal B}(P)}
\]
before the fold threshold is reached (ignoring the futility test), where $g_{\mathcal B}(P)=\max_{b\in\mathcal B}\E_P\ln(1+bZ)$ is the growth of the best bet, and $g_{\mathcal B}\le\KLinf(P)$, with equality only if $c^*\in\mathcal B$. The restart-only certificate therefore does not attain the limit of \cref{thm:limit} or \cref{cor:app-scaling} as $\delta\to0$. It is a finite-candidate, finite-grid version of the idealized certificate, whose extra cost is the additive $2\ln j+\ln(\pi^2/6)+\ln 7+\ln1.9$ and the factor $\KLinf/g_{\mathcal B}$ ($1.03$--$1.07$ on the three paired laws we checked).

\section{Experimental details}
\label{app:experiments}

\paragraph{Compute and software.} All LLM runs used one NVIDIA RTX~4080 SUPER GPU (16\,GB). Models were served with vLLM~0.30 \citep{kwon2023efficient} (FlashAttention backend) and PyTorch~2.13 under Ubuntu~22.04. The 8B models were served with FP8 weights: Qwen3-8B-FP8 natively \citep{yang2025qwen3}, and Granite-3.3-8B-Instruct \citep{ibm2025granite}, Mistral-7B-Instruct-v0.3 \citep{jiang2023mistral} and OLMo-3-7B-Instruct \citep{olmo2025olmo3} with on-the-fly FP8 quantization. Qwen3-4B ran in BF16. The natural-edit timeline used $4.75$ GPU-hours and the code domain $17.2$ GPU-hours of wall time. Mistral has no system role, so its system message is prepended to the first user turn. Granite-3.3's chat template embeds the current date, so its prompts change with the calendar day; all Granite prompts were rendered with the date pinned to 2026-09-25, and prompt hashes confirm that this reproduces the main-run prompts exactly. Qwen3 ``thinking'' was disabled. Main-table decoding used $T{=}0.7$, top-$p$ $0.8$, top-$k$ $20$, with a deterministic per-request seed. Every LLM request carries a content hash. Results are appended to checkpoint shards and skipped on restart, so all runs are resumable. Tables and figures are generated by script from the raw logs.

\paragraph{Mechanism computations.} Run lengths of integer-score rules were computed by solving the Markov chain of the reflected walk. The gap of SPRT certification uses \cref{lem:exact} with the exact sample-size distribution of the one-sided SPRT, obtained by dynamic programming on the lattice of log-likelihood-ratio values. Sums over templates run to $10^7$ with rigorous tail brackets. The collapse grid crossed $(r_0,r_1)\in\{(0.85,0.95),(0.6,0.9),(0.3,0.7)\}$ plus random pairs, verifier flip rates $\{0,0.05,0.1,0.2\}$, $\delta\in[10^{-32},0.03]$, audit rates $p\in[0.03,1]$ (tabulated settings: $p=1$, $\Th\in\{10^{-5},\dots,10^{-1}\}$), 10 drift rates for the random settings and $s\in\{1.2,1.5,2,3\}$. Plotted against $x=\nu/p$ alone, the same points spread over two orders of magnitude at fixed $x$ because of the $\ln(1/\delta)/\KL$ factor. The literal-stream check simulated Poisson arrivals, audits with $p\in\{0.2,0.3,0.5,1\}$, Poisson changes and verifier flips, for $500$--$10^4$ templates, and matched the exact formula within $0.51\%$.

\paragraph{Natural template streams.} KQA Pro templates are the programs of the train and validation sets (106,173 questions) with entity names, strings, numbers and dates abstracted. Function, relation, attribute and concept names are kept, giving 45,544 templates. MuSiQue templates abstract each single-hop sub-question of the answerable training set (19,938 questions; 10,686 templates). We abstract the relation phrase after ``$\gg$'' when it is present, and otherwise the question with bridge entities replaced by placeholders. Exponents were estimated from rank-frequency data (Sibuya-type estimator on counts ${\le}10$), from the Clauset--Shalizi--Newman tail fit, and from Heaps' law. The Sibuya and Heaps estimators give $s\in[1.15,1.26]$ for the fine-grained templates. The CSN tail fit gives $\hat s\approx0.85$--$0.88$ with a poor goodness of fit on MuSiQue ($p<0.01$), so we do not use it. Coverage curves average 100 random stream orders.

\paragraph{\textsc{KQA-Stream} construction.} We index the KQA Pro knowledge base: 16,960 entities, 794 concepts, 363 relations (726 relation--direction types) and 415,702 directed edges. Our path executor agrees with the official KoPL engine on 600 sampled one- and two-hop programs (2,187 steps). A template is an anchor concept plus a relation path of 2--4 hops. It is kept if at least 60 anchors have 1--3 answers of a consistent concept, the path does not immediately backtrack, and no single answer covers more than half of the anchors. From 2,250 qualifying distinct paths we selected 300 templates (120/120/60 with 2/3/4 hops), covering 17 anchor concepts, 51 answer concepts and 114 relations, and sampled 60 anchors each (18,000 instances). A random pattern verbalizes each instance question; 296 of the 300 templates have three LLM-written patterns with a single entity slot (the rest have fewer valid patterns). An edit event replaces the object of a relation for a fraction $f\in\{0.25,0.6\}$ of the relevant subjects with a random entity of the same concept, keeping inverse edges consistent. Each of three sequences has 24 events. An event changes the gold answer of 223 ($f{=}0.25$) or 386 ($f{=}0.6$) instances on average and touches $3.9\%$ of templates, which become $32\%$ or $59\%$ stale respectively. Templates that share an edited relation go stale together. The per-template change rate grows with derivation length (events that hit a 2/3/4-hop template: $2.7/4.2/5.5\%$; Spearman $0.34$), which makes staleness itself a multi-step-reasoning phenomenon.

\paragraph{LLM paths.}\label{app:llm-paths} The \emph{live} agent answers hop by hop. At each step its prompt contains the question, the linked anchor, the path so far, the current frontier (up to 5 entity names) and a numbered list of the (relation, direction) pairs available at the frontier (up to 60, sorted by frequency). It outputs JSON that either takes one hop or answers, with at most 5 hops, and may filter the final frontier by one of its concepts. The answer is the resulting entity set, and it is correct if it equals the gold set of the current knowledge-base version. For Qwen3-8B a live query uses $3.24$ calls with $1{,}486$ prompt and $62$ completion tokens on average; on the unedited knowledge base its live accuracy is $0.625$, $0.305$ and $0.190$ at 2, 3 and 4 hops ($0.410$ over all 18{,}000 instances). \emph{Routing}, the one LLM step of a macro, retrieves the top-5 templates by BM25 over their verbalized patterns and asks the LLM for a template id or NONE. It runs against the full, fixed template catalog, so its outcome does not depend on the policy. Its accuracy is $0.939$ for Qwen3-8B, and the correct template is among the top 5 for all but 40 of the 18{,}000 instances. The \emph{direct} baselines use one call: a parametric answer without the knowledge base, or whole-path generation from the question and the BM25 top-60 relation directions, executed on the current version. For each call we log prompt and completion tokens and latency. After an edit, live traces are re-run only for instances whose earlier trace or gold path touches an edited edge; all other traces are reused unchanged, which is exact because the agent's inputs are unchanged. A brute-force re-run of all instances at versions $0,8,16,24$ of every sequence matched in all $216{,}000$ comparisons.

\paragraph{Baselines and statistics.} The ExpeL-style counter starts at importance $2$, moves by $\pm1$ per audited success or failure and retires the macro at $0$. The ACE-style counter keeps helpful and harmful tallies from audited outcomes and retires the macro when it would be flagged (harmful $\ge$ helpful, harmful $>0$). \emph{Fold once} folds a verified candidate at formation and never retires it. A semantic cache keyed on question embeddings reduces to this policy behind our router. vCache's per-request bound assumes i.i.d.\ prompts and has no retirement; \cref{app:focus-vcache} evaluates a re-implementation, which avoids staleness mainly by rarely answering from cache. No policy was tuned on the evaluation streams. The validation-gated Hebbian rule uses its published defaults ($\eta=0.1$, $\lambda=10^{-3}$, $\tau=0.85$, $c_0=0.5$, fusion bonus $0.05$). Our certificates use $\delta=\beta=0.05$ and a likelihood-ratio CUSUM with threshold $h=7$ and post-change failure rate $\max(1-\hat r^\star,\hat q_0+0.15)$; its in-control rate $\hat q_0$ is estimated from the candidate's shadow labels. The betting certificate mixes seven fixed bets $\lambda\in\{0.05,0.1,0.2,0.3,0.5,0.7,0.9\}$ with equal weights. Any fixed mixture is valid; the grid mixture of \cref{app:utility} is needed only for the rate result. All defaults were fixed on synthetic tables before any real-LLM run. \emph{Risk ledger.} The library holds at most one candidate or macro per template. The \emph{lifetime ledger} tests the $j$-th test of a template, counted over the whole stream, at $\delta_j=6\delta/(\pi^2j^2)$. The restart-only certificate adds \emph{restarts}: the replay notifies it of every edit that changes the gold answer of an instance routed to a template, and a candidate under test then discards its wealth and shadow statistics and restarts with the next ledger index (\cref{app:ledger}). By \cref{prop:app-ledger}(iv) this satisfies \cref{def:safety} when the certificate's null changes only at notices (\cref{app:r4-silent}). The recommended deployment (``ours'' in the main text) also drops the notified template's folded macro or candidate and re-forms it under the ledger (invalidate + LR, \cref{app:r3-info}). Without restarts, the engine never tells a policy that an edit has occurred, and a candidate keeps its betting wealth (or SPRT statistic) across edits. The ledger then still bounds by $\delta$ the probability that a template ever folds a candidate that was at or below break-even at all its evaluations, and, given the history at the start of any epoch, the probability that a candidate formed in that epoch folds while bad. A candidate carried across an edit with wealth $w$ is covered only by the conditional bound $\delta_jw$, so this variant does not satisfy \cref{def:safety} in general. The \emph{per-candidate} variants test each new candidate at a fresh level $\delta=0.05$. This gives per-candidate conditional validity only, because a template's total fold risk grows with its number of candidates. Compiled plans were run with the restart-only certificate (restarts), with the ledger without restarts, and per candidate. The SpF columns count every fold event of a candidate whose reliability at the fold's knowledge-base version is below the break-even $r^\star$ of its template at that version ($L=20\bar m$), including candidates carried across edits (\cref{app:acl2-epoch}). \emph{Same information and tuning.} Every policy sees the same queries, edits, audit coins and verifier labels; all certificates share $\delta$, the bet grid and $h$, and no threshold was tuned on the evaluation streams. Only our certificates receive edit notices: restart only uses them to restart candidates under test, ours also to invalidate, and \cref{tab:models} reports ours with every notice missed; \cref{app:r3-info} gives the notices to baselines that invalidate on them, and \cref{app:r3-safety,app:r3-decode} report the answer-only safety replays and the decoding-variance subset.
On Qwen3-8B the ledger without restarts lowers accuracy by $0.35$ and $0.41$ points relative to per-candidate testing and raises $J_{20}$ by about $1\%$ (betting: $0.646\to0.643$, $J_{20}$ $25.6\to25.9$; compiled + betting: $0.677\to0.673$, $J_{20}$ $23.4\to23.7$). On the other four models the change in accuracy ranges from $-0.59$ to $+0.14$ points. Restarts cost another $0.1$--$0.4$ points on every model (\cref{tab:app-models,tab:app-reset}). Paired differences use a 95\% bootstrap clustered by template (the 300 templates are resampled) over the 10 seeds $\times$ 3 edit sequences. Each instance--version is decoded once per model, so these intervals do not include LLM sampling variance.

\paragraph{Replay setting.} The replays fix the following setting. Audits reveal the gold answer of the current knowledge-base version at no LLM cost. Templates are given by the benchmark, and each has up to three verbalization patterns; routing errors enter every metric as measured losses (a misrouted macro answer counts as wrong) and are not modeled by the theory. A macro executes a verified relation path exactly, which is the main source of its accuracy gain over live reasoning; text notes and distilled models would need a costlier, noisier verifier. Each instance--version is decoded once per model, so intervals resample templates, streams and edit sequences, not decoding randomness (measured separately in \cref{app:r3-decode}), and calls and latencies are reconstructed from the recorded requests. The main edits are synthetic but structured: correlated across templates that share a relation, and more frequent for longer derivations; \cref{app:natural-edits} adds a natural timeline and \cref{app:codestream} a code domain with an exact verifier. Edit notices list the templates whose gold answers change; changes of live reasoning are not announced (\cref{app:r4-silent}). Two comparisons are deliberately narrow. Compiled plans re-execute a stored path on the current facts; when such re-execution is cheap it can dominate caching the result, and \cref{tab:realllm} reports this baseline. The vCache-style cache stores unverified live answers, so on this benchmark it inherits live accuracy (\cref{app:focus-vcache}); a verified exact-repeat cache is evaluated in \cref{app:r3-info}. The fundamental limit concerns candidates informed by the outcomes of their own template; transfer across correlated templates, or a macro that a change does not affect, lies outside its model.

\paragraph{Completed coverage and validation.}
The final analysis includes all five models, three edit sequences and ten seeds, with 57,750 distinct configuration--seed checkpoints after deduplicating identical effective settings. Every checkpoint and its trace-derived summary passed completion and version checks. Recomputing the 80 primary model--policy cells reproduces the exported mean accuracy, calls and $J_{20}$. Paired differences use per-candidate betting + LR-CUSUM as the reference, except the price of restarts (\cref{tab:app-reset}), which is paired against the same ledger without restarts.

\paragraph{All policies and all models.} \Cref{tab:app-full} adds to \cref{tab:realllm} one-call path generation, the validation-gated Hebbian rule with time decay, and compiled plans certified by the plug-in SPRT. Hebbian decay almost never folds, because its time-to-live starves all but the most popular templates (\cref{app:hebbian}). \Cref{tab:app-models} gives accuracy, calls and $J_{20}$ for all five models.

\begin{table}[h]
\centering
\small
\setlength{\tabcolsep}{2.4pt}
\begin{tabular}{lrrrrr}
\toprule
Policy (Qwen3-8B) & Acc & Calls & Stale & SpF & $J_{20}$ \\
\midrule
direct (path generation) & .186 & 1.00 & 0 & 0 & 54.1 \\
Hebbian + decay & .444 & 3.80 & 0.4 & 0.3 & 40.1 \\
compiled plan + SPRT & .650 & 2.51 & 10 & 0.3 & 25.3 \\
loss-tuned & .631 & 1.62 & 659 & 118 & 25.7 \\
SPRT + symmetric & .604 & 2.84 & 20 & 0.7 & 28.7 \\
\textbf{betting, no restarts} & .643 & 2.55 & 72 & 0.1 & 25.9 \\
\textbf{betting, per candidate} & .646 & 2.50 & 76 & 0.2 & 25.6 \\
\textbf{plan, ledger, no restarts} & .673 & 2.42 & 14 & 0.0 & 23.7 \\
\textbf{plan, per candidate} & .677 & 2.38 & 12 & 0.03 & 23.4 \\
invalidate on notice (no LR) & .651 & 2.42 & 0 & 0.0 & 25.2 \\
\bottomrule
\end{tabular}
\caption{Further Qwen3-8B rows for \cref{tab:realllm} (same setting and columns; SpF counts the bad folds of the main text). Loss-tuned: fold and retire thresholds set from a renewal-reward approximation of the loss. \emph{Per candidate}: a fresh level $\delta$ for each candidate. The last row drops and re-forms every macro an edit announces, with perfect notices (\cref{app:r3-info}). Bold: ours.}
\label{tab:app-full}
\end{table}

\begin{table*}[t]
\centering
\small
\setlength{\tabcolsep}{3pt}
\resizebox{\textwidth}{!}{%
\begin{tabular}{lrrrrrrrrrrrrrr}
\toprule
& \multicolumn{3}{c}{live} & \multicolumn{3}{c}{\begin{tabular}[b]{@{}c@{}}\textbf{ours}\\\textbf{(recommended)}\end{tabular}} & \multicolumn{4}{c}{\begin{tabular}[b]{@{}c@{}}\textbf{macros + betting}\\\textbf{restart only / no restarts / per-cand.}\end{tabular}} & \multicolumn{4}{c}{\begin{tabular}[b]{@{}c@{}}\textbf{compiled plan + betting}\\\textbf{restarts / no restarts / per-cand.}\end{tabular}}\\
\cmidrule(lr){2-4}\cmidrule(lr){5-7}\cmidrule(lr){8-11}\cmidrule(l){12-15}
Model & Acc & Calls & $J_{20}$ & Acc & Calls & $J_{20}$ & Acc & Calls & Stale & $J_{20}$ & Acc & Calls & Stale & $J_{20}$ \\
\midrule
Qwen3-8B & .379 & 3.26 & 43.8 & .648 & 2.56 & 25.5 & .642/.643/.646 & 2.57 & 66/72/76 & 26.0 & .668/.673/.677 & 2.46 & 16/14/12 & 24.1 \\
Qwen3-4B & .226 & 4.48 & 74.0 & .556 & 3.10 & 43.0 & .551/.554/.553 & 3.08 & 82/85/91 & 43.5 & .591/.596/.596 & 2.90 & 10/12/12 & 39.7 \\
Granite-3.3-8B & .152 & 4.47 & 80.3 & .439 & 3.57 & 53.8 & .438/.441/.447 & 3.52 & 79/82/84 & 53.8 & .469/.472/.477 & 3.39 & 13/15/16 & 50.9 \\
Mistral-7B & .097 & 4.96 & 94.6 & .241 & 5.00 & 80.5 & .245/.247/.250 & 4.94 & 38/42/43 & 80.0 & .257/.259/.258 & 4.88 & 9/9/10 & 78.7 \\
OLMo-3-7B & .076 & 2.39 & 46.4 & .280 & 2.68 & 36.7 & .276/.280/.283 & 2.68 & 51/51/51 & 36.9 & .294/.297/.299 & 2.64 & 11/12/12 & 36.0 \\
\bottomrule
\end{tabular}}
\caption{All five models in the main setting (30 streams of 200k queries each). Stale: stale wrong answers per $10^4$ queries. Ours: invalidate on notice + LR-CUSUM, perfect notices (it serves no stale answers). Calls and $J_{20}$ of the other groups are for the first variant listed: the restart-only certificate with restarts (restart only), for macros and for compiled plans (compiled plan, \cref{app:r3-info}). Entries are descriptive means; calls are reconstructed from recorded requests. Bold: ours.}
\label{tab:app-models}
\end{table*}

\paragraph{Where the accuracy comes from.} On Qwen3-8B (\cref{tab:realllm}), \emph{reuse} without certification (fold once) adds $15.5$ points over live reasoning for macros and $26.1$ for compiled plans. \emph{Certification} with retirement and restarts adds $10.7$ and $2.9$ more and cuts stale answers $20\times$ and $6\times$; over the counters, which reuse and retire too, it adds $2.3$--$2.9$. Reuse helps this much because a macro executes a verified relation path symbolically, whereas the live agent re-derives it with a noisy LLM at $38\%$ accuracy. For text or distilled macros this advantage need not hold, so we regard the accuracy gain as specific to this benchmark and the staleness result as the transferable one. On OLMo-3-7B the certificate uses more calls than live reasoning ($2.39\to2.68$; \cref{tab:app-models}). Stale counts also depend on how much a policy folds: SPRT + symmetric serves $20$ stale answers per $10^4$ queries against restart only's $66$ on Qwen3-8B, but answers $43\%$ of queries in one step against $51\%$. This is why the main text compares retirement rules on a matched population (\cref{app:acl3}).

\subsection{Natural template streams}
\label{app:natural}

\begin{figure}[h]
\centering
\includegraphics[width=\columnwidth]{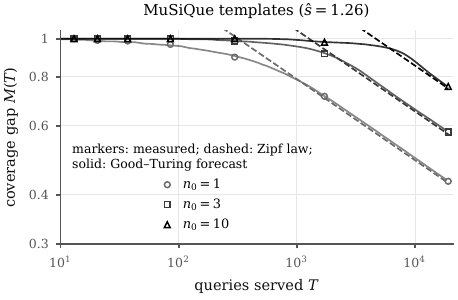}
\caption{MuSiQue templates: measured coverage gap (markers; mean of 100 stream orders), fitted Zipf law (dashed) and online Good--Turing forecast (solid) for certification sample sizes $n_0$.}
\label{fig:app-natural}
\end{figure}

We extracted derivation templates from 106k KQA Pro programs \citep{cao2022kqa} and 19.9k MuSiQue decompositions \citep{trivedi2022musique}, with entities and values abstracted (details above). Both are heavy-tailed: the Sibuya-type estimator gives $\hat s=1.26$ and Heaps' law $\hat s=1.15$ on both corpora. The static branch of \cref{thm:fss} therefore predicts that coverage improves only as $T^{-(0.13\text{--}0.21)}$, as observed (\cref{fig:app-natural}): slopes at $n_0=1$ are $-0.16$ (KQA Pro) and $-0.18$ (MuSiQue), steepening to $-0.19$ and $-0.21$ late in the stream. The Good--Turing estimator of the rare mass \citep{good1953population,orlitsky2016optimal}, $\hat\cov(T)=\sum_{k\le n_0}kK_k(T)/T$ with $K_k$ the number of templates seen $k$ times, predicts the gap online with a median relative error of $0.4$--$3.4\%$ ($2.1$--$3.4\%$ at $n_0=1$), without knowing $s$. Raising the certification sample size $n_0$ slows contraction by $C_{s,n_0}/C_{s,1}\approx n_0^{1-1/s}/\Gamma(2-\frac1s)$. These streams test the demand side of the theory without an LLM in the loop.

\subsection{Learning rates, counters, noisy verifiers and a semantic cache}
\label{app:focus}

These experiments replay the recorded \textsc{KQA-Stream} outcome tables of the five models; no LLM is called. They use the main setting (200k queries, Zipf $s=1.2$, Poisson edits at rate $28/T$ so that nearly all 24 edit events of a sequence occur, uniform audits $p=0.1$, $L=20$) with three edit sequences and ten seeds, i.e.\ 30 runs per cell. Intervals are 95\% bootstraps that resample the 300 templates (1000 resamples), pooled over sequences and seeds, and paired differences use the same template bootstrap as the main text. Here \emph{ours} is betting + LR-CUSUM at the per-candidate level; it reproduces the per-candidate row of \cref{tab:realllm} exactly (Qwen3-8B: accuracy $0.646$, $J_{20}$ $25.58$, $76$ stale answers per $10^4$).

\emph{Definitions.} A false retirement (FR) is the retirement of a macro whose true reliability is still $\ge\rstar$; we count them per $10^4$ folded uses of such healthy macros. The detection delay of a stale episode is the number of folded uses from its onset, the version at which a macro folded while healthy drops below $\rstar$, to its retirement. Episodes still open at the end of a stream contribute their uses so far, so delays are lower bounds; \cref{app:acl2-km} gives censoring-aware estimates. At a \emph{matched FR}, each rule's delay and stale answers are interpolated linearly in $\log$ FR along its threshold grid, at the FR$^\ast$ of our LR-CUSUM at its default $h=7$. Matching uses the true reliabilities. This makes the comparison of detectors fair, but it is not a deployable tuning procedure.

\subsubsection{Learning-rate sweep}
\label{app:focus-lr}

\paragraph{Rules.} A fixed-ratio rule keeps our betting certificate and retires with the clipped score $+1$ per audited failure and $-1/k$ per audited success, retiring at $S>h$, for $k\in\{0.1,0.25,0.5,0.75,1,2,3,5,7,10,15,25\}$ and $15$ thresholds $h\in[0.5,80]$. Ours is the same except that the score is the log-likelihood ratio of the macro's own estimated $\hat q_0$ against $\max(1-\hat r^\star,\hat q_0+0.15)$. The failure rates $q_0$ and $q_1$ of \cref{tab:app-e1} are the true failure rates ($1-$reliability, including routing errors) of the stale episodes of our policy, $q_0$ at the fold version and $q_1$ at the onset version. The prediction $1/\rho(\bar q_0,\bar q_1)$ uses their means.

\begin{table*}[t]
\centering
\small
\setlength{\tabcolsep}{3.2pt}
\resizebox{\textwidth}{!}{%
\begin{tabular}{lrrrrlllrrrrrr}
\toprule
& & & & & & \multicolumn{2}{c}{empirical optimum} & \multicolumn{4}{c}{delay at matched FR (uses)} & \multicolumn{2}{c}{stale per $10^4$}\\
\cmidrule(lr){7-8}\cmidrule(lr){9-12}\cmidrule(l){13-14}
Model & episodes & $\bar q_0$ & $\bar q_1$ & $q_1{>}\frac12$ & predicted $1/\rho$ [95\% CI] & $k$ & $P(\arg\min)$ & $k{=}1$ & $k{=}7$ & best $k$ & \textbf{ours} & best $k$ & \textbf{ours} \\
\midrule
Qwen3-8B & 81 & .068 & .746 & .86 & 1.84 [1.36, 2.47] & 25 & .24 & 853 & 482 & 468 & 81 & 185 & 76 \\
Qwen3-4B & 65 & .044 & .746 & .88 & 2.14 [1.52, 3.32] & 0.75 & .34 & 330 & 398 & 317 & 54 & 231 & 91 \\
Granite-3.3-8B & 32 & .161 & .863 & .97 & 0.93 [0.57, 1.50] & 0.25 & .40 & 169 & 317 & 126 & 63 & 173 & 84 \\
Mistral-7B & 17 & .237 & .896 & .94 & 0.67 [0.39, 1.11] & 0.1 & .33 & 141 & 252 & 106 & 84 & 78 & 43 \\
OLMo-3-7B & 17 & .135 & .821 & .94 & 1.14 [0.59, 2.83] & 0.25 & .23 & 131 & 227 & 115 & 57 & 88 & 51 \\
\bottomrule
\end{tabular}}
\caption{Learning-rate sweep on \textsc{KQA-Stream}. Episodes: stale episodes over 30 runs. $q_1{>}\frac12$: share of stale episodes with $q_1>0.5$. $P(\arg\min)$: bootstrap frequency of the empirical optimum. Best $k$: the fixed ratio with the lowest delay at matched FR over all tested $k$, and with the fewest stale answers over $k\ge1$. Ours: per-macro LR-CUSUM (per-candidate certificate). Delays are naive means over censored episodes. Bold: ours.}
\label{tab:app-e1}
\end{table*}

\begin{figure*}[t]
\centering
\includegraphics[width=\textwidth]{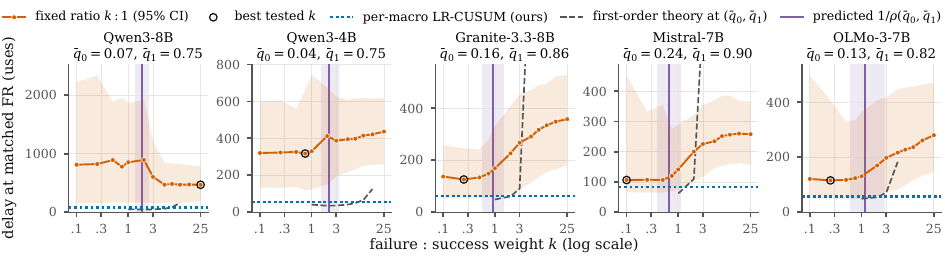}
\caption{\textbf{Per-memory likelihood ratios beat every tested global counter ratio.} Mean uses to detect staleness for a clipped counter that weighs a failure $k$ times a success (orange; circle: best tested $k$) and for the per-memory LR-CUSUM (blue dotted), at the false-retirement rate of the latter ($1.9$--$3.4$ per $10^4$ healthy uses). Purple: the weight $1/\rho$ that \cref{cor:lr} predicts from the mean failure rates $(\bar q_0,\bar q_1)$; dashed: its first-order delay, an \emph{optimum}-level rate that the \emph{measured} delays exceed (by $2.3$--$17\times$ at $k=1$; see the results below). On four of five models the delay is flat or lowest up to the predicted weight and grows beyond it; the per-memory gain is significant on Qwen3-8B, Qwen3-4B and OLMo-3-7B. 30 streams per model, per-candidate certificate; bands: 95\% template bootstrap; naive means, which favour the slower rules (\cref{app:focus-lr}).}
\label{fig:e1}
\end{figure*}

\paragraph{Results.} Staleness on these streams is nearly total, not partial: $86$--$97\%$ of stale episodes have $q_1>0.5$ (\cref{tab:app-e1}). The predicted weights $1/\rho$ are therefore $0.67$--$2.14$, not $7$. On four of five models the delay at matched FR is smallest below $k=1$ ($k=0.1$--$0.75$), consistent with the ordering of the prediction, and it grows with $k$ above $1$; $k=7$ is $1.3$--$2.5\times$ slower than the best tested ratio on these models (\cref{fig:e1}). Below $1$ the curves are flat, so the optimum there is not sharply identified. Qwen3-8B is the exception: its curve is flat for $k\ge5$ and worse at $k\le2$, and the bootstrap spreads its optimum over $k=25$, $5$ and $10$. The larger effect is not the tuning of $k$. The per-macro LR-CUSUM detects $1.3$--$5.9\times$ faster than each model's best tested fixed ratio (paired intervals exclude $1$ on Qwen3-8B, Qwen3-4B and OLMo-3-7B; \cref{app:acl2-decomp}) and serves $1.7$--$2.5\times$ fewer stale answers than the fixed ratio $k\ge1$ with the fewest. Every fixed ratio at $k=7$ is significantly slower than ours, and at $k=1$ this holds on three of five models (paired intervals exclude $0$). Healthy failure rates are heterogeneous: their medians ($0$--$0.21$) lie far below their means. We interpret this, without a direct test, as the reason why one global ratio cannot be likelihood-ratio tuned for all macros at once, whereas the per-macro rule adapts $\hat q_0$ to each macro. The first-order delay curve follows the rising trend on Granite, Mistral and OLMo but not on Qwen3-8B. At $k=1$ it underestimates the measured delays by $2.3$--$17\times$, and at larger $k$ it leaves the logarithmic regime on three models. With the episode-median in place of the mean $(q_0,q_1)$ the predicted weights are $0.51$--$3.01$. With only $17$--$81$ stale episodes per model the intervals are wide, and only the large differences are resolved.

\subsubsection{Counters in the main setting}
\label{app:focus-counters}

ExpeL- and ACE-style counters serve $294$--$967$ stale answers per $10^4$ queries ($7$--$13\times$ ours), fold $111$--$595$ spurious macros per run (ours: at most $1.7$), and have $10$--$35$ false retirements per $10^4$ (ours: $1.8$--$3.4$) with naive mean delays of $307$--$478$ uses (ours: $54$--$84$; lower bounds, see \cref{app:acl2-km}). Their $J_{20}$, however, is not significantly different from ours on any model, and their accuracy is significantly lower only on Qwen3-8B (\cref{tab:app-e2}). The reason is coverage, as the price of safety of \cref{sec:limit} suggests: the counters skip certification and answer $28$--$76\%$ of queries in one step, against $21$--$53\%$ for the restart-only certificate and $20$--$53\%$ for ours (recommended), so they make fewer LLM calls (e.g.\ $1.75$ and $1.86$ against $2.57$ on Qwen3-8B, \cref{tab:realllm}), and in $J_{20}$ this saving roughly offsets their stale and spurious answers.
\begin{table*}[t]
\centering
\small
\setlength{\tabcolsep}{5pt}
\begin{tabular}{lllll}
\toprule
& \multicolumn{2}{c}{ExpeL-style counter} & \multicolumn{2}{c}{ACE-style counter}\\
\cmidrule(lr){2-3}\cmidrule(l){4-5}
Model & $\Delta$Acc & $\Delta J_{20}$ & $\Delta$Acc & $\Delta J_{20}$ \\
\midrule
Qwen3-8B & $-.033$ [$-.061$, $-.004$] & $+1.38$ [$-0.56$, $+3.20$] & $-.028$ [$-.055$, $-.001$] & $+1.15$ [$-0.72$, $+2.95$] \\
Qwen3-4B & $+.006$ [$-.020$, $+.035$] & $-1.18$ [$-3.93$, $+1.34$] & $+.002$ [$-.026$, $+.032$] & $-0.68$ [$-3.61$, $+1.99$] \\
Granite-3.3-8B & $-.014$ [$-.034$, $+.009$] & $+0.87$ [$-1.27$, $+2.79$] & $-.014$ [$-.035$, $+.009$] & $+1.05$ [$-1.14$, $+3.01$] \\
Mistral-7B & $-.008$ [$-.023$, $+.006$] & $+0.30$ [$-1.17$, $+1.79$] & $-.008$ [$-.022$, $+.006$] & $+0.49$ [$-0.94$, $+1.93$] \\
OLMo-3-7B & $+.007$ [$-.019$, $+.037$] & $-0.47$ [$-2.13$, $+0.90$] & $-.000$ [$-.028$, $+.032$] & $-0.09$ [$-1.82$, $+1.36$] \\
\bottomrule
\end{tabular}
\caption{Main setting: paired differences of the counters against ours (betting + LR-CUSUM, per-candidate level), with 95\% template-bootstrap intervals; positive $\Delta J_{20}$ is worse. Only the counters' accuracy on Qwen3-8B differs significantly.}
\label{tab:app-e2}
\end{table*}

\subsubsection{A noise-aware certificate}
\label{app:focus-noise}

\paragraph{Setup.} Noisy verifiers are the next step named in the Limitations; this experiment measures where the current certificate stands. Each audited judgement (of the live answer, the macro and any other evaluated answer) is flipped independently with probability $\eps$. The noisy live verdict also decides which traces count as verified when candidates are formed. The noise-aware variant bets on the paired utility of the observed channel, $X=(1-2\eps)(m-1)+L(\tilde Y^M-\tilde Y^L)$, and maps the CUSUM's alternative into the observed channel, $q_1=\max(\eps+(1-2\eps)(1-\rstar),\,q_0+0.15(1-2\eps))$, with live accuracy corrected as $a=(a_{\rm obs}-\eps)/(1-2\eps)$. It uses either the true $\eps$ or an estimate from a simulated clean calibration subset of $200$ items, drawn per sequence and seed (mean estimates $0.054$, $0.103$ and $0.201$). It corrects the certificate and the CUSUM, not the formation of candidates. The labels-to-fold slowdown is the mean over templates of the number of audited shadow labels a candidate needs before it folds, relative to the uncorrected certificate at $\eps=0$.

\begin{table}[h]
\centering
\small
\setlength{\tabcolsep}{2.6pt}
\begin{tabular}{llrrrrr}
\toprule
$\eps$ & certificate & Acc & Stale & SpF & slowdown & $(1{-}2\eps)^{-2}$ \\
\midrule
0 & \textbf{uncorrected} & .436 & 69 & 0.71 & 1.00 & 1.00 \\
\midrule
\multirow{3}{*}{.05} & \textbf{uncorrected} & .370 & 63 & 0.21 & 1.88 & 1.23 \\
& \textbf{known} $\boldsymbol{\eps}$ &.370 & 61 & 0.21 & 2.00 & \\
& \textbf{estimated} $\boldsymbol{\eps}$ &.369 & 61 & 0.21 & 1.99 & \\
\midrule
\multirow{3}{*}{.1} & \textbf{uncorrected} & .326 & 61 & 0.17 & 2.56 & 1.56 \\
& \textbf{known} $\boldsymbol{\eps}$ &.328 & 59 & 0.12 & 2.82 & \\
& \textbf{estimated} $\boldsymbol{\eps}$ &.327 & 59 & 0.14 & 2.70 & \\
\midrule
\multirow{3}{*}{.2} & \textbf{uncorrected} & .279 & 53 & 0.27 & 3.67 & 2.78 \\
& \textbf{known} $\boldsymbol{\eps}$ &.286 & 50 & 0.21 & 3.36 & \\
& \textbf{estimated} $\boldsymbol{\eps}$ &.286 & 49 & 0.22 & 3.21 & \\
\bottomrule
\end{tabular}
\caption{Noisy verifier, per-candidate betting, means over the five models (30 streams each; at $\eps=0$ the mean of the per-candidate column of \cref{tab:app-models}). Stale: per $10^4$ queries. SpF: spurious folds per run. Slowdown: audited labels a candidate needs before it folds, relative to $\eps=0$. Bold: ours (all rows).}
\label{tab:app-e3}
\end{table}

\paragraph{Results (negative).} Certification slows faster than $(1-2\eps)^{-2}$ (\cref{tab:app-e3}): the labels-to-fold slowdown is $1.88$, $2.56$ and $3.67$ at $\eps=0.05$, $0.1$ and $0.2$, against $1.23$, $1.56$ and $2.78$ predicted, and the slowdown in key-queries is larger still ($2.15$, $3.20$, $5.05$). The prediction is for a single-label test; our certificate bets on the difference of two noisy labels. The noise-aware variant, with known or estimated $\eps$, is indistinguishable from the uncorrected certificate on accuracy, stale answers and spurious folds, and nearly every per-model accuracy interval contains $0$. Noise does not inflate spurious folds ($0.71$ per run clean, $0.12$--$0.27$ noisy), because far fewer candidates fold at all. What dominates is the loss of accuracy for every variant ($0.436$ to $0.279$ at $\eps=0.2$), driven by slower certification and by macros formed from falsely verified live traces, neither of which the correction addresses. Counter rules, which fold nearly everything, keep their accuracy under noise (Qwen3-8B, 15 streams: $.592\to.575$ and $.597\to.580$ at $\eps=0.2$) but fold more spurious macros (ExpeL $271\to385$, ACE $377\to601$ per stream). A certificate built for the noisy \emph{paired} channel is left to future work.

\subsubsection{A vCache-style semantic cache}
\label{app:focus-vcache}

\paragraph{Re-implementation.} We re-implemented vCache from Algorithms~1--2 of its paper \citep{schroeder2026vcache}; the following details are our design choices. A query is looked up against the nearest cached question by cosine similarity (ties go to the newest entry). Each entry fits a sigmoid of similarity by ridge-penalized logistic maximum likelihood ($\lambda=10^{-3}$), with a delta-method upper confidence bound on its threshold. The exploration probability is the minimum of the paper's Eq.~11 over a grid of 49 values of its error parameter, and $1$ while an entry has no observations. An exploration answers live, evaluates the cached answer in shadow at no cost, and adds the query to the cache only if the cached answer was judged wrong; an exploitation returns the cached answer at no LLM call, and embedding compute is not charged. Labels are vCache's own (\emph{agree}: cached answer equals the live answer) or gold correctness on audited queries (\emph{audit}). There is no retirement. The \emph{plug-in} variant uses the point estimate without the confidence band, which is optimistic. The vCache paper used GTE-large, E5-large-v2 and OpenAI embeddings. None was available offline, so we use all-MiniLM-L6-v2 (384 dimensions), with TF-IDF as a sensitivity check; the median nearest-neighbour similarity over the 17{,}927 unique questions is $0.81$ (MiniLM) and $0.66$ (TF-IDF). A faithful replication would need a 1024-dimensional embedding and the authors' code to confirm the confidence band. A cache hit on a different question counts as stale if the cached answer would have been correct for that query when the entry was formed.

\begin{table}[h]
\centering
\small
\setlength{\tabcolsep}{2.6pt}
\resizebox{\columnwidth}{!}{%
\begin{tabular}{lrrrrrrr}
\toprule
& & \multicolumn{3}{c}{vCache, $\delta_c=0.02$} & \multicolumn{3}{c}{plug-in, $\delta_c=0.02$}\\
\cmidrule(lr){3-5}\cmidrule(l){6-8}
Model & live Acc & hits & Acc & $\Delta J_{20}$ & hits & Acc & Stale \\
\midrule
Qwen3-8B & .379 & .124 & .377 & $+17.9$ & .847 & .309 & 536 \\
Qwen3-4B & .226 & .119 & .225 & $+30.3$ & .848 & .187 & 241 \\
Granite-3.3-8B & .152 & .110 & .151 & $+26.9$ & .841 & .126 & 203 \\
Mistral-7B & .097 & .111 & .097 & $+14.7$ & .847 & .082 & 115 \\
OLMo-3-7B & .076 & .119 & .076 & $+9.5$ & .835 & .063 & 100 \\
\bottomrule
\end{tabular}}
\caption{vCache-style cache (MiniLM, \emph{agree} labels). Hits: share of queries served from cache. $\Delta J_{20}$: against ours (per-candidate betting); all intervals exclude $0$. Stale: per $10^4$ queries. Our accuracy is $.646$, $.553$, $.447$, $.250$ and $.283$.}
\label{tab:app-e5}
\end{table}

\paragraph{Results.} With its confidence band, vCache serves only $6$--$26\%$ of queries from cache for $\delta_c=0.01$--$0.05$, and $3\%$ with audited labels. Its accuracy equals live accuracy, and its $J_{20}$ is $9.2$--$30.7$ worse than ours on every model (\cref{tab:app-e5}). TF-IDF embeddings give similar results. The reason is that this cache stores earlier live answers without verifying them, so here it inherits live accuracy ($0.08$--$0.38$), whereas a certified macro re-executes a verified derivation (our accuracy $0.25$--$0.65$). A cache that stores only verified answers was tested for exact repeats (with notice invalidation; \cref{app:r3-info}): it serves almost no stale answers but stays at live accuracy, because it answers only $13\%$ of queries in one step; a verified \emph{semantic} cache remains untested. Removing the confidence band raises the hit rate to about $85\%$, but accuracy falls below live and stale answers rise to $100$--$536$ per $10^4$. vCache's $\delta_c$ bounds disagreement with the LLM, not error against gold: $62$--$92\%$ of its hits are wrong against gold, mirroring live accuracy. An exact-question answer cache beats vCache on $J_{20}$ but is still $7.7$--$27.0$ worse than ours. Cache hits are charged no LLM call, which favours the caches.

\subsection{Censoring, per-epoch safety and per-memory weights}
\label{app:acl2}

These replays reuse the recorded outcome tables of the five models in the main setting (30 streams per model and policy); no LLM is called. Intervals are 95\% template-clustered bootstraps (1000 resamples); pooled rows resample templates jointly across models.

\subsubsection{Censoring-aware detection delays}
\label{app:acl2-km}

A stale episode starts when a macro that was above break-even at its fold drops below $\rstar$ after an edit; its onset is the first query of that version. Retirement is the event, and the end of the 200k-query stream censors the episode. The delay is the number of folded uses from onset to retirement, as in \cref{app:focus}. We estimate the survival function by Kaplan--Meier. The restricted mean delay within a window $w$ is $\E\min(T,w)$, and ``unretired at stream end'' is the censored share. This closed-loop analysis covers our LR-CUSUM retirement with the per-candidate certificate and with the SPRT; each rule has its own population of stale episodes. \Cref{app:acl3} compares the rules on one matched population, including the restart-only certificate.

\begin{table}[h]
\centering
\small
\setlength{\tabcolsep}{2.2pt}
\resizebox{\columnwidth}{!}{%
\begin{tabular}{lrrlll}
\toprule
Rule (pooled) & episodes & cens. & unretired at end & RM delay, $10^4$ & median \\
\midrule
\textbf{LR (per-cand.\ betting)} & 212 & 13 & 6.1 [2.5, 10.8]\% & 76 [55, 107] & 45 [40, 55] \\
LR (SPRT) & 222 & 14 & 6.3 [2.4, 10.6]\% & 103 [64, 176] & 50 [44, 64] \\
ExpeL-style & 1{,}954 & 1{,}241 & 63.5 [57.5, 69.8]\% & 3{,}014 [2{,}204, 3{,}895] & 375 [216, 682] \\
ACE-style & 1{,}865 & 999 & 53.6 [47.6, 59.9]\% & 2{,}569 [1{,}882, 3{,}332] & 211 [122, 373] \\
\bottomrule
\end{tabular}}
\caption{Censoring-aware detection, pooled over five models (150 streams per rule). RM delay: restricted mean delay within $10^4$ folded uses; median: Kaplan--Meier median (folded uses). Bold: ours.}
\label{tab:app-km}
\end{table}

\begin{figure}[h]
\centering
\includegraphics[width=\columnwidth]{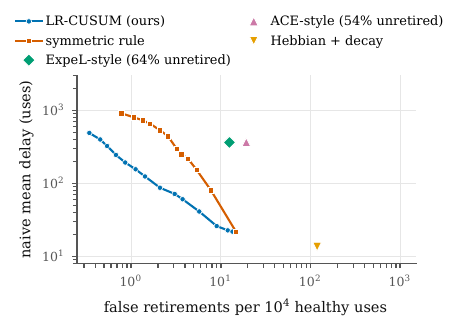}
\caption{Detection frontier on real streams (five models, 30 streams each; SPRT certification): naive mean delay after an edit against false retirements of healthy macros, as frontiers over each rule's thresholds. The naive mean counts an episode that is still open at the end of the stream with its uses so far, so it is a lower bound; the counters' censoring-aware delays are $7$--$8\times$ longer (\cref{tab:app-km}). Legend: share of each counter's stale episodes not retired by stream end.}
\label{fig:app-detect}
\end{figure}

On the audited-use clock the restricted means within $10^3$ audited uses are $7.6$ (betting), $10.4$ (SPRT), $299.9$ (ExpeL-style) and $257.6$ (ACE-style), with medians of $5$, $5$, $37$ and $21$ audited uses. The censored episodes of the LR-CUSUM are onsets near the end of a stream (at most $192$ folded uses of follow-up), whereas the counters' censored episodes include some with $39{,}467$ uses. Per model the counters leave $28$--$69\%$ of their own stale episodes unretired at stream end (\cref{tab:models} gives the matched population). The naive mean delay of \cref{fig:app-detect} counts an open episode with its uses so far. It is therefore a lower bound: $363$ and $366$ uses for the counters, $7$--$8\times$ below their restricted means.

\subsubsection{Detection on a matched population}
\label{app:acl3}

The comparison above uses different populations: the counters fold every candidate, so their $1{,}954$ and $1{,}865$ stale episodes include many macros that a certificate would never fold, whereas the LR-CUSUM rows contain $212$. Here all rules see the same macros and the same audits. Every macro that a certified policy folded (the restart-only certificate, or per-candidate betting) is replayed \emph{open loop} from its fold to the end of the stream, on its own audited label stream (the same audit coins; labels of its build instance withheld), under each retirement rule started at the fold: its own LR-CUSUM (reproducing the closed-loop retirement of all $7{,}385$ and $9{,}448$ folded macros exactly), the ExpeL-style rule (importance $2$ at the fold) and the ACE-style rule. Counters therefore bank no credit from the shadow phase, which favours them. Stale onset is defined without retirement; spurious folds are excluded. \emph{Population.} The restart-only certificate makes $7{,}385$ folds over the $150$ streams; the $43$ spurious ones (below $\rstar$ at the fold version) are not stale episodes. A \emph{matched stale episode} is a fold above $\rstar$ whose macro is below $\rstar$ at a later version before the stream ends: there are $221$. Some rule retired $65$ of them before onset (LR-CUSUM $47$, ExpeL $27$, ACE $41$; $24$ by the LR-CUSUM alone and $11$ by ACE alone), and an episode retired before it began cannot be compared, so the common population is the remaining $156$ ($49$ templates, $25$ edited relations, $34$ edit events, $3$ sequences). Code: \texttt{paper\_figs/acl3\_matched.py}.

\begin{table}[h]
\centering
\small
\setlength{\tabcolsep}{2.4pt}
\resizebox{\columnwidth}{!}{%
\begin{tabular}{lrrrrrrr}
\toprule
& & \multicolumn{3}{c}{unretired at stream end} & \multicolumn{3}{c}{RM delay within $10^4$ uses}\\
\cmidrule(lr){3-5}\cmidrule(l){6-8}
Model & eps. & ExpeL & ACE & \textbf{LR} & ExpeL & ACE & \textbf{LR} \\
\midrule
Qwen3-8B & 66 & 53\% & 53\% & 3\% & 4{,}501 & 4{,}507 & 92 \\
Qwen3-4B & 53 & 57\% & 51\% & 0\% & 4{,}391 & 4{,}139 & 46 \\
Granite-3.3-8B & 18 & 22\% & 22\% & 0\% & 3{,}278 & 3{,}251 & 44 \\
Mistral-7B & 8 & 25\% & 25\% & 0\% & 2{,}110 & 2{,}095 & 61 \\
OLMo-3-7B & 11 & 64\% & 55\% & 9\% & 5{,}071 & 4{,}507 & 942 \\
\midrule
pooled & 156 & 50\% & 47\% & 2\% & 4{,}069 & 3{,}949 & 67 \\
per-cand., pooled & 177 & 52\% & 46\% & 5\% & 4{,}115 & 3{,}773 & 72 \\
\bottomrule
\end{tabular}}
\caption{Matched-population detection: stale episodes of the macros folded by the restart-only certificate (last row: by per-candidate betting), replayed open loop under each retirement rule with identical audits; 30 streams per model. The OLMo LR-CUSUM value comes from one censored episode with long follow-up. Bold: ours.}
\label{tab:app-acl3}
\end{table}

\paragraph{Results.} Pooled, the counters leave $50.0\%$ [$38.2$, $62.8$] and $47.4\%$ [$36.3$, $60.2$] of the episodes unretired at stream end, against $1.9\%$ [$0.0$, $4.7$] for the LR-CUSUM (\cref{tab:app-acl3}, \cref{fig:km}). Their restricted mean delays are $4{,}069$ [$2{,}924$, $5{,}448$] and $3{,}949$ [$2{,}830$, $5{,}296$] uses against $67$ [$46$, $99$], ratios $60.8$ [$44.7$, $87.1$] and $59.1$ [$43.7$, $83.7$] (paired template bootstrap, $2{,}000$ resamples; the ratio is computed in each resample); KM medians are $1{,}750$, $1{,}303$ and $43$ uses, and on the audited clock the restricted means within $10^3$ audits are $403$, $391$ and $6.6$. Each episode costs $1{,}359$ and $1{,}346$ stale wrong answers under the counters, against $38.5$. The ratio interval excludes $1$ for Qwen3-8B ($49\times$), Qwen3-4B ($95\times$, $90\times$) and Granite ($75\times$, $74\times$); Mistral ($8$ episodes) and OLMo ($11$) are not resolved. The matched population thus confirms the closed-loop picture of \cref{tab:app-km}, with a larger gap in delay.

\paragraph{Robustness of the $60\times$.} Episodes of one template, relation or edit event are correlated, so we also resample these clusters. The ExpeL/LR ratio is $60.8$ with 95\% intervals [$44.7$, $87.1$] clustered by template, [$44.5$, $91.0$] by edited relation and [$41.1$, $90.0$] by edit event (ACE/LR: $59.1$; [$43.7$, $83.7$], [$43.1$, $89.3$], [$40.1$, $86.6$]). With only three edit sequences, sequence clustering is coarse ([$31.2$, $93.4$]); the ratio is $93$, $31$ and $89\times$ on sequences 101, 202 and 303 alone and $46$--$92\times$ when one sequence is left out. Without the exclusion, i.e.\ on all $221$ matched episodes with pre-onset retirements counted as zero delay, the ratio is $63.5$ [$46.9$, $88.4$] ($58.9$ for ACE), with $40.7\%$, $37.6\%$ and $3.2\%$ unretired.

\subsubsection{Per-epoch safety}
\label{app:acl2-epoch}

A template epoch is a run, a routed template and an interval between edits that change the gold answer of some instance routed to that template. A \emph{false fold} is a fold of a candidate whose true reliability (over all instances routed to the template, at stream popularity) is below $\rstar$ at the fold version; this is the SpF event of \cref{tab:models}. An epoch is \emph{at risk} if a candidate that is below break-even at some version of the epoch is under test in it, and the per-epoch false-fold frequency is the share of at-risk epochs with at least one false fold. A false fold is \emph{carried over} if the candidate formed in an earlier epoch.

This subsection uses the standard scoring; \cref{app:r3-safety} repeats the analysis with answer-only scoring, which the main text reports (\cref{fig:epoch}).

\begin{figure}[h]
\centering
\includegraphics[width=\columnwidth]{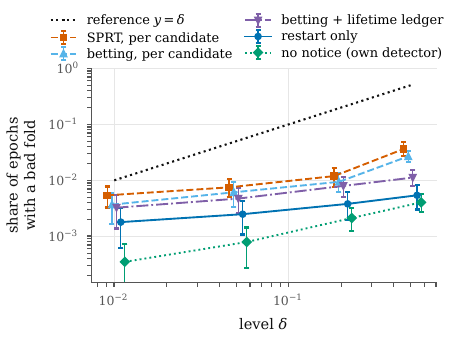}
\caption{Standard scoring (routing errors count as wrong answers): share of at-risk epochs with a task-level bad fold (reliability below $\rstar$) for five certificates; the dotted line $y=\delta$ is a \emph{reference line}, not a guarantee for these curves. \Cref{prop:app-ledger}(iv) bounds by $\delta$ only folds that violate a certificate's \emph{own} null (paired utility $\le0$; \cref{fig:epoch}), and per epoch only for the restarted certificate (blue), when notices cover every change of that null; the per-candidate variants are valid per candidate, the ledger without restarts over a template's lifetime, and the no-notice variant for folds on post-change evidence (\cref{prop:app-detect}). Five models, 150 streams per policy and $\delta$, 95\% template-bootstrap intervals.}
\label{fig:app-epoch-std}
\end{figure}

\paragraph{Results.} At $\delta=0.05$ the per-epoch false-fold frequency is $0.007$ [$0.005$, $0.010$] for the SPRT and $0.006$ [$0.003$, $0.010$] for betting, both at a fresh level per candidate, $0.005$ [$0.003$, $0.007$] for betting with the lifetime ledger and $0.002$ [$0.001$, $0.004$] for the restart-only certificate with restarts; all stay below $\delta$ for every $\delta\le0.5$ (\cref{fig:app-epoch-std}). Carried-over folds are $40$ of $106$ false folds for per-candidate betting ($16$ by candidates that were good at formation), $32$ of $70$ with the ledger ($11$), $62$ of $87$ for the SPRT ($50$), and $6$ of $43$ with restarts ($0$). \Cref{tab:app-reset} gives the price of restarts per model.

\begin{table}[h]
\centering
\small
\setlength{\tabcolsep}{2pt}
\resizebox{\columnwidth}{!}{%
\begin{tabular}{lllrrrr}
\toprule
Model & $\Delta$Acc & $\Delta J_{20}$ & stale & SpF & folds & restarts \\
\midrule
Qwen3-8B & $-.001$ [$-.002$, $.000$] & $+0.08$ [$-0.02$, $0.18$] & 72$\to$66 & 0.10$\to$0.00 & 60.0$\to$53.8 & 149 \\
Qwen3-4B & $-.003$ [$-.007$, $-.001$] & $+0.30$ [$0.07$, $0.62$] & 85$\to$82 & 0.20$\to$0.07 & 72.9$\to$66.5 & 110 \\
Granite-3.3-8B & $-.003$ [$-.006$, $-.002$] & $+0.33$ [$0.17$, $0.57$] & 82$\to$79 & 0.67$\to$0.47 & 69.4$\to$62.3 & 100 \\
Mistral-7B & $-.002$ [$-.004$, $.000$] & $+0.27$ [$0.05$, $0.48$] & 42$\to$38 & 0.97$\to$0.60 & 36.1$\to$28.1 & 138 \\
OLMo-3-7B & $-.004$ [$-.006$, $-.002$] & $+0.20$ [$0.12$, $0.32$] & 51$\to$51 & 0.40$\to$0.30 & 41.1$\to$35.5 & 79 \\
\bottomrule
\end{tabular}}
\caption{The price of restarts: the restart-only certificate against the same lifetime ledger without restarts (paired differences with 95\% template-bootstrap intervals; stale answers per $10^4$ queries, SpF, folds and restarts per stream).}
\label{tab:app-reset}
\end{table}

\paragraph{The scoring convention behind most false folds.} \textsc{KQA-Stream} scores a macro answer as correct only if the query was also routed to the right template, and a template's true reliability averages over every query routed to it, including misrouted ones. The certificate instead tests the paired utility of its own shadow evaluations. The two notions of ``bad'' disagree for some candidates: at $\delta=0.05$, $91$ of the $106$ false folds of per-candidate betting and all $43$ of the restart-only certificate have a positive exact paired-utility drift, so by the test's own criterion these candidates were useful. Among candidates that are bad under the test's own null at every version after formation, none folds ($0$ of $8{,}434$ for per-candidate betting and $0$ of $8{,}084$ for the restart-only certificate). Scoring macro answers by the answer alone removes this convention. With answer-only scoring, every remaining false fold of per-candidate betting ($15$) is carried over by a candidate that was good at formation, which is exactly the event that restarts remove.

\subsubsection{Restarts without edit notices}
\label{app:acl2-nonotice}

The restart-only certificate restarts a candidate when the replay reports an answer-changing edit of its template. The no-notice certificate (\emph{discard on alarm}; ``no notice'' in \cref{tab:algos,tab:realllm}) receives no edit information. It runs, on every candidate under test, the retirement detector of its folded macros: a Bernoulli LR-CUSUM on the candidate's audited shadow outcomes with $\hat q_0$ from the current test segment (estimated before each label), $q_1=\max(1-\hat r^\star,\hat q_0+0.15)$ and $h=7$. On an alarm it discards the candidate, and the template's next candidate forms from a later verified live trace. A retired macro is removed, so its successor starts a fresh test. \Cref{prop:app-detect} covers this rule; a fold is \emph{spanning} if its test segment began before the edit that opened the fold's epoch, the event that (iii) of that proposition bounds.

\paragraph{Results.} Discard-on-alarm has bad folds in $0.0008$ [$0.0003$, $0.0014$] of at-risk epochs at $\delta=0.05$ ($9$ of $11{,}375$), against $0.0025$ [$0.0011$, $0.0043$] for the restart-only certificate ($29$ of $11{,}670$), and raises accuracy by $1.3$ points [$0.8$, $2.0$] and lowers $J_{20}$ by $1.06$ [$0.63$, $1.60$] relative to it (\cref{tab:app-nonotice}). All $9$ of its bad folds are spanning folds of candidates that were good at formation, and all $9$ are bad under the test's own null, so they are exactly the residual risk of \cref{prop:app-detect}(iii); no fold rests on post-change evidence of a bad candidate ($0$ of $13{,}935$ candidates that are bad under the test's own null at every version after formation fold). The restart-only certificate's $29$ epochs come from the scoring convention above (all with a positive test drift). Spanning folds are frequent ($1{,}174$ of $7{,}766$ folds), but the edit left all but $9$ of these candidates good. Retirement is unchanged: $195$ stale episodes with $5.1\%$ censored and a mean delay of $68.0$ uses, against $174$, $4.0\%$ and $68.0$. The accuracy gain is positive on every model, because an alarm replaces a candidate whose cached answers broke by one formed on the current facts, while a restarted candidate keeps its stale answers. At every tested level $\delta\in\{0.01,0.05,0.2,0.5\}$, discard-on-alarm has fewer bad-fold epochs than the restart-only certificate ($0.0003$--$0.0040$ against $0.0018$--$0.0054$; \cref{fig:app-epoch-std}) and $1.1$--$1.3$ more accuracy points.

\begin{table}[h]
\centering
\small
\setlength{\tabcolsep}{2pt}
\resizebox{\columnwidth}{!}{%
\begin{tabular}{lllrrr}
\toprule
Model & $\Delta$Acc & $\Delta J_{20}$ & bad-fold epochs & alarms & stale \\
\midrule
Qwen3-8B & $+.009$ [$.004$, $.015$] & $-0.62$ [$-1.09$, $-0.26$] & $.0009$ / $.0000$ & 68.3 & 70$\leftarrow$66 \\
Qwen3-4B & $+.016$ [$.003$, $.039$] & $-1.62$ [$-3.86$, $-0.26$] & $.0023$ / $.0011$ & 61.0 & 83$\leftarrow$82 \\
Granite-3.3-8B & $+.009$ [$.004$, $.017$] & $-0.86$ [$-1.60$, $-0.33$] & $.0006$ / $.0064$ & 91.8 & 81$\leftarrow$79 \\
Mistral-7B & $+.012$ [$.006$, $.022$] & $-1.32$ [$-2.27$, $-0.62$] & $.0003$ / $.0026$ & 224.2 & 42$\leftarrow$38 \\
OLMo-3-7B & $+.017$ [$.004$, $.042$] & $-0.89$ [$-2.22$, $-0.18$] & $.0000$ / $.0053$ & 99.7 & 49$\leftarrow$51 \\
\bottomrule
\end{tabular}}
\caption{Discard-on-alarm (no edit notices) against the restart-only certificate (edit-notified restarts), $\delta=0.05$: paired differences with 95\% template-bootstrap intervals; share of at-risk epochs with a bad fold (no notice / restart only); detector alarms per stream; stale answers per $10^4$ queries (no notice $\leftarrow$ restart only).}
\label{tab:app-nonotice}
\end{table}

\subsubsection{Partial staleness on real streams}
\label{app:partial}

Half of the edits of each sequence change $25\%$ of the affected subjects ($f=0.25$), and some of them leave a macro partially stale. We test the phase diagram on these episodes open loop: every macro that per-candidate betting folds is re-scored on its full audited label stream from its fold to the end of the stream ($9{,}448$ macros over the five models). Ground truth is the macro's exact failure rate $q_v=1-R_v$ at each knowledge-base version. An \emph{episode} starts at the first version after the fold at which $q$ rises by at least $0.1$ over its value $q_0$ at the fold, has failure rate $q_1$ there, and ends at the next version at which $q$ moves by more than $0.02$, or at the stream end. We keep episodes with $q_0\in[0.02,0.15]$, since with $q_0=0$ every rule can alarm at the first failure without false alarms, and call them \emph{partial} if $q_1\in[0.15,0.5]$ and \emph{near-total} if $q_1>0.5$. Each rule restarts at the change (Lorden's worst case for a CUSUM); its delay is the number of folded uses from the change to its alarm, censored at the episode end, summarized by the Kaplan--Meier mean restricted to $1{,}000$ uses. Thresholds are matched to $1$, $3$ and $10$ false alarms per $10^4$ healthy uses, measured on the uses between fold and change of all folded macros with $q_0$ in the same band. Intervals are 95\% template-clustered bootstraps with thresholds re-matched in every resample.

\begin{figure}[h]
\centering
\includegraphics[width=\columnwidth]{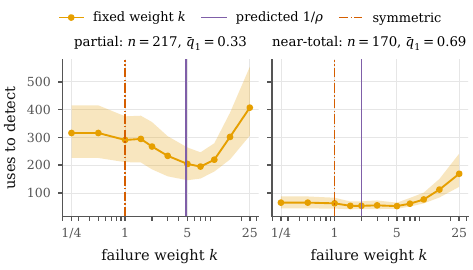}
\caption{Partial staleness on real streams: delay (Kaplan--Meier mean within $1{,}000$ uses) of fixed-weight counters against the failure weight $k$ at $1$ false alarm per $10^4$ healthy uses (95\% template-bootstrap bands), with the weight $1/\rho$ predicted by \cref{cor:lr} from the mean failure rates. Left: $217$ partial episodes ($q_1\in[0.15,0.5]$); right: $170$ near-total episodes; both with $q_0\in[0.02,0.15]$.}
\label{fig:app-partial}
\end{figure}

\paragraph{Results.} Of the $2{,}076$ single-change episodes, $217$ are partial episodes in the band ($63$ templates; $32$--$53$ per model) with mean $q_0=0.067$ and $q_1=0.331$, for which \cref{cor:lr} predicts the failure weight $1/\rho=4.8$. At $1$ false alarm per $10^4$ healthy uses the best tested weight is $7$ ($195$ uses [$152$, $246$]); the delay is U-shaped in the weight, as the efficiency \eqref{eq:eff} predicts: the symmetric rule is $1.49\times$ [$1.18$, $1.82$] slower and detects $90\%$ of the episodes within $1{,}000$ uses (weight $7$: $99.8\%$), and weight $25$ is $2.09\times$ [$1.71$, $2.65$] slower (\cref{fig:app-partial}). At $3$ and $10$ false alarms the best weight is $5$, and the symmetric rule is $1.22\times$ [$1.03$, $1.50$] and $1.29\times$ [$1.04$, $1.45$] slower. In the bootstrap the best weight is $5$ or $7$ in $98.6\%$, $93.2\%$ and $46.3\%$ of resamples at the three rates ($3$: $52.5\%$ at the highest rate). On the $170$ near-total episodes of the same band (mean $q_1=0.690$, predicted $1/\rho=2.0$) the optimum moves down to $3$--$5$ and the symmetric rule is within $1.04$--$1.19\times$ of it (intervals contain $1$), so the advantage of failure-heavy weights is specific to partial staleness, as \cref{thm:trichotomy} predicts. The per-memory LR-CUSUM tuned at the low end $q_1=\hat q_0+0.15$ (\cref{prop:minimax}) is within $1.21\times$ [$0.90$, $1.64$] of the best fixed weight; the deployed LR-CUSUM, tuned for a drop below break-even ($q_1=\max(1-\hat r^\star,\hat q_0+0.15)$), is $1.59\times$ [$1.20$, $2.16$] slower on these smaller changes, for which it is not tuned.

\subsubsection{Per-memory weights against the best fixed ratio}
\label{app:acl2-decomp}

The learning-rate sweep of \cref{app:focus-lr} compares the per-memory LR-CUSUM with fixed-ratio counters on the same certificate, closed loop, at matched false retirements (\cref{tab:app-e1}). Delays here are naive means that count open episodes at their length so far (as in \cref{app:focus}); slower rules are censored more often, so this bias favours the fixed-ratio rules over the LR-CUSUM. With $k$ down to $0.1$, the closed-loop optimum lies below $1$ on four of five models ($k=0.75$, $0.25$, $0.1$ and $0.25$; $25$ for Qwen3-8B), consistent with the predicted $1/\rho$ of $0.67$--$2.14$ under near-total staleness. The LR-CUSUM remains $5.8$, $5.9$, $2.0$, $1.3$ and $2.0\times$ faster than the best tested ratio. Paired intervals of this ratio at each model's best $k$ exclude $1$ for Qwen3-8B ($5.8$ [$3.2$, $9.4$]), Qwen3-4B ($5.9$ [$2.3$, $11.2$]) and OLMo-3-7B ($2.0$ [$1.1$, $4.6$]), but not for Granite ($2.0$ [$0.9$, $3.4$]) or Mistral ($1.3$ [$0.9$, $3.4$]) (\texttt{paper\_figs/acl3\_bestk.py}). The weights the LR-CUSUM assigns differ across memories and models: the median per-memory weight ratio is $1.90$, $1.33$, $0.92$, $0.50$ and $0.67$ (Qwen3-8B to OLMo-3-7B).

\subsection{Aligned safety, same-information baselines, decoding variance and natural edits}
\label{app:r3}

The replays of \cref{app:r3-safety,app:r3-info} reuse the recorded outcome tables of the five models in the main setting (30 streams per model and policy; no shard is missing). Intervals are 95\% template-clustered bootstraps; pooled rows resample templates jointly across models.

\subsubsection{Safety scored as the certificate tests it}
\label{app:r3-safety}

\paragraph{Scoring.} Under the standard scoring of \textsc{KQA-Stream}, a macro answer counts as correct only if the query was also routed to the right template, so a template's reliability averages over misrouted queries, which the certificate's paired test does not see (\cref{app:acl2-epoch}). Here a one-step answer is correct if and only if it equals the gold answer (\emph{answer-only} scoring). A candidate is then \emph{bad for the task} if its answer-scored reliability is below $\rstar=a-(m-1)/L$ ($L=20\bar m$), and \emph{bad under its own null} if the exact drift $\E Z$ of its paired utility on its routed queries is $\le0$ (with the certificate's own $L$, $B$ and clipping; build instance excluded). An epoch is an interval between answer-changing edits of a template, and it is at risk if a candidate that is bad at some version of it is under test. We count three events: (i)~a \emph{false certification}, a fold of a candidate that is bad under its own null at the fold; it is \emph{on null evidence} if the null held at every version of the fold's test segment (the event bounded by \cref{prop:app-ledger}(i)--(ii) and \cref{prop:app-detect}(ii)); (ii)~a task-level bad fold, below break-even; (iii)~a wrong answer among the one-step answers served.

\begin{table}[h]
\centering
\small
\setlength{\tabcolsep}{2.2pt}
\resizebox{\columnwidth}{!}{%
\begin{tabular}{llrrrrr}
\toprule
& $\delta$ & (i) own null & on null ev. & (ii) task & (iii) wrong & stale \\
\midrule
\multirow{5}{*}{\textbf{restart only}} & .01 & 0/6704 & 0 & 0/6693 & 11.2\% & 69 \\
& .05 & 0/6657 & 0 & 0/6645 & 11.6\% & 75 \\
& .1 & 0/6620 & 0 & 0/6611 & 11.7\% & 78 \\
& .2 & 2/6554 & 2 & 2/6546 & 11.8\% & 82 \\
& .5 & 12/6401 & 12 & 10/6396 & 12.8\% & 103 \\
\midrule
\multirow{5}{*}{\textbf{no notice}} & .01 & 6/6471 & 0 & 5/6458 & 10.7\% & 70 \\
& .05 & 12/6411 & 0 & 13/6396 & 11.2\% & 80 \\
& .1 & 13/6379 & 0 & 15/6365 & 11.4\% & 85 \\
& .2 & 26/6319 & 2 & 29/6304 & 11.7\% & 92 \\
& .5 & 58/6122 & 19 & 59/6109 & 12.5\% & 116 \\
\bottomrule
\end{tabular}}
\caption{Safety under answer-only scoring (pooled, 150 streams per row). (i),~(ii): epochs with the event / at-risk epochs; on null ev.: epochs with a false certification on null evidence; (iii): wrong answers among served one-step answers; stale: stale answers per $10^4$ queries. Restart only: the restart-only certificate; no notice: discard on own alarm; ours (the recommended deployment): \cref{app:r4-main}. Bold: ours (both rows).}
\label{tab:app-r3-safety}
\end{table}

\paragraph{Results.} The restart-only certificate makes no false certification at $\delta\le0.1$ on any model ($0$ of $6{,}657$ at-risk epochs at $\delta=0.05$; \cref{tab:app-r3-safety}, \cref{fig:epoch}); at $\delta=0.2$ and $0.5$ the shares are $0.0003$ [$0$, $0.0009$] and $0.0019$ [$0.0006$, $0.0036$], all on null evidence and in fresh test segments, as \cref{prop:app-ledger}(ii) allows. Intervals clustered by relation or by edit sequence are comparable. The two notions of ``bad'' nearly coincide: at $\delta=0.05$, of $27{,}339$ candidates, $3{,}611$ are bad for the task, $3{,}619$ under their own null and $3{,}571$ under both; the task-level counts in \cref{tab:app-r3-safety} differ from the own-null counts by at most three epochs. Without notices, no false certification rests on null evidence at $\delta\le0.1$; every task-level bad fold at these levels ($5$, $13$ and $15$) spans an edit, i.e.\ its test segment began before the change, which is the residual risk that \cref{prop:app-detect}(iii) bounds. Every point estimate of both certificates is below $\delta$.

\paragraph{False certification is not the error rate.} At every $\delta$, $11.2$--$12.8\%$ of the one-step answers that the restart-only certificate serves are wrong ($7.6$--$8.7\%$ on Qwen3-8B, $13.2$--$17.1\%$ on Mistral-7B). Stale answers are only $14$--$17\%$ of them; the rest are errors of macros that were above break-even when folded, which a fold accepts by design ($r>\rstar$ allows $r<1$). $\delta$ therefore controls whether a bad memory is trusted, not how often a trusted memory errs; the second is set by $\rstar$ and by the detection delay of \cref{sec:hebbian}.

\paragraph{Standard scoring.} With routing errors counted as wrong answers, the restart-only certificate has bad folds in $0.0025$ [$0.0011$, $0.0043$] of at-risk epochs at $\delta=0.05$ ($29$ of $11{,}670$), all of them candidates with a positive paired-utility drift, and discard-on-alarm has $0.0008$ [$0.0003$, $0.0014$] ($9$ of $11{,}375$; \cref{app:acl2-epoch,app:acl2-nonotice}, \cref{fig:app-epoch-std}). The order of the two certificates flips between the scorings because the routing convention creates most of the restart-only certificate's standard-scoring bad folds, whereas the no-notice certificate's bad folds are carried across edits under both. Both scorings stay below $\delta$ at every level.

\subsubsection{Same-information baselines}
\label{app:r3-info}

The restart-only certificate receives an edit notice: the templates whose gold answer an edit changes. It uses the notice only to restart candidates under test; folded macros are retired by the LR-CUSUM alone. The baselines below receive the same notices, on the same streams, audits and edits, with the same certificate parameters ($\delta=0.05$, seven bets, $h=7$) where they certify. \emph{Invalidate on notice} drops the folded macro or the candidate under test of a notified template; a later verified live trace forms a fresh candidate, tested under the lifetime ledger, and folded macros are retired by notices only or also by their LR-CUSUM. Because carried-over candidates are discarded at every announced change, \cref{prop:app-ledger}(iv) gives per-epoch safety for those epochs. \emph{Drop only} sends the template back to live reasoning for the rest of the stream. The answer caches evict notified entries: a verified-answer cache, which stores a question's live answer only when that run was audited and verified correct and returns it on an exact repeat, and the vCache-style cache of \cref{app:focus-vcache}. To model missed notices, each answer-changing edit is announced independently with probability $f$ (the same draws for all policies). The \emph{compiled plan} row is the restart-only certificate on compiled plans (with restarts).

\begin{table}[h]
\centering
\small
\setlength{\tabcolsep}{2.4pt}
\resizebox{\columnwidth}{!}{%
\begin{tabular}{lrrrl}
\toprule
Policy ($f=1$) & Acc & Calls & Stale & $\Delta$Acc vs restart only (points) \\
\midrule
\textbf{restart only (restart-only certificate)} & .430 & 3.36 & 63.3 & -- \\
invalidate & .430 & 3.20 & 0.0 & $-0.06$ [$-0.94$, $+0.80$] \\
\textbf{ours: invalidate + LR} & .433 & 3.38 & 0.0 & $+0.24$ [$-0.24$, $+0.81$] \\
drop only & .332 & 3.84 & 0.0 & $-9.8$ [$-12.9$, $-7.0$] \\
verified-answer cache & .188 & 3.43 & 0.0 & $-24.3$ [$-28.1$, $-20.3$] \\
vCache-style & .186 & 3.50 & 0.0 & $-24.5$ [$-28.3$, $-20.5$] \\
\textbf{compiled plan, no restarts} & .461 & 3.20 & 12.3 & $+3.1$ [$+2.5$, $+3.8$] \\
\textbf{compiled plan} & .456 & 3.25 & 11.8 & $+2.5$ [$+2.0$, $+3.2$] \\
\bottomrule
\end{tabular}}
\vspace{4pt}
\resizebox{\columnwidth}{!}{%
\begin{tabular}{lrrrr}
\toprule
Stale per $10^4$ queries at notice fraction $f$ & 1 & 0.8 & 0.5 & 0 \\
\midrule
\textbf{restart only (restart-only certificate)} & 63.3 & 65.3 & 67.8 & 66.3 \\
invalidate & 0.0 & 124.5 & 324.5 & 809.6 \\
\textbf{ours: invalidate + LR (recommended)} & 0.0 & 13.3 & 28.2 & 66.3 \\
\textbf{compiled plan} & 11.8 & 12.0 & 12.4 & 12.3 \\
\bottomrule
\end{tabular}}
\caption{Same-information baselines, pooled over five models (150 streams per row; standard scoring). Top: perfect notices, paired differences with 95\% template-bootstrap intervals. Bottom: missed notices. Bold: ours.}
\label{tab:app-r3-info}
\end{table}

\paragraph{Results.} With perfect notices, invalidation matches the restart-only certificate's accuracy and serves no stale answers (\cref{tab:app-r3-info}); its $J_{20}$ differs by $-0.12$ [$-0.82$, $+0.51$]. Dropping without re-forming loses $9.8$ points, and the answer caches stay at live accuracy ($0.19$, the pooled live accuracy) because they answer only $10$--$13\%$ of queries in one step. With missed notices, invalidation alone degrades fast: at $f=0$ every one of its $218$ stale episodes is still in service at stream end, and its accuracy falls by $4.4$ points [$2.2$, $6.9$]. Adding the LR-CUSUM keeps stale answers at or below the restart-only certificate's at every $f$, and at $f=0$ the two policies coincide. The share of at-risk epochs with a bad fold rises as notices are missed, because carried-over evidence is no longer discarded (\cref{prop:app-ledger}(iv)): $0.0026$, $0.0030$, $0.0036$ and $0.0047$ for invalidate + LR at $f=1$, $0.8$, $0.5$ and $0$ (restart-only certificate: $0.0025$--$0.0047$), all below $\delta$. On Qwen3-8B, invalidate + LR adds $0.6$ points [$0.15$, $1.1$] over the restart-only certificate (\cref{tab:realllm}).

\paragraph{Compiled plans with restarts.} With restarts, compiled plans keep $+2.5$ points over macros and $11.8$ stale answers per $10^4$ queries, and they halve the bad folds of plans without restarts: $0.0032$ against $0.0064$ of at-risk epochs ($29$ of $9{,}179$ against $59$ of $9{,}172$) and $0.29$ against $0.63$ bad folds per stream, for $0.6$ accuracy points. Per model the gain over macros is $1.2$--$4.0$ points (\cref{tab:app-models}).

\subsubsection{Decoding variance}
\label{app:r3-decode}

Each instance--version of the main runs is decoded once, so the intervals elsewhere resample templates, streams and edit sequences, not decoding randomness. To measure the latter we re-decoded a Qwen3-8B subset (60 templates $\times$ 10 instances, edit sequence 101, knowledge-base versions 0, 8, 16 and 24) with three decoding seeds. Live accuracy per version is $0.368/0.343/0.322/0.312$, $0.367/0.338/0.320/0.312$ and $0.368/0.335/0.313/0.305$ for the three seeds: the standard deviation across seeds is at most $0.0044$ per version, against a template-bootstrap 95\% half-width of $0.09$--$0.10$ for one seed. Two seeds give the same final answer on $91.9\%$ of instance--versions and the same correctness on $97.6\%$. Replays of each seed's tables (20 streams each) rank the six policies (live, fold once, ExpeL- and ACE-style counters, the restart-only certificate and its no-notice variant) in the same order of accuracy for every seed. The restart-only certificate's accuracy is $0.447$--$0.452$ and its stale answers $32$--$33$ per $10^4$ queries, against $698$--$771$ for fold once and the counters. On the matched population of its folded macros, the LR-CUSUM leaves $0\%$ unretired and the counters $57$--$75\%$, a delay ratio of $131$--$155\times$ on this small subset (32--35 episodes). Decoding noise is thus two orders of magnitude below the template variation that our intervals already include, and no conclusion changes on the subset.

\subsubsection{Natural edits from Wikidata temporal qualifiers}
\label{app:natural-edits}

\begin{figure*}[t]
\centering
\includegraphics[width=\textwidth]{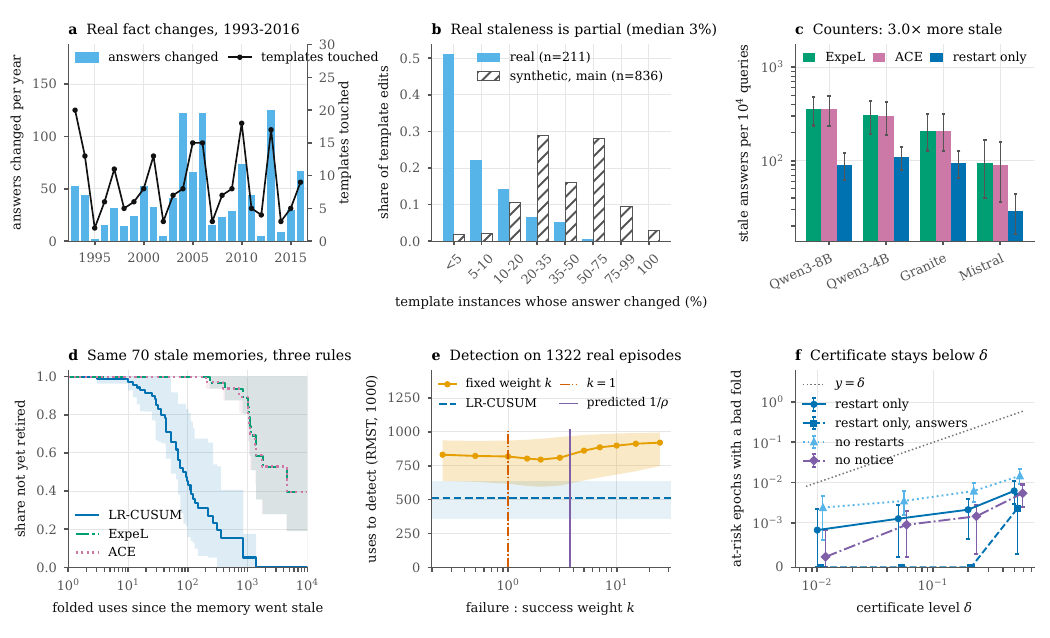}
\caption{Natural-edit transfer (Wikidata temporal qualifiers, 1992--2016; four models, 30 streams each). \textbf{(a)}~Real fact changes per year. \textbf{(b)}~Share of a template's instances whose answer changes in one edit: half of the real changes touch under $5\%$ of the template, most synthetic ones $20$--$75\%$. \textbf{(c)}~Stale answers per $10^4$ queries: counters serve $3.0\times$ more than the restart-only certificate. \textbf{(d)}~The same $70$ stale memories under three retirement rules (Kaplan--Meier). \textbf{(e)}~Delay to detect at $3$ false alarms per $10^4$ healthy uses on $1{,}322$ real episodes: fixed failure weights against the per-memory LR-CUSUM. \textbf{(f)}~Share of at-risk epochs with a bad fold against $\delta$ (``answers'': answer-only scoring; no notice: discard at the own alarm). Bands and bars: 95\% template bootstrap.}
\label{fig:app-natural-edits}
\end{figure*}

\paragraph{Timeline.} Of the $207{,}851$ facts of the knowledge base, $15{,}290$ ($7.4\%$) carry time information ($7{,}017$ start or end dates and $8{,}273$ dated events). The $300$ main templates are too sparse for a natural timeline: only $321$ of their instances change answer between 1992 and 2016 while staying answerable. We therefore built a natural subset with the construction rules of \cref{app:experiments}: of $2{,}250$ candidate paths, $1{,}769$ are valid at every date, $88$ change at least $5\%$ of their answers, and $72$ templates ($68$ new, $4$ from the main set; $4{,}304$ instances) remain after the diversity caps. The stream applies the 1992 state at query $0$ and one knowledge-base version per year, 1993--2016, at uniform stream times ($8{,}000$ queries per year). Real changes are small and partial: $43.6$ answer changes and $8.8$ templates touched per year; in the median template-year only $3.3\%$ of the template's answers change, and no change is total (main synthetic edits: median $40\%$, $3\%$ total; \cref{fig:app-natural-edits}b). We ran four models (Qwen3-8B, Qwen3-4B, Granite-3.3-8B, Mistral-7B); OLMo-3-7B was not run on this timeline. The offline exact check reproduced all $111{,}904$ traces at all $26$ versions. Policies and settings are those of the main replays ($30$ streams of $200$k queries per model, $\delta=0.05$).

\begin{table}[h]
\centering
\small
\setlength{\tabcolsep}{2.4pt}
\resizebox{\columnwidth}{!}{%
\begin{tabular}{lrrrrr}
\toprule
Policy (pooled) & Acc & Calls & Stale & Bad/stream & $J_{20}$ \\
\midrule
live always & .175 & 4.47 & 0 & 0 & 79.3 \\
fold once & .340 & 2.60 & 231 & 8.5 & 62.6 \\
ExpeL-style counter & .391 & 3.39 & 241 & 149 & 59.0 \\
ACE-style counter & .391 & 3.52 & 238 & 207 & 59.2 \\
\textbf{restart only (restart-only certificate)} & .356 & 4.03 & 80.5 & 0.10 & 62.8 \\
\textbf{ours (recommended)} & .337 & 4.22 & 0 & 0.11 & 64.8 \\
\textbf{no notice (own detector)} & .384 & 3.84 & 99.5 & 0.03 & 60.2 \\
\textbf{compiled plan} & .374 & 3.95 & 2.8 & 0.10 & 61.2 \\
\bottomrule
\end{tabular}}
\caption{Natural edits, pooled over four models (30 streams of 200k queries each; standard scoring). Stale: per $10^4$ queries. Bold: ours.}
\label{tab:app-natural-edits}
\end{table}

\paragraph{Results.} \emph{Counters.} Counters serve $3.0\times$ [$2.2$, $4.0$] more stale answers than the restart-only certificate ($4.0$, $2.8$, $2.2$ and $3.2\times$ per model) and make $149$--$207$ bad folds per stream, against $0.10$. Replayed on the same macros as in \cref{app:acl3}, $70$ of the $142$ stale episodes remain after removing those that some rule retired before the change; the counters leave $84.3\%$ [$46.3$, $100$] in service at stream end and the LR-CUSUM $37.1\%$ [$4.5$, $55.3$], with a restricted-mean delay ratio of $20.5\times$ [$13.2$, $72.9$] and $405$ against $30$ stale answers per episode. The LR-CUSUM leaves more episodes open here than on synthetic edits because real staleness is mostly partial and $26$ of its $70$ episodes are cut off by the end of the stream; \cref{app:r4-natural-sens} reports the sensitivity to the event set and to leaving out one year or one relation. \emph{Weights.} On $1{,}322$ single-change episodes ($35$ templates; mean $q_0=0.121\to q_1=0.330$, optimal weight $1/\rho=3.7$), the per-memory LR-CUSUM is faster than the best fixed weight at every matched false-retirement rate: $1.08\times$ [$1.01$, $1.32$], $1.56\times$ [$1.34$, $1.85$] and $1.14\times$ [$0.93$, $1.56$] at $1$, $3$ and $10$ false alarms per $10^4$ healthy uses ($1.72\times$ on the $907$ partial episodes at $3$). The best fixed weight is about $2$, not the $5$--$7$ of the synthetic partial episodes (\cref{app:partial}), and on the $170$ near-total episodes the symmetric rule is best, as \cref{cor:lr} predicts ($1/\rho=0.73$). \emph{Safety.} Bad-fold epochs of the restart-only certificate are $0.0008$, $0.0013$, $0.0021$ and $0.0062$ of at-risk epochs at $\delta=0.01$, $0.05$, $0.2$ and $0.5$ ($4/4{,}790$, $6/4{,}770$, $10/4{,}744$, $29/4{,}686$), and $0$ up to $\delta=0.2$ under answer-only scoring. Every point estimate is below $\delta$; one per-model upper bound exceeds it (Qwen3-4B at $\delta=0.01$, $2$ of $690$ epochs). \emph{Accuracy and cost.} The restart-only certificate adds $18.1$ points [$14.7$, $21.7$] over live reasoning with $10\%$ fewer calls (on Mistral-7B it uses slightly more, $5.31$ against $5.04$), and certified compiled plans serve $2.8$ stale answers per $10^4$ queries. The counters are $3.5$ points more accurate than the restart-only certificate ($5.4$ than the recommended deployment, \cref{app:r4-natural-sens}), with lower loss, because when real staleness is this mild, folding everything without certification pays; this is the price of safety of \cref{sec:limit}, bought here with $149$--$207$ bad folds per stream. The synthetic ratios of the main text ($7.0$--$13.5\times$, $60\times$, a best weight of $5$--$7$) are not claimed for natural edits.

\subsection{Silent changes of the null, the recommended deployment, natural-edit sensitivity}
\label{app:r4}

\subsubsection{When the null changes without a notice}
\label{app:r4-silent}

\paragraph{What triggers a notice.} In the replays, the notice of version $v$ lists the templates $k$ for which the gold answer of some instance routed to $k$ differs between versions $v-1$ and $v$ (\texttt{notice\_tables}); nothing else restarts a test or invalidates a memory. Epochs are the intervals between such notices. The certificate's null, however, is the paired-utility drift $\E Z\le0$, which depends on live reasoning as well: with $C^L-C^M=2$, $L=50$ and a macro of reliability $0.9$, a live accuracy of $0.8$ gives $\E U=+7$, and $0.95$ gives $\E U=-0.5$. Live reasoning can thus make a memory useless without any change of its answers or of the gold answers. \Cref{prop:app-ledger}(iv) therefore gives per-epoch safety only if notices cover every change of the null, including changes of live reasoning; across a change without a notice, the evidence banked before it is covered only by the carried-over bound $\min\{1,\delta_jK_c\}$ of \cref{prop:app-ledger}(iii). Two events must be kept apart: a candidate that is null \emph{at} the fold, and a test segment that was null \emph{throughout} (``on null evidence''). Only the second is bounded by $\delta$ without further assumptions.

\paragraph{How often it happens.} Over the three edit sequences and five models, $16{,}827$ of $107{,}496$ template-versions ($15.7\%$) change some routed instance's live trace (answer, correctness or call count) without a notice, $1{,}828$ change a live correctness and $1{,}715$ the routed live accuracy; $5{,}179$ carry a notice (\cref{tab:app-r4-silent}, top). Within the replays ($\delta=0.05$, 150 streams per row), we tracked the exact drift of every candidate's own test at every version of its life. \Cref{tab:app-r4-silent} (bottom) counts the candidates whose null flips without a notice while they are under test, and the folds whose test segment spans such a flip. Flips are rare ($0.3$--$0.7\%$ of candidates), and of $7{,}385$--$9{,}835$ folds per row at most $2$ span one, all of candidates that are useful at the fold; no fold of a candidate that is null at the fold occurs at $\delta=0.05$. So on these streams the per-epoch statement holds as measured, but the assumption it needs is not automatic.

\begin{table}[h]
\centering
\small
\setlength{\tabcolsep}{2.2pt}
\resizebox{\columnwidth}{!}{%
\begin{tabular}{lrrrr}
\toprule
Model & template-versions & notice & silent live-trace change & silent live-accuracy change \\
\midrule
Qwen3-8B & 21{,}528 & 884 & 1{,}392 & 286 \\
Qwen3-4B & 21{,}600 & 867 & 3{,}124 & 363 \\
Granite-3.3-8B & 21{,}456 & 966 & 4{,}719 & 448 \\
Mistral-7B & 21{,}384 & 1{,}333 & 5{,}328 & 385 \\
OLMo-3-7B & 21{,}528 & 1{,}129 & 2{,}264 & 233 \\
\bottomrule
\end{tabular}}
\vspace{4pt}
\resizebox{\columnwidth}{!}{%
\begin{tabular}{llrrrrrr}
\toprule
Policy & scoring & candidates & with flip & to null / useful & folds & span a flip & null at fold \\
\midrule
\textbf{restart only} & standard & 31{,}593 & 226 & 97 / 155 & 7{,}385 & 2 & 0 \\
\textbf{restart only} & answer & 27{,}339 & 192 & 82 / 133 & 8{,}632 & 0 & 0 \\
\textbf{ours} & standard & 42{,}306 & 132 & 68 / 80 & 8{,}294 & 2 & 0 \\
\textbf{ours} & answer & 37{,}928 & 114 & 58 / 75 & 9{,}835 & 0 & 0 \\
\bottomrule
\end{tabular}}
\caption{Silent changes of the null. Top: template-versions (three sequences, routed templates) with a notice and without one but with a change of live reasoning. Bottom: candidates whose own null flips without a notice while under test (flip events by direction), and folds whose test segment spans such a flip; $\delta=0.05$, 150 streams per row, pooled over five models. Bold: ours (all policies).}
\label{tab:app-r4-silent}
\end{table}

\subsubsection{The recommended deployment}
\label{app:r4-main}

The main tables and \cref{fig:epoch} report the recommended deployment (``ours'': the restart-only certificate plus invalidation on notice, invalidate + LR of \cref{app:r3-info}); restart only (the restart-only certificate) is the comparison row. Under answer-only scoring, ours makes $0$ own-null false certifications at $\delta\le0.1$ ($0/5{,}814$, $0/5{,}779$, $0/5{,}758$ at-risk epochs at $\delta=0.01$, $0.05$, $0.1$), $1/5{,}718$ at $\delta=0.2$ and $11/5{,}614$ ($0.0020$ [$0.0006$, $0.0038$]) at $\delta=0.5$, all on null evidence; task-level bad folds are $0$, $0$, $0$, $1$ and $8$. Of the one-step answers it serves, $9.6$--$10.6\%$ are wrong (restart only: $11.2$--$12.8\%$), and none is stale. In \cref{tab:models} its stale answers are reported with every notice missed (the counters' information; then it coincides with the ledger certificate without restarts); its bad folds per stream are those with perfect notices ($0.00$, $0.07$, $0.43$, $0.63$, $0.30$). The matched retirement comparison (\cref{fig:km}) uses the macros that restart only folds, because their folded macros, like ours after a missed notice, are retired by detection alone.

\subsubsection{Natural edits: sensitivity}
\label{app:r4-natural-sens}

All numbers below reuse the natural replays of \cref{app:natural-edits}; the recommended deployment was replayed in addition (same streams). \emph{Recommended deployment.} It serves no stale answers, has bad folds in $0.0007$, $0.0013$, $0.0029$ and $0.0060$ of at-risk epochs at $\delta=0.01$, $0.05$, $0.2$, $0.5$ ($4/5{,}493$, $7/5{,}498$, $16/5{,}492$, $33/5{,}487$), and $0.11$ bad folds per stream, but it is $1.9$ points [$1.0$, $2.8$] less accurate than restart only ($0.337$ against $0.356$): when an edit changes $3\%$ of a template's answers, invalidation discards a memory that is still mostly right.

\emph{Event set.} The matched retirement comparison uses the $70$ of $142$ stale episodes that no rule retired before onset. On all $142$, counting a pre-onset retirement as detection at $0$ uses, the counters leave $70.4\%$ (ExpeL) and $62.7\%$ (ACE) in service at stream end against $26.8\%$ for the LR-CUSUM, with a delay ratio of $13.6\times$ [$5.4$, $88.7$]; each rule on its own episodes gives $90.1\%$, $89.0\%$ and $45.2\%$ unretired and a ratio of $10.3\times$.

\emph{Leave one out.} Leaving out the episodes whose onset falls in one year, the counters leave $57.7$--$100\%$ and the LR-CUSUM $3.8$--$45.6\%$ in service, and the delay ratio is $18.5$--$32.0\times$; leaving out the templates that use one changing relation (21 relations), $56.0$--$100\%$ against $21.2$--$45.6\%$ and $17.2$--$38.1\times$. Most common episodes ($44$ of $70$) start in the last year, 2016, and are censored early; without them the shares are $58\%$ against $4\%$, so the unretired shares depend on censoring, while the delay ratio is stable. Leaving out one relation, the counters serve $2.5$--$3.2\times$ more stale answers than restart only ($2.3$--$2.9\times$ than the ledger without restarts, i.e.\ ours with every notice missed), every certificate's bad-fold share stays below $\delta$ at every level (largest: $0.0080$ at $\delta=0.5$), and the per-memory LR-CUSUM stays faster than the best fixed weight at $1$, $3$ and $10$ false alarms per $10^4$ healthy uses ($1.01$--$1.17\times$, $1.33$--$1.74\times$, $1.04$--$1.29\times$). The weight comparison uses the $1{,}322$ single-change episodes; episodes with several changes do not have one change point and are not used.

\subsubsection{Which variant for which goal}
\label{app:guide}

\begin{table}[h]
\centering
\small
\setlength{\tabcolsep}{2.2pt}
\resizebox{\columnwidth}{!}{%
\begin{tabular}{@{}>{\raggedright\arraybackslash}p{2.3cm}>{\raggedright\arraybackslash}p{3.0cm}>{\raggedright\arraybackslash}p{3.3cm}@{}}
\toprule
Variant & Use when & Evidence \\
\midrule
\textbf{Recommended: invalidate on notice + certify with restarts + LR-CUSUM} & false deployments must be bounded and notices may be incomplete & 0 stale with notices; 13--66 per $10^4$ when 20--100\% are missed (\cref{tab:app-r3-info}); no false certification at $\delta\le0.1$; code: $2.0$ stale per $10^4$ with only requirement notices, bad folds in $1.3\cdot10^{-5}$ of epochs under test at $\delta=0.05$ (\cref{app:codestream}) \\
\textbf{Restart only} & changes are mild and partial, notices reliable & $+1.9$ points over invalidation on natural edits (\cref{app:r4-natural-sens}) \\
\textbf{Compiled plans + certificate} & the derivation can be re-executed cheaply on current facts & $+2.5$ points over restart-only macros, $12$ stale per $10^4$ (\cref{tab:app-r3-info}) \\
Verified answer cache & exact repeats, reliable notices & live accuracy; $13\%$ of queries in one step \\
Counters (ExpeL/ACE) & staleness is mild and folding everything pays; never when stale answers are costly & natural edits: $+3.5$ points, $149$--$207$ bad folds per stream \\
\bottomrule
\end{tabular}}
\caption{Decision guide (pooled evidence; main text: \cref{sec:experiments}). Bold: ours.}
\label{tab:app-guide}
\end{table}

\subsection{A code domain with an exact verifier: CodeStream}
\label{app:codestream}

CodeStream tests the theory on a second real domain, chosen so that the verifier is exact and memories are what tool-building agents store: a verified program, reused as a tool \citep{wang2024voyager,cai2024large,wang2024trove,yuan2024craft}. It writes the same potential-outcome tables as \textsc{KQA-Stream}, so every policy, metric and replay setting of \cref{app:experiments} runs on it unchanged; only the domain, the stream length and the cost unit differ. The setting was fixed after a two-model pilot, before the full run.

\paragraph{Tasks, queries and the verifier.} We use 521 EvalPlus tasks \citep{liu2023evalplus}: 359 from MBPP+ (v0.2.0; \citealp{austin2021program}) and 162 from HumanEval+ (v0.1.10; \citealp{chen2021evaluating}). We exclude 21 tasks: 17 with a special oracle or a failing canonical solution, and 4 with fewer than 10 usable hidden inputs. A query is a task plus one hidden plus-test input, 20 per task ($10{,}420$ queries); its answer is the function's output. The verifier is EvalPlus's own check against the reference output of the task version in force: exact equality, or closeness within the task's tolerance for floating-point outputs. Each (program, version, input set) runs once and is cached, so every policy sees the same verdicts; re-executing the $6{,}330$ sandbox runs of the table build in fresh processes changes $57$ of $126{,}600$ verdicts ($0.05\%$), from programs whose output is not deterministic (e.g.\ set order).

\paragraph{Live reasoning and memories.} The live reasoner is a solve--test--repair agent. It writes the function from the task prompt (temperature $0.7$, at most 768 new tokens; Qwen3 with thinking disabled), runs it on the visible tests (the EvalPlus base inputs) in a sandbox of the current version, and, if a test fails, gets the failing call, the expected output and the output or exception it produced. It makes at most three attempts; the final program is run on the query input. Each task has 20 samples, one per query, so live outcomes of different queries are independent samples. A \emph{memory} is the final program of a live trace whose answer was audited correct; folded, it is executed directly on later queries of the same task, in the current environment. Memories are thus compiled plans in the sense of \cref{sec:experiments}: an environment change can break them, and a requirement revision changes the answers they are checked against. Some programs pass the visible tests but fail other hidden inputs, EvalPlus's own finding that base tests under-specify tasks; they are the domain's candidates that are bad at formation, so certification has real work to do.

\paragraph{Edit catalogue.} Each timeline has 14 edits (15 versions) at uniform stream times, in the order A1, S1, A2, A3, S2, A4, S3, A5, A6, S4, A7, S5, A8, S6. \emph{Requirement revisions} S1--S6 are the main source of staleness. Each batch revises 22 tasks ($6\%$ of the revisable tasks; 132 per timeline, each at most once) by appending one natural post-processing change to the prompt as ``Requirement update: \dots'': sort, deduplicate, lowercase, reduce modulo $10^9{+}7$, clamp negatives to $0$, round to two decimals, return \texttt{'Yes'}/\texttt{'No'}, or return a list instead of a tuple. The reference output changes exactly on the inputs whose output the change affects, so staleness is partial or total by construction from the real input distribution (in the design-time catalogue the share of a revised task's queries whose answer changes ranged from $0.05$ to $1$, median about $0.6$). The three timelines (seeds 101, 202, 303) revise different tasks. \emph{API edits} A1--A8 re-enact documented Python breaking changes in the sandbox (\cref{tab:app-cs-edits}); they change no reference output, and a memory goes stale only if its own program changes behavior. Error messages carry the migration hint, as real deprecations do, so live reasoning can repair from test feedback.

\begin{table}[h]
\centering
\small
\setlength{\tabcolsep}{2.4pt}
\resizebox{\columnwidth}{!}{%
\begin{tabular}{@{}llll@{}}
\toprule
Id & Edit (re-enacted) & Precedent & Programs broken \\
\midrule
A1 & \texttt{typing.List} etc.\ removed & PEP 585 deprecation & 55 / 68, total \\
A2 & \texttt{re.sub/split}: count, flags keyword-only & deprecated in 3.13 & 0 / 0 \\
A3 & \texttt{str} patterns ASCII-only by default & Python 2/3 default & 0 / 0 \\
A4 & int--str conversion capped at 640 digits & 3.11 (CVE-2020-10735) & 1 / 1, partial \\
A5 & \texttt{round} half away from zero & Python 2 & 2 / 2, partial \\
A6 & \texttt{math.floor/ceil} return float & Python 2 & 1 / 1, total \\
A7 & \texttt{Counter.most\_common} tie order & dict order before 3.7 & 0 / 0 \\
A8 & \texttt{zip} strict by default & PEP 618 option & 3 / 3, partial \\
\bottomrule
\end{tabular}}
\caption{API edits of CodeStream. Programs broken: distinct verified programs whose behavior the edit changes, out of $1{,}494$ (Qwen3-4B) / $1{,}380$ (Qwen3-8B) in the two-model pilot.}
\label{tab:app-cs-edits}
\end{table}

\paragraph{Notices.} A requirement revision bumps the task's version: this is the answer-changing notice of the main experiments. API edits are unannounced: they are the domain's unannounced changes, and for them the LR-CUSUM is the only safety net. They also change live reasoning (in the pilot, after A1 the mean number of attempts on the 210 affected Qwen3-4B traces rose from $1.21$ to $2.13$) without changing any answer, a silent change of the certificate's null in the sense of \cref{app:r4-silent}. The \emph{changelog} variant additionally invalidates, at each API edit, every memory whose source mentions a changed symbol, a static match that real maintainers would make: it is exact for removals but flags many unaffected programs for semantic edits.

\paragraph{Streams, policies and statistics.} Streams have $T=10^6$ queries, Zipf demand over tasks ($s=1.2$), audits with $p=0.1$, $L=20\bar m$ and $\delta=0.05$; the certificates, the bet grid and the LR-CUSUM threshold $h=7$ are those of the main experiments. Each model is replayed on 10 stream seeds per timeline. All memory policies store programs. Labels: \emph{fold once}; ExpeL- and ACE-style counters; \emph{per candidate}, the per-candidate certificate of \cref{tab:algos}, without notices or restarts, i.e.\ the counters' information; \emph{restart only}, the restart-only certificate, which restarts at requirement notices; \emph{ours}, the recommended deployment (invalidate on notice and re-form, ledger, restarts, LR-CUSUM), as in the main text; and \emph{ours+changelog}. A \emph{stale answer} is a folded answer that is wrong although the same memory answered this query correctly when it was formed. 95\% intervals resample templates (clusters) with 2{,}000 bootstrap draws, pooled over seeds and timelines; pooled intervals resample (model, template) clusters.

\paragraph{Models and compute.} All five models ran all three timelines, so pooled numbers rest on 150 streams, 30 per model; every stream has the same weight.

\paragraph{Two cost accountings.} Writing, testing and repairing a program costs more than running a stored one, and the domain admits two units. \emph{Step accounting}, our main setting and faithful for tool libraries, charges a live query its attempts (LLM calls) plus one execution step, and a folded memory one execution step; live reasoning costs $2.68$ steps per query. \emph{Call accounting}, the unit of \textsc{KQA-Stream}, charges live reasoning its LLM calls ($1.68$ per query) and a folded memory one call. $J_{20}$ is computed in each unit with $L=20\bar m$, as in the main text. Every cost statement below names its unit.

\begin{figure*}[t]
\centering
\includegraphics[width=\textwidth]{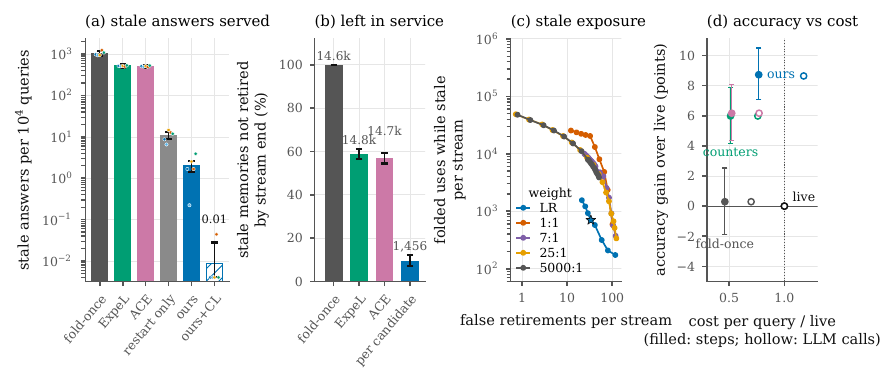}
\caption{CodeStream (EvalPlus; five models, 150 streams of $10^6$ queries). \textbf{(a)}~Stale answers per $10^4$ queries (step accounting; dots: models); ours+CL: ours with changelog invalidation. \textbf{(b)}~Share of each policy's own stale memories not retired by stream end (numbers: stale episodes). \textbf{(c)}~Stale exposure at matched false retirements: folded uses while stale against false retirements, both per stream, for the per-memory LR-CUSUM (threshold sweep; star: $h=7$) and fixed failure:success weights, all on the same certificate (weights $1{:}1$--$5000{:}1$ were run; above $25{:}1$ the curves coincide with $5000{:}1$, the no-forgetting limit). \textbf{(d)}~Accuracy gain over live reasoning (points) against cost per query relative to live reasoning, in steps (filled) and in LLM calls (hollow). Bars: 95\% template-clustered bootstrap; in (d) on the filled markers only, since $\Delta$acc in LLM calls is within $0.1$ points of it.}
\label{fig:app-codestream}
\end{figure*}

\begin{table*}[t]
\centering
\small
\setlength{\tabcolsep}{3.8pt}
\begin{tabular}{@{}lrrrrrrrrrr@{}}
\toprule
& & & \multicolumn{3}{c}{step accounting} & \multicolumn{3}{c}{call accounting} & & \\
\cmidrule(lr){4-6}\cmidrule(lr){7-9}
Policy & Acc & $\Delta$acc (points) & steps/q & $J_{20}$ & one-step & calls/q & $J_{20}$ & one-step & Stale & Unretired \\
\midrule
live always & .762 & $0$ & 2.68 & 16.0 & 0 & 1.68 & 10.21 & 0 & 0 & -- \\
fold once & .765 & $+0.3$ [$-1.9$, $2.5$] & 1.24 & 14.1 & .94 & 1.17 & 9.35 & .94 & 1{,}043 & 100\% \\
ExpeL-style counter & .822 & $+6.0$ [$4.2$, $7.9$] & 1.37 & 11.1 & .90 & 1.27 & 7.44 & .90 & 518 & 59\% \\
ACE-style counter & .823 & $+6.2$ [$4.3$, $8.1$] & 1.40 & 11.0 & .89 & 1.29 & 7.40 & .89 & 510 & 57\% \\
\textbf{per candidate (no notices)} & .850 & $+8.9$ [$7.2$, $10.6$] & 2.01 & 10.2 & .63 & 1.95 & 7.19 & .33 & 10.1 & 10\% \\
\textbf{restart only} & .849 & $+8.7$ [$7.1$, $10.4$] & 2.06 & 10.4 & .61 & 1.97 & 7.26 & .32 & 11.0 & 9\% \\
\textbf{ours (recommended)} & .849 & $+8.7$ [$7.1$, $10.5$] & 2.06 & 10.3 & .61 & 1.97 & 7.27 & .32 & 2.0 & 6\% \\
\textbf{ours + changelog} & .849 & $+8.7$ [$7.0$, $10.4$] & 2.08 & 10.4 & .60 & 1.98 & 7.30 & .32 & 0.01 & -- \\
\bottomrule
\end{tabular}
\caption{CodeStream, pooled over five models (150 streams of $10^6$ queries). Accuracy and $\Delta$acc (paired, template-clustered 95\% intervals) in step accounting; in call accounting every $\Delta$acc is within $0.1$ points of these. One-step: share of queries answered by a folded memory. Stale: per $10^4$ queries (step accounting). Unretired: share of the policy's own stale episodes still in service at stream end (step accounting; $14{,}826$, $14{,}704$, $1{,}456$, $1{,}328$ and $96$ episodes for the counters, per candidate, restart only and ours; fold once never retires). Bold: ours.}
\label{tab:app-codestream}
\end{table*}

\paragraph{(a) Counters leave stale memories in service.} The counters serve $518$ [$471$, $567$] (ExpeL) and $510$ [$463$, $561$] (ACE) stale answers per $10^4$ queries, against $10.1$ [$8.8$, $11.4$] for the certificate with the counters' information (per candidate): $51\times$ pooled (95\% intervals within [$43$, $60$]) and $35$--$73\times$ on every model, with every per-model lower confidence bound at least $25$. Per folded answer, i.e.\ after dividing each policy's stale answers by its one-step coverage, the ratio is still $36\times$ pooled ($27$--$48\times$ per model), so the gap is not an artifact of the counters' higher coverage. Against restart only ($11.0$ [$9.0$, $13.2$]) the ratio is $47\times$ [$39$, $58$] (ExpeL) and $47\times$ [$38$, $57$] (ACE); against ours ($2.0$ [$1.4$, $2.7$]) it is $258\times$ [$191$, $363$] (ExpeL). Ours with changelog notices serves $0.01$ per $10^4$ (all on Mistral-7B). Each policy's own stale memories remain in service at stream end in $59\%$ [$56$, $61$] (ExpeL) and $57\%$ [$54$, $59$] (ACE) of cases, against $10\%$ [$7$, $12$] for per candidate; fold once retires none. These shares compare each rule on its own folded memories, not on a matched population as in \cref{fig:km}; the matched comparison is claim (b). In call accounting the stale answers of the certified policies are $9.2$ (per candidate), $12.3$ (restart only) and $2.0$ (ours) per $10^4$; the counters serve $55$--$56\times$ more than per candidate pooled ($38$--$86\times$ per model), and the counter/restart-only ratio is $42\times$ [$32$, $56$] (ExpeL).

\paragraph{(b) Per-memory weights at matched false retirements.} All rules run on the same per-candidate certificate and differ only in how they retire a folded memory. The per-memory LR-CUSUM uses its default threshold $h=7$. Each fixed failure:success weight $k{:}1$ (score $+1$ per audited failure, $-1/k$ per success), for $k\in\{1,2,7,25,50,100,200,500,1000,5000\}$, gets its best threshold (between $1$ and $2{,}584$ failure units; a finer grid from $34$ to $89$ for $7{:}1$ and $5000{:}1$) among those with no more false retirements per stream than the LR-CUSUM (a retirement of a memory at or above break-even), a choice that favors the fixed rules. We measure \emph{stale exposure}: folded uses of a memory while it is stale, per stream. Pooled, the LR-CUSUM has $702$ stale uses per stream at $33.5$ false retirements; the best fixed weight ($5000{:}1$, threshold $78$, $31.7$ false retirements) has $7{,}300$, i.e.\ $10.4\times$ [$7.7$, $13.7$] more, and the symmetric $1{:}1$ rule $29.1\times$ [$14.6$, $51.7$] more (\cref{tab:app-cs-b}). On every model the best fixed weight is $7.6$--$11.4\times$ worse, with every lower confidence bound at least $3.5$. On the shared threshold grid, exposure falls from $25{:}1$ to $100{:}1$ and is flat from $100{:}1$ to $1000{:}1$ (within $3\%$): there a success barely discounts a failure, and all weights converge to one rule, a failure count that never forgets; $5000{:}1$, this limit tuned on a finer threshold grid, is lower still. The optimum is interior on Mistral-7B ($7{:}1$) and this limit on the other four models.

The theory predicts both the limit and the gap. In $86\%$ of the $1{,}456$ stale episodes the memory was perfect before the change ($q_0=0$): a failure cannot occur before the change, so the optimal weight $1/\rho(q_0,q_1)$ of \cref{eq:rho} is infinite, and the best global weight moves to the no-forgetting limit. The other episodes want the opposite: $1/\rho=0$ when $q_1=1$ ($9\%$) and a median of $1.0$ otherwise ($5\%$). No single weight fits both groups, and the limit is still $10\times$ worse because a global rule also needs a global threshold. To spare the healthy memories that do fail now and then, it must wait for about $78$ failures. The per-memory LR-CUSUM instead charges each failure $\ln(q_1/\hat q_0)$ of its own memory, which is large when the memory never failed on its shadow labels. Among retired episodes of perfect memories, it retires after a median of $30$ stale uses, about three audits; the best fixed rule needs $874$. This is also why the gap is far larger than on \textsc{KQA-Stream} ($1.3$--$5.9\times$ in detection delay) and on natural edits ($1.1$--$1.6\times$), where few memories are perfect before the change. $42\%$ of the episodes are partial ($q_1<0.95$; median $q_1=1$).

\begin{table}[h]
\centering
\small
\setlength{\tabcolsep}{2.4pt}
\resizebox{\columnwidth}{!}{%
\begin{tabular}{@{}lrrrlr@{}}
\toprule
& \multicolumn{3}{c}{\textbf{LR-CUSUM} ($h=7$)} & \multicolumn{2}{c}{best fixed weight ($1{:}1$--$5000{:}1$)} \\
\cmidrule(lr){2-4}\cmidrule(l){5-6}
Model & FR & stale uses & episodes & weight ($h$) & $\times$ LR [95\% CI] \\
\midrule
Qwen3-8B & 38.8 & 968 & 196 & $5000{:}1$ (84) & $7.6$ [$3.5$, $21.7$] \\
Qwen3-4B & 29.8 & 656 & 234 & $5000{:}1$ (84) & $9.8$ [$5.8$, $16.7$] \\
Granite-3.3-8B & 47.5 & 526 & 281 & $5000{:}1$ (42) & $8.2$ [$6.4$, $10.9$] \\
Mistral-7B & 29.7 & 813 & 468 & $7{:}1$ (84) & $11.4$ [$9.3$, $13.9$] \\
OLMo-3-7B & 21.8 & 550 & 277 & $5000{:}1$ (46) & $9.6$ [$7.5$, $12.0$] \\
pooled & 33.5 & 702 & 1{,}456 & $5000{:}1$ (78) & $10.4$ [$7.7$, $13.7$] \\
\bottomrule
\end{tabular}}
\caption{CodeStream claim (b), step accounting. FR: false retirements per stream; stale uses: folded uses while stale per stream; $\times$LR: the fixed rule's stale uses over the LR-CUSUM's at no more false retirements. $5000{:}1$ is the no-forgetting limit, tuned on a finer threshold grid; on the shared grid exposure is flat from $100{:}1$ to $1000{:}1$. Bold: ours.}
\label{tab:app-cs-b}
\end{table}

\paragraph{(c) Per-epoch false certification.} An \emph{epoch under test} is a (task, version) pair in which a candidate is under test, from its formation to its fold or discard; a \emph{bad fold} folds a candidate whose reliability at that version is below break-even, the task-level event of \cref{fig:epoch}b. Ours has bad folds in $6$ of $965{,}730$ epochs at $\delta=0.01$ ($6.2\cdot10^{-6}$), $12$ of $946{,}678$ at $\delta=0.05$ ($1.3\cdot10^{-5}$; Clopper--Pearson upper bound $2.2\cdot10^{-5}$), $16$ of $934{,}466$ at $\delta=0.1$ ($1.7\cdot10^{-5}$) and $21$ of $917{,}079$ at $\delta=0.2$ ($2.3\cdot10^{-5}$): the rate rises with $\delta$ and stays more than $1{,}000\times$ below it. Per model, every rate is at least $250\times$ below $\delta$; the closest to $\delta$ is Mistral-7B's at $\delta=0.01$ ($6$ of $161{,}262$). With changelog notices it is $0$ of $941{,}085$ at $\delta=0.05$, and in call accounting $10$ of $968{,}421$. The counters fold a below-break-even candidate in $4.8\%$ (ExpeL, $4{,}625/97{,}172$) and $5.4\%$ (ACE, $5{,}733/106{,}424$) of their epochs under test, fold once in $3.7\%$; these policies fold at formation, so their epochs under test are formation epochs. The guarantee of \cref{prop:app-ledger}(iv) does not cover the API edits, which change live reasoning silently; the measured rates show that, here too, this boundary is not where certification fails.

\paragraph{(d) Accuracy and cost.} Live reasoning is right on $76.2\%$ of queries ($57.5$--$83.1\%$ per model). Ours gains $8.7$ points [$7.1$, $10.5$]: $2.4$ (Qwen3-4B), $3.1$ (Qwen3-8B), $7.2$ (OLMo-3-7B), $10.0$ (Granite-3.3-8B) and $21.0$ (Mistral-7B) per model, because a certified program has passed hidden inputs that the live agent's visible tests do not cover. The cost depends on the unit. In \emph{steps} ours costs $0.62$ fewer per query [$0.52$, $0.71$] ($-23\%$) and lowers $J_{20}$ by $5.3$ [$4.4$, $6.3$]; in \emph{LLM calls} it costs $0.29$ more per query [$0.20$, $0.39$] ($+17\%$), because shadow mode pays for live reasoning while a candidate is under test, and still lowers $J_{20}$ by $2.6$ [$2.0$, $3.3$]. The certified policies have the highest accuracy and the lowest $J_{20}$ in both units. Counters gain $6.0$--$6.2$ points at a lower cost in both units (on Qwen3-8B and Qwen3-4B they are $0.4$--$0.8$ points \emph{below} live reasoning) while serving $47\times$ more stale answers than restart only; fold once is within noise of live reasoning ($+0.3$ points [$-1.9$, $2.5$]).

\paragraph{Ties: the certification load on a real domain.} The cost unit changes how much certified reuse the stream allows, and \cref{thm:betting}(ii) says why. Audited memory--live pairs of the certified policies agree on the outcome in $94$--$97\%$ of cases ($84$--$99\%$ per model), and live reasoning re-derives the memory's exact program in $49$--$59\%$ ($21$--$79\%$ per model; $21$--$33\%$ for Mistral-7B and OLMo-3-7B). In call accounting, a tie in which live reasoning succeeds at its first attempt has paired utility exactly $0$; this holds for $84$--$86\%$ of audited pairs ($67$--$89\%$ per model). The utility law then has a large atom at zero, so $\KLinf(P_Z)$ is small and certification is slow, and one-step coverage drops to $0.32$, against $0.61$ in step accounting, where every tie is positive evidence because live reasoning costs at least one step more than the memory. For ours, the drop tracks the share of zero-utility pairs across models: Qwen3-8B and Qwen3-4B ($88$--$89\%$) keep $34$--$36\%$ of their step-accounting coverage, Granite-3.3-8B and OLMo-3-7B ($86$--$87\%$) $52$--$53\%$, and Mistral-7B ($68\%$) $81\%$. This is the certification load $\Th\propto1/(p\KLinf)$ of \cref{sec:limit,sec:certify} observed on a real domain: an exact verifier and a strong live reasoner make audits nearly uninformative, and safe contraction must wait for the queries on which the two differ.

\paragraph{Scope.} API edits break few programs: in the pilot, A1 breaks $3.7$--$4.9\%$ of the distinct verified programs and the other seven edits together at most $7$ programs per model, because MBPP and HumanEval programs mostly use builtins. They test the unannounced channel and changelog notices, but we claim no staleness magnitudes for real library drift; a benchmark with version-specific APIs would measure them. Requirement revisions are controlled changes of real tasks, as the fact edits of \textsc{KQA-Stream} are.

\section{Worked examples from \textsc{KQA-Stream}}
\label{app:examples}

This appendix follows real Qwen3-8B macros through the replays of \cref{tab:realllm} (main setting).
Questions, unedited facts, LLM outputs, routing decisions and audit outcomes come from the logs; post-edit facts are counterfactual (\cref{app:experiments}).
Cases are named by edit sequence, stream seed and policy; $t$ is the query index in the stream.
Every replay used here was re-run with an instrumented recorder and matched the stored run exactly.
The values $r$, $a$, $m$ and $\rstar$ are exact popularity-weighted values over the instances routed to the template, with $L=20\bar m$ ($\bar m$: the stream's mean live calls).
A \emph{stale} answer is a folded answer that was correct at formation and is wrong now.
``Betting'' and ``SPRT'' denote betting + LR-CUSUM (per-candidate level $\delta$, fold threshold $\ln20$, no ledger; the ``per candidate'' row of \cref{tab:app-full}) and SPRT + LR-CUSUM. These are mechanism examples: the main-text results use the restart-only ledger certificate and the recommended deployment, whose first threshold is $\ln20+\ln(\pi^2/6)=3.49$ (so the $33$-label fold of \cref{tab:app-lifecycle}, at log-wealth $3.43$, is a per-candidate fold).

\subsection{One macro from formation to re-formation}
\label{app:ex-lifecycle}

\begin{table*}[t]\centering\small\setlength{\tabcolsep}{3.5pt}
\begin{tabular}{@{}r l p{0.445\textwidth} l@{}}
\toprule
$t$ & Stage & Real evidence (Qwen3-8B, template T118, edit sequence 303, stream seed 6) & Statistic \\
\midrule
44 & form & Audited live trace of ``What individuals have their residence in the state where Louisiana Tech University is situated?'' (3 calls, 1{,}300/59 tokens): \texttt{hop located in the administrative territorial entity (fwd)} $\to$ Louisiana; \texttt{hop residence (bwd)} $\to$ Elmore Leonard; \texttt{answer type human}. Correct. & $r{=}1.00$, $\rstar{=}0.772$ \\
44--10{,}769 & certify & 33 audited shadow labels: 26 both correct ($U{=}{+}2$), 7 live wrong and macro correct ($U{\approx}{+}70$), 0 macro failures & $\ln W^+\!: 0\to3.43\ge\ln 20$ \\
10{,}769 & fold & routing call \texttt{\{"type": 1\}} + lookup in a 272-anchor table frozen at $v_0$ & LR-CUSUM $q_0{=}.015$, $q_1{=}.294$, $h{=}7$ \\
$\to$93{,}194 & reuse & 2{,}982 one-call answers before the edit, all correct (live: .798 on the same queries); 5{,}566 LLM calls (64.8\%) saved over the macro's 3{,}026 folded uses & \\
93{,}194 & edit $v_{16}$ & \emph{located in the administrative territorial entity}, $f{=}0.6$ (451 facts); e.g.\ Cardiff University: Cardiff $\Rightarrow$ Hong Kong, gold John Williams $\Rightarrow$ Bruce Lee, Ip Man, Stephen Chow; the macro still answers John Williams & $r$: $1.00\to0.367{<}\rstar{=}0.491$ \\
93{,}227--94{,}247 & retire & audited folded labels 0,1,0,0 (Univ.\ of Basel, Univ.\ of Windsor, East Tennessee State Univ., Cardiff Univ.); 1{,}053 queries, 43 folded uses and 29 stale answers after the edit & $S$: 2.99, 2.66, 5.66, 8.65${>}7$ \\
94{,}378--98{,}104 & re-form & trace (valid at $v_{16}$) of ``Who lives in the state where University of Freiburg is located?''; 12 labels (7 discordant); then 3{,}694 folded uses to the end, 0 wrong (0 of 334 audited labels fail) & $\ln W^+{=}3.00$ \\
\midrule
93{,}194--200k & other rules & ExpeL-style, ACE-style and fold once do not retire the macro before the stream ends: 3{,}863 folded uses and 2{,}427 stale answers after the edit. SPRT: fold at 20{,}978, then the same retirement (29 stale answers). Compiled plan + betting: 0 stale answers, 0 of 346 post-edit labels fail & \begin{tabular}[t]{@{}l@{}}ExpeL: $333\to249$\\ ACE: $331/0\to462/215$\end{tabular} \\
\bottomrule
\end{tabular}
\caption{Lifecycle of one real macro under betting + LR-CUSUM, and the same macro under other rules (last row). $t$: query index in the 200k-query stream. Stale answers include the retiring query. Post-edit facts are counterfactual by construction.}
\label{tab:app-lifecycle}
\end{table*}

One real macro passes through every stage of the lifecycle and illustrates \cref{eq:breakeven,thm:betting,cor:lr,prop:unclipped}.
Template T118 maps a university to the people living in its state (edit sequence 303, seed 6).
It is the fifth most popular template in this stream ($\pi=0.036$).

\paragraph{Formation and certification.}
The macro forms at $t=44$ from the first audited T118 query (\cref{tab:app-lifecycle}).
At its second step the live agent outputs \texttt{\{"action": "hop", "relation": "residence (backward)", "direction": "backward"\}}.
Both one-call baselines fail; the parametric one outputs \texttt{\{"answers": ["State of Louisiana"]\}}.
At $v_0$ the macro is always correct, while live has $a=0.800$ at $m=2.85$ calls, so $\rstar=0.772$ with $L=66.98$.
The 26 concordant pairs add 0.010--0.019 each to $\ln W^+$; the 7 pairs where only live fails add 0.33--0.52 each.
The SPRT plugs in $r_0=0.756$ and $r_1=0.788$ (true: 0.772 and 0.800) and needs 72 labels instead of 33.

\paragraph{Retirement and re-formation.}
Edit $v_{16}$ lowers the macro's reliability to $0.367<\rstar=0.491$, while live re-derives answers on the edited facts.
The LR-CUSUM was set at the fold to $q_0=0.0147$ and $q_1=1-\hat r^\star=0.294$.
It scores $+2.995$ per failure and $-0.333$ per success.
One failure thus weighs about nine successes ($\rho=0.111$ in \cref{eq:rho}), and three failures in four labels retire the macro.
The SPRT retires it on the same labels at the same $t$.
The re-formed candidate needs only 12 labels, 7 of them discordant, because live accuracy has fallen to 0.517.

\paragraph{Counters and plans.}
The ExpeL-style, ACE-style and fold-once policies do not retire the macro before the stream ends (last row of \cref{tab:app-lifecycle}).
Its 331 successful audits before the edit bank an ExpeL importance of 333, and \cref{prop:unclipped}(a) predicts a delay of $333/(2\cdot0.633-1)=1{,}249$ labels; only 346 remain.
The compiled plan, which re-executes the path on current facts, has no failure in 677 audited labels.

\paragraph{How typical is this case?}
Among betting's 81 stale-after-edit onsets in the 30 streams, this one is at the median in folded uses to retirement (43); their median stale count is 30 (IQR 21--47).
The counter side is extreme (98.5th percentile of ExpeL onsets): T118 is popular, and 106,806 queries remain.
In seeds 1 and 2 the counters serve 358 and 273 stale answers after this edit, and betting 52 and 40.

\paragraph{Takeaway.}
Certification took 33 labels, and a likelihood-ratio detector limited a real edit to 29 stale answers.
Counters with banked credit kept the stale macro to the end of the stream.

\subsection{Counters under partial staleness}
\label{app:ex-counters}

\begin{table*}[t]\centering\small\setlength{\tabcolsep}{4pt}
\begin{tabular}{@{}lrrrrrr@{}}
\toprule
& & & & & \multicolumn{2}{c}{$A$ at in-control rate} \\
\cmidrule(l){6-7}
Rule & at edit & retires at label & uses / stale & $\E\tau$ & $q_0{=}.05$ & $q_0{=}.013$ \\
\midrule
ExpeL-style importance ($\pm1$) & 92 & not by end (239) & 3{,}283 / 894 & $\infty$ & $\infty$ & $\infty$ \\
ACE-style helpful/harmful & 90/0 & not by end (330/93) & 3{,}283 / 894 & $\infty$ & $\infty$ & $\infty$ \\
symmetric, $H{=}2$ (SPRT + symmetric)$^\dagger$ & 0 & 19 & 169 / 47 & 17.8 & 420 & $6.2{\times}10^3$ \\
symmetric, $H{=}4$ & 0 & 152 & 1{,}460 / 374 & 180.6 & $1.5{\times}10^5$ & $3.7{\times}10^7$ \\
LR $+7/{-}1$, $H{=}43$ & 0 & 44 & 389 / 103 & 37.1 & $2.0{\times}10^5$ & $9.7{\times}10^9$ \\
\textbf{LR-CUSUM as run} (SPRT), $h{=}7$$^\dagger$ & 0 & 19 & 169 / 47 & 13.7 & 249 & $2.0{\times}10^4$ \\
\bottomrule
\end{tabular}
\caption{Retention rules on the same 333 real post-edit labels of the T048 macro (edit sequence 303, seed 0). Uses / stale: folded uses before retirement / stale answers incl.\ the retiring query (to the stream's end for ``not by end''). $\E\tau$: exact mean delay (labels) at $q_1=0.267$. $A$: exact mean labels to a false retirement; .013 is the deployed LR-CUSUM's in-control estimate. $^\dagger$A real replay retired the macro at label 19. Bold: ours.}
\label{tab:app-counters}
\end{table*}

On one real sequence of audit outcomes, this example compares all retention rules and checks \cref{prop:unclipped,thm:trichotomy,cor:lr}.
The T048 macro maps a film to Academy Awards via \emph{producer} and \emph{nominated for} (edit sequence 303, seed 0).
It forms at $t=738$ from the trace for ``What Academy Awards did the producer of Man of Steel get nominated for?'' and passes all 90 audits before the edit.
Edit $v_5$ gives a new producer to 25\% of the relevant subjects.
Goodfellas' producer Irwin Winkler, for example, becomes Eminem, yet the macro still answers Best Picture.
Its reliability falls to $0.733<\rstar=0.768$, so $q_1=0.267$: partial staleness.
The 333 post-edit labels contain 93 failures.

\Cref{tab:app-counters} replays every rule on these labels.
By \cref{prop:unclipped}(b), ExpeL (credit 92) and ACE (90/0) are ever retired with probability at most $10^{-40.4}$ and $10^{-39.5}$.
The symmetric rule is in the negative-drift regime of \cref{thm:trichotomy} ($\mu_1=-0.467$, $\gamma=0.344$).
At a similar $A$, the $+7/{-}1$ rule of \cref{cor:lr} retires at label 44 instead of 152.
Every realized delay lies between the 57th and 81st percentile of its exact law.
On the longest real onset (T164, sequence 101, seed 5; 4,004 labels), the counters serve 9,997 stale answers.
There the symmetric rule with $H=4$ needs 610 labels and the LR-CUSUM 8.
Over the 30 streams, ExpeL has not retired 95.5\% of its 287 partial-staleness onsets by the end of the stream, and ACE 80.8\% of its 312.
The bounds of \cref{prop:unclipped}(b), computed from each onset's real credit, average 0.89 and 0.76.
The LR-CUSUM detectors leave 1 of 24 (SPRT) and 0 of 11 (betting) unretired.

\paragraph{Takeaway.}
Under partial staleness, $\pm1$ counters with banked credit are practically never retired: the probability of ever retiring them is small (\cref{prop:unclipped}(b)), and on real streams they stay in service until the stream ends.
A failure must weigh $1/\rho(q_0,q_1)$ successes, about 6.5 here.

\subsection{Certifying utility instead of reliability}
\label{app:ex-utility}

\begin{table*}[t]\centering\small\setlength{\tabcolsep}{4pt}
\begin{tabular}{@{}rrllrrrrr@{}}
\toprule
\# & $t$ & instance (T010) & $(Y^M,Y^L)$ & $U_t$ & $Z_t$ & $\ln W^+$ & $\ln W^-$ & plug-in LLR\\
\midrule
1 & 3179 & Q2656 & (1,1) & $+2$ & $+.027$ & .011 & $-.011$ & .096\\
11 & 27863 & Q30002 Chengdu & (0,0) & $+2$ & $+.027$ & .121 & $-.115$ & .910\\
17 & 34168 & Q25416 & (1,1) & $+2$ & $+.027$ & .190 & $-.174$ & 1.342\\
18 & 35098 & Q48320 K\=obe (stale) & (0,1) & $-64.4$ & $-.877$ & $-.290$ & .096 & 1.295\\
19 & 35271 & Q1467 Cebu City (stale) & (0,1) & $-64.4$ & $-.877$ & $-.599$ & .402 & 1.248\\
35 & 59956 & Q1467 & (0,1) & $-64.4$ & $-.877$ & $-1.573$ & \textbf{3.226} discard & 1.781\\
94 & 187133 & Q123709 & (1,1) & -- & -- & -- & -- & \textbf{3.007} fold\\
\bottomrule
\end{tabular}
\caption{Real certificates for the T010 candidate (edit sequence 202, seed 1). \#: audited label, shared by betting and the SPRT; labels 1--11 precede edit $v_7$. Threshold $\ln 20=2.996$ for $\ln W^+$ (fold), $\ln W^-$ (discard) and the plug-in LLR (fold).}
\label{tab:app-utility-example}
\end{table*}

\begin{table*}[t]\centering\small\setlength{\tabcolsep}{3.2pt}
\begin{tabular}{@{}lrrrrrrrl@{}}
\toprule
Template (rank) & $\pi_i$ & audits & $\nstar_i$ & $\nstar_i/(p\pi_i)$ & epoch & $e^{-\Th_i/\pi_i}$ & one-step & labels to fold \\
 & & in stream & & (queries) & (queries) & predicted & observed & SPRT / \textbf{betting} \\
\midrule
T132 (1)   & .258   & 5{,}042 & 642 & 24.9k & 200k       & .883          & .994 & 19 / 9 \\
T048 (8)   & .0214  & 430     & 210 & 98.2k & 100k$^{a}$ & .375          & .738 & 38 / 106 \\
T027 (10)  & .0165  & 323     & 213 & 129k  & 200k       & .524          & .906 & 38 / 106 \\
T110 (73)  & .00154 & 31      & 213 & 1.38M & 100k$^{b}$ & $<10^{-5}$    & 0    & not in stream (8 of 39$^{b}$) / not in stream \\
T115 (204) & .00043 & 6       & 208 & 4.87M & 200k       & $<10^{-10}$   & 0    & not in stream (5 of 38) / not in stream \\
T014 (294) & .00025 & 3       & 171 & 6.92M & 200k       & $<10^{-15}$   & 0    & not in stream (2 of 35) / not in stream \\
\bottomrule
\end{tabular}
\caption{Six templates of one stream (edit sequence 101, seed 0). $\pi_i$: share of foldable queries. Epoch: $1/(\nu s_i+1/T)$, $s_i$: fraction of harmful edits. One-step: share of the template's queries answered by a folded macro, stale ones included. Bold: ours. $^a$Edit $v_{19}$ ($t=149{,}426$); retired at $t=161{,}145$, re-certified after 77 labels at $t=194{,}254$. $^b$Edit $v_7$ ($t=55{,}476$) drops the candidate's reliability from 1.0 to .733 after 8 of 39 labels.}
\label{tab:app-coverage}
\end{table*}

Two real candidates show why \cref{thm:betting} bets on the paired utility and how a plug-in break-even misleads the SPRT (\cref{prop:plugin}).
The replays bet on $Z_t=U_t/(L+7)$, clipped to $[-1,1]$, with $U_t=(m_t-1)+L(Y^M_t-Y^L_t)$ and a uniform mixture of seven fixed bets.
This differs from the grid mixture of \cref{app:utility} but keeps validity; mirrored bets ($W^-$) decide discards.

\paragraph{(a) Less accurate than live, yet worth folding.}
Template T096 maps a record label to its time zone and has rank 2 in seed 9 of sequence 101 ($\pi=0.109$).
Its macro forms at $t=100$ from the trace for ``Which time zone is associated with Imperial Records?''.
It follows the gold path, and no edit changes its answers.
Still $r=0.997<a=0.998$, because 21 misrouted instances (mass 0.003) count as macro failures.
For example, ``What time zone does the headquarters of Fox Searchlight Pictures observe?'' belongs to T281, but the router outputs \texttt{\{"type": 3\}} and picks T096.
Live, which does not route, answers correctly, so the pair enters with $U_t=-65.3$.
The macro is still above break-even ($\rstar=0.968$) and worth $+1.93$ steps per query.
Betting folds it at $t=22{,}459$ after 249 labels: 247 pairs with both correct, one routing loss and one joint failure.
No discordant pair favors the macro, so a sign test on discordant pairs would never fold it.
Betting folds on the concordant gain $\eta/(1-\eta)$ of \cref{app:utility}.
All three numbers below are for the \emph{per-candidate} certificate at level $\delta$ (fold threshold $\ln(1/\delta)=\ln20$, no ledger term), the version this example ran. \Cref{lem:app-klinf} gives $\KLinf(P_U)=0.0244$, i.e.\ $\ln20/\KLinf=123$ labels to first order (the \emph{optimum} level of \cref{sec:certify}); the implemented seven-bet mixture adds $\ln 7$ nats to the threshold and grows at the rate $g_{\mathcal B}=0.0223$ of its best bet ($0.9$), which predicts $(\ln20+\ln7)/g_{\mathcal B}=221$ (\emph{as run}: the implementation loss; the overshoot term $\ln1.9$ of \cref{app:ledger} is not included); the 249 labels it took are the \emph{measured} level, $13\%$ above the as-run prediction through finite-sample effects. The deployed ledger's $j$-th test adds $2\ln j+\ln(\pi^2/6)$ nats; for $j=1$ the first two levels become $(\ln20+\ln\frac{\pi^2}6)/\KLinf=143$ and $(\ln20+\ln7+\ln\frac{\pi^2}6)/g_{\mathcal B}=244$ labels.
The SPRT folds earlier ($t=6{,}537$, 67 labels); its plug-in error is harmless here because the macro is good.

\paragraph{(b) A plug-in break-even folds a stale macro.}
The T010 candidate (sequence 202, seed 1) forms at $t=651$ from ``Which time zone is associated with the administrative twin of Nottingham?''.
Before the edit, $r=0.965>\rstar=0.947$: a thin margin that needs about 686 labels, while only 11 arrive.
Edit $v_7$ then gives a new time zone to 25\% of the relevant subjects.
Seattle, for example, moves from Pacific Time Zone to East Africa Time.
The cached macro drops to $r=0.597$, while live stays at $a=0.977$.
Betting sees 9 stale pairs with $U_t=-64.4$ and discards the candidate at $t=59{,}956$ (\cref{tab:app-utility-example}).
The SPRT froze $\hat a=0.345$ from one shrunk key label, so $r_0=0.313$ and $r_1=0.345$.
Its zero-drift point, 0.329, lies below the true $r$ in every phase.
As in the proof of \cref{prop:app-plugin}, it therefore folds almost surely.
It folds at $t=187{,}133$, 0.51 below break-even, and serves 45 stale answers.
Across the 30 streams, the SPRT makes 21 below-break-even folds and betting 6 (SpF in \cref{tab:realllm,tab:app-full}).
In all 21 SPRT cases the plug-in $r_0$ lay below the true $\rstar$, and betting folded none of those candidates.

\paragraph{Takeaway.}
Pair the macro with live on the same instance.
Routing errors and live mistakes then enter the certificate directly, and $a$ need not be estimated.

\subsection{Why rare templates stay uncovered}
\label{app:ex-coverage}

Six templates of one stream show the mechanism of \cref{thm:limit}: a template is covered only if its epoch outlasts $\nstar_i/(p\pi_i)$.
We use edit sequence 101 and seed 0, with SPRT certification.
T048, T027, T110 and T115 have matched difficulty: live solves 59 of 60 instances, the gold-path macro has $r=1.000$, and $\nstar_i$ is 208--213.
Only their popularity ranks differ (\cref{tab:app-coverage}).

The head certifies within 38 labels; the tail never does.
T110 forms at $t=5{,}301$ from a correct trace for ``What record labels are connected to graduates of Radcliffe College?''.
It gets 8 of the 39 labels it needs before edit $v_7$ lowers its reliability to $0.733<\rstar=0.952$.
T115 and T014 get 5 and 2 labels in the whole stream.

In the 30 main-setting streams, the SPRT needs a median of 39 labels per fold (all 1,201 folds), fewer than the first-order $\nstar_i$ (208--213 for the templates above), which assumes the least favorable candidate.
Ranks 101--300 get 7.5--14.5 audits per template-stream on average, and none of their 5,771 template-streams was ever folded.

\paragraph{Takeaway.}
With a fixed audit budget, templates rarer than the load are not certified between changes; coverage comes from popular templates.

\section{Extended related work}
\label{app:related}

\paragraph{Reasoning memory and skill libraries.}
LLM systems increasingly store reusable reasoning: thought templates \citep{yang2024buffer}, strategies and insights \citep{suzgun2026dynamic,zhao2024expel,ouyang2026reasoningbank,zhang2026agentic,didolkar2025metacognitive}, workflows and programmatic skills \citep{wang2025agent,wang2025inducing,wang2024voyager,wang2024trove}, verified lemmas \citep{wang2024legoprover}, and long-term memories \citep{zhong2024memorybank,park2023generative,chhikara2025mem0,li2025memos}.
Many never retire an item \citep{yang2024buffer,ouyang2026reasoningbank,wang2025agent}. The rest use hand-set rules. ExpeL \citep{zhao2024expel} keeps an integer importance count that starts at 2, moves by $\pm1$ on up- and down-votes and deletes the item at 0. ACE \citep{zhang2026agentic} keeps unbounded helpful/harmful tallies. Other rules include clipped Hebbian strengths with time decay \citep{singh2025validation,zhu2026helamem}, lifetime failure rates thresholded at $1/2$ \citep{chen2026skilldroid}, EWMA utilities \citep{zhang2026memrl}, Beta counts \citep{forouzandeh2026learning}, usage-frequency pruning \citep{wang2024trove}, time-to-live \citep{li2025memos}, sliding-window utilities \citep{pu2026skillops} and paired with/without-memory audits \citep{zhang2026liveevo,gao2026skillaudit}. MQuAKE \citep{zhong2023mquake}, StreamingQA and RealTime QA \citep{liska2022streamingqa,kasai2023realtime} evaluate a model after knowledge edits; \textsc{KQA-Stream} evaluates a memory lifecycle over a stream with recurring templates and interleaved edits.
The statistical analyses we know of either prove stationary consistency of a two-counter ``memory worth'' \citep{simsek2026when} or prove that an EMA converges \citep{zhang2026memrl}. \citet{sun2025beyond} recast memory management in decision-theoretic terms but give no tests. \citet{asadolahi2026memory} show that self-graded memory scores are inflated unless verifier errors are independent.
Sequential tests enter LLM pipelines only to audit trajectories \citep{sadhuka2025evaluator} or to stop sampling within one query \citep{lee2025consol}.
\emph{We show that these maintenance rules are change detectors: clipped counters are Page CUSUMs with mis-specified scores, and unclipped tallies have no renewal. We quantify their detection delay and replace them with certified folding and likelihood-ratio retirement.}

\paragraph{Step skipping versus system-level contraction.}
Model-level methods train an LLM to skip or internalize steps \citep{liu2024can,yu2024distilling,hao2025training}, to compress chains of thought \citep{xia2025tokenskip}, or to decide when to think \citep{zhang2025adaptthink}; see \citet{sui2025stop} for a survey. These methods change what one forward pass can compute. They choose training data offline. At inference time nothing decides whether a template's one-step answer can be trusted, or notices when it has gone stale.
The problem is real. Latent multi-hop composition is unreliable \citep{yang2024do}, and consensus-based early exit is unsafe \citep{mo2026stable}.
\emph{Contraction is complementary. It treats any distilled routine as a macro and decides per template, over time and with explicit error control, when to serve that macro and when to retire it.}

\paragraph{Caching, routing and cascades with guarantees.}
Semantic caches reuse responses for similar prompts \citep{bang2023gptcache}, and agent systems cache plan templates \citep{zhang2025agentic}. Caching theory gives optimal content under stationary demand \citep{zhu2023towards,liu2026semantic}. Cascades and routers decide per query \citep{chen2024frugalgpt,aggarwal2024automix}.
Where these systems carry guarantees, the guarantees are fixed-sample risk control under exchangeability. Examples are CALM \citep{schuster2022confident}, which uses Learn-then-Test \citep{angelopoulos2025learn}, PAC and group-conditional reasoning \citep{zeng2025provable,huang2026conditional}, and conformal early stopping \citep{wang2026conformal}.
Two works are closer.
\begin{itemize}
  \item B-PAC \citep{yu2026anytime} certifies a global thinking/non-thinking router anytime, with inverse-propensity-weighted test supermartingales. Our certificate uses the same tools (betting, inverse-propensity weights, Ville's inequality); what differs is the object: one certificate per template, restarted at changes and followed by retirement, over a memory's whole lifecycle.
  \item vCache \citep{schroeder2026vcache} bounds the per-request error of cache hits using randomized LLM verification. Its bound assumes i.i.d.\ prompts and a correct similarity model, and exploration can stop entirely.
\end{itemize}
Staleness has been studied only empirically \citep{mansoor2026risk,wu2026invalidation}, although model behavior does drift \citep{chen2024how}. Anytime-valid monitors watch whole models, not reusable items \citep{podkopaev2022tracking,richter2025auditing,khosravi2026conformal}.
\emph{We certify a reusable derivation with a test that stays valid at all times, under adaptive shadowing and arbitrary drift below the null. The indifference zone comes from the loss. Certification is followed by quickest-detection retirement, and the result yields population-level coverage laws.}

\paragraph{Sequential testing and change detection.}
Certified folding rests on the SPRT \citep{wald1945sequential,wald1948optimum}, Ville's inequality \citep{ville1939etude}, and testing by betting \citep{shafer2021testing,ramdas2023game,howard2021time,waudbysmith2024estimating}.
Retirement rests on Page's CUSUM \citep{page1954continuous} and its optimality \citep{lorden1971procedures,moustakides1986optimal,lai1998information}. Two of our results are instances of older work. The counter identity matches cumulative-score and integer CUSUM charts \citep{munford1980control,brook1972approach}. Our likelihood-ratio learning rates are the Bernoulli CUSUM \citep{reynolds1999cusum}.
Analyses of mis-specified CUSUM assume positive post-change drift \citep{molloy2017misspecified,cooper2024quickest}, or note only qualitatively that delay can exceed the false-alarm time \citep{molloy2021misspecified}. \citet[Thm.~2]{wu2016quickest} state that the delay is infinite in the negative-drift regime; their proof exchanges an expectation and a minimum over the CUSUM's restarts.
Observation-controlled detection \citep{banerjee2013data,xu2021optimum} and e-detectors \citep{shin2024edetectors} are the natural comparisons for audited and composite settings.
\emph{Using Cram\'er--Lundberg asymptotics \citep{iglehart1972extreme,siegmund1985sequential,asmussen2003applied}, we make explicit the full delay--ARL phase diagram for additive-score detectors. It includes the power law $D\asymp A^{\theta_1/\theta_0}$ under negative drift, where \citet{wu2016quickest} state an infinite delay; \cref{app:hebbian} locates the step of their proof that fails and gives a counterexample.}

\paragraph{Occupancy, freshness and scaling laws.}
Our static amortization law is a corollary of occupancy asymptotics \citep{karlin1967central,gnedin2007notes}. Its constant follows from rare-mass results \citep{ohannessian2012rare}, and Good--Turing theory estimates and forecasts it \citep{good1953population,good1956number,mcallester2000convergence,orlitsky2016optimal}.
For $n_0=1$ it is the cold-start or characteristic-time miss ratio of caching \citep{fagin1977asymptotic,che2002hierarchical,jelenkovic1999asymptotic}. Admission after $k$ hits generalizes this case \citep{martina2014unified}, and memorization learning curves share the exponent \citep{hutter2021learning,michaud2023quantization}.
Under drift, the $n_0=1$ steady state is the freshness objective of crawling \citep{azar2018tractable} and the hit law of TTL and random-replacement caches \citep{fofack2014performance,gallo2014performance}.
Square-root refresh and replication rules are classical \citep{yates2017age,cohen2002replication,kolobov2019staying}, and so is abandoning fast-changing items \citep{cho2003effective,bastopcu2021information}. Square-root allocations also appear in label-efficient inference \citep{xu2024active,zrnic2024active}.
The law of practice has been explained by chunk recurrence and by averaging over items \citep{newell1981mechanisms,logan1988toward,heathcote2000power}.
\emph{We couple sequential-test sample complexity with occupancy. This gives the fundamental limit of safe one-step coverage under drift, $M^\star\sim\kappa_s\Theta^{1-1/s}$, with a first-order matching converse for every $\delta$-safe policy.}

\paragraph{Safe self-improvement and multi-agent memory.}
Self-modifying agents accept a change when a benchmark score rises \citep{zhang2025darwin}. Per-decision betting gates \citep{shawn2026pace} and trajectory-level error spending \citep{wu2025sgm} control false commits. Earlier antecedents are high-confidence and Seldonian improvement \citep{thomas2015hcpi,thomas2019seldonian,chandak2020nonstationary} and the ``bio-creep'' study of approval policies \citep{feng2021biocreep}. Recent work proves limits for classifier-style gates \citep{scrivens2026limits} and near the reward ceiling \citep{cai2026safeharness}.
Shared multi-agent memories admit items through judges, one-shot A/B replay or count-and-mean rules \citep{zhang2025gmemory,xu2025sedm,yu2026comem}. Yet LLM errors are correlated \citep{kim2025correlated}, judges are noisy \citep{zhang2026ratchet}, and shared memory is a poisoning channel \citep{dong2025minja}.
\emph{These gates decide one modification or one shared item at a time. We study a single memory whose items are certified per template from verifier feedback, and leave shared and self-modifying memories to future work.}

\paragraph{Speed-up learning and the utility problem.}
Symbolic AI compiled multi-step reasoning into macro-operators \citep{fikes1972learning,korf1985macro}, explanation-based rules \citep{mitchell1986explanation}, chunks \citep{laird1986chunking} and compiled productions \citep{anderson2004integrated,taatgen2003production}. Program synthesis grows libraries in the same way \citep{ellis2021dreamcoder}.
The utility problem \citep{minton1990quantitative} led to statistically gated adoption \citep{gratch1992composer,greiner1996palo} and PAC analyses of speed-up learning \citep{tadepalli1996formal}. These analyses assume a fixed problem distribution and never un-adopt a rule.
\emph{We add four things: validity at all times under adaptive schedules and drift, an indifference zone derived from the loss, retirement with optimal detection delay, and closed-form population laws for how much reasoning can be contracted.}

\end{document}